\documentclass[letterpaper]{article} 
\usepackage[preprint]{aaai2027} 
\usepackage{pifont}
\usepackage[hyphens]{url} 
\usepackage{graphicx} 
\usepackage[numbers,square,sort&compress]{natbib} 
\usepackage{caption} 
\usepackage{booktabs}
\usepackage{makecell}
\usepackage{tabularx}
\usepackage{array}
\usepackage{colortbl}
\usepackage{listings}
\definecolor{tablegroup}{RGB}{232, 232, 232}
\definecolor{localintent}{HTML}{325A8C}
\definecolor{proceduralintent}{HTML}{B93C3C}
\definecolor{nextintent}{HTML}{E6B450}
\graphicspath{{figs/}}

\newcommand{\figref}[1]{\textbf{Fig.~\ref{#1}}}
\newcommand{\tabref}[1]{\textbf{Table~\ref{#1}}}
\newcommand{\tabrefpanel}[2]{\textbf{Table~\ref{#1}#2}}
\DeclareRobustCommand{\localcolor}[1]{\textcolor{localintent}{\textbf{#1}}}
\DeclareRobustCommand{\proceduralcolor}[1]{\textcolor{proceduralintent}{\textbf{#1}}}
\DeclareRobustCommand{\nextcolor}[1]{\textcolor{nextintent}{\textbf{#1}}}

\title{EgoIntent: A Pre-Outcome Micro-Step Benchmark for Understanding What, Why, and Next}
\author{
    Ye Pan\textsuperscript{\rm 1},
    Chi Kit Wong\textsuperscript{\rm 1},
    Yuanhuiyi Lyu\textsuperscript{\rm 1},
    Hanqian Li\textsuperscript{\rm 1},
    Chenfei Liao\textsuperscript{\rm 1},\\
    Jiahao Huo\textsuperscript{\rm 1},
    Lutao Jiang\textsuperscript{\rm 1},
    Zixin Zhang\textsuperscript{\rm 1},
    Jiacheng Chen\textsuperscript{\rm 4},\\
    Yuqian Fu\textsuperscript{\rm 3},
    Xu Zheng\textsuperscript{\rm 2}\thanks{Corresponding author.}
}
\affiliations{
    \textsuperscript{\rm 1}Hong Kong University of Science and Technology (Guangzhou)\\
    \textsuperscript{\rm 2}Great Bay University\\
    \textsuperscript{\rm 3}King Abdullah University of Science and Technology\\
    \textsuperscript{\rm 4}Sun Yat-sen University
}

\begin{document}

\maketitle

\begin{abstract}

Egocentric videos capture the world from a human perspective, providing a natural modality for studying human behavior. Yet conventional visual understanding, centered on recognizing scenes, objects, and actions, captures only the observable aspects of behavior. A more complete understanding requires reasoning about the latent intentions that motivate these actions and the goals humans seek to achieve.
Existing benchmarks for intent understanding primarily focus on coarse goals defined over entire events, overlooking how intent evolves across individual procedural steps. Capturing this evolution requires understanding three complementary dimensions: \ding{182} \textbf{\localcolor{Local Intent (What)}}, the immediate goal that a person is trying to accomplish at the current step; \ding{183} \textbf{\proceduralcolor{Procedural Intent (Why)}}, the role that the current step plays in the broader procedure; and \ding{184} \textbf{\nextcolor{Next-Plan (Next)}}, the action that is most likely to happen next.
To address this gap, we introduce \textbf{EgoIntent}, a step-level intent understanding benchmark comprising \textbf{3,014 steps} from 32 egocentric videos across 15 indoor and outdoor daily-life scenarios. We carefully curate videos rich in procedural steps, manually annotate each step along the three intent dimensions, and conduct multiple rounds of human review to refine temporal boundaries, resolve annotation inconsistencies, and ensure overall quality.
We comprehensively evaluate 15 MLLMs using reference-based score and complementary reference-free diagnostics. Controlled studies on four representative MLLMs further reveal that: (1) correct temporal order provides surprisingly limited benefits—only one model shows a significant temporal gain, while a single boundary frame significantly outperforms the full ordered clip for three models; (2) more context is not necessarily better—step-only input performs best for all four models, whereas adding 15 seconds of history significantly degrades three; and (3) revealing the current outcome and the next step selectively improves \localcolor{Local Intent} by 7.81 points and \nextcolor{Next-Plan} by 13.17 points, respectively. These counterintuitive findings show that current MLLMs can achieve strong intent-prediction scores through static boundary cues, without robustly exploiting temporal order or procedural history.

\end{abstract}

\begin{figure*}[t]
  \centering
  \includegraphics[width=\textwidth]{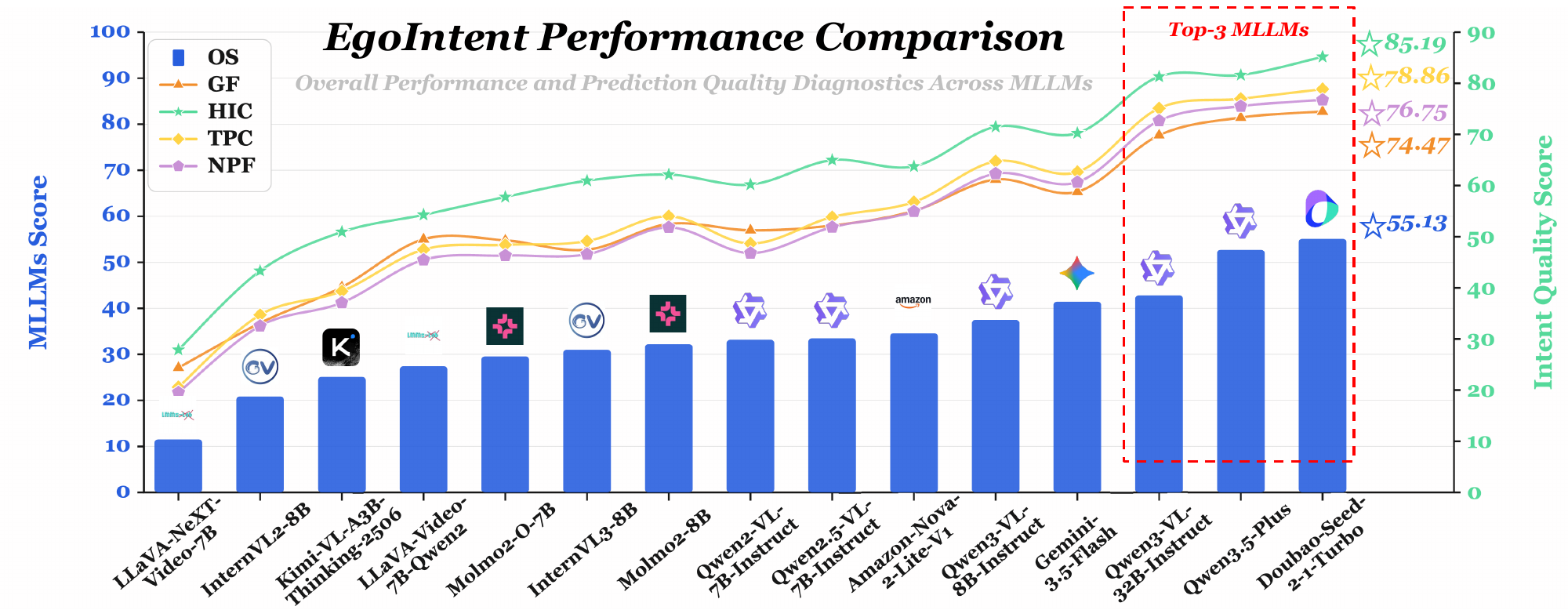}
  \caption{\textbf{Overall performance and intent-quality diagnostics across 15 MLLMs.} Blue bars report the reference-based Overall Score (OS) against the left MLLMs Score axis (0--100). The four curves report reference-free Visual Grounding Faithfulness (GF), Hierarchical Intent Consistency (HIC), Temporal Progression Consistency (TPC), and \nextcolor{Next-Plan} Feasibility (NPF) against the right Intent Quality Score axis (0--90). Models are ordered by OS; the dashed red box marks the three highest-OS systems, and the endpoint annotations give the leading model's scores. OS and the diagnostics use different Judges and evaluation targets, so they are complementary rather than directly comparable.}
  \label{fig:scores}
\end{figure*}


\section{Introduction}
\label{sec:intro}

\begin{figure*}[t]
  \centering
  \includegraphics[width=\textwidth]{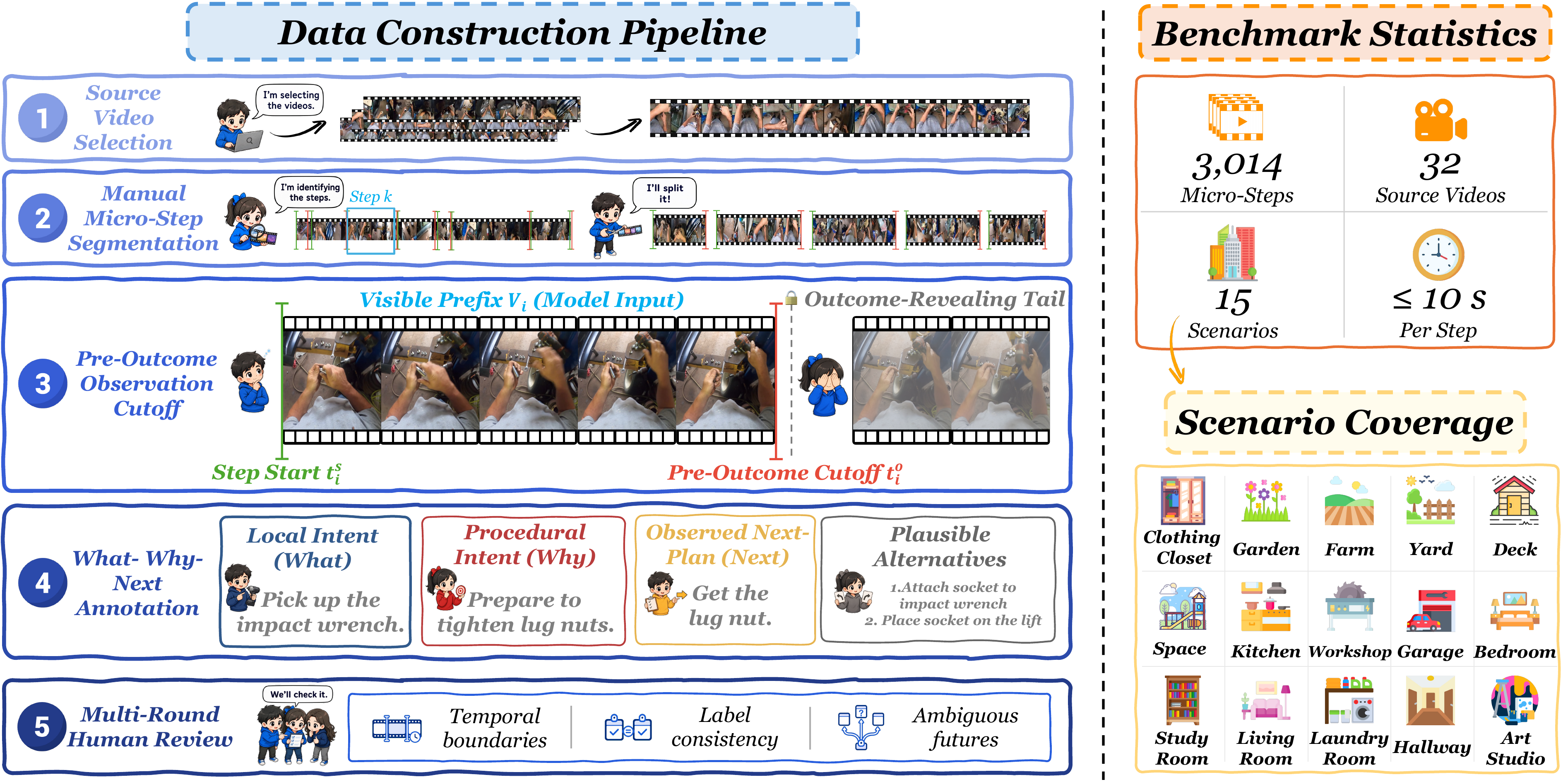}
  \caption{\textbf{EgoIntent data construction pipeline and benchmark composition.} The left panel follows five stages: (1) select coherent procedural source videos; (2) manually segment goal-directed micro-steps; (3) retain only the visible prefix $V_i$ from step start $t_i^{s}$ to the pre-outcome cutoff $t_i^{o}$, hiding the outcome-revealing tail and all later frames; (4) annotate \localcolor{Local Intent (What)}, \proceduralcolor{Procedural Intent (Why)}, the observed \nextcolor{Next-Plan (Next)}, and plausible alternatives; and (5) conduct multi-round human review of temporal boundaries, label consistency, and ambiguous futures. The right panel summarizes the resulting benchmark: 3,014 micro-steps, each at most 10 seconds, from 32 source videos across 15 indoor and outdoor scenarios.}
  \label{fig:method}
\end{figure*}

Recent Multimodal Large Language Models (MLLMs) have made rapid progress in visual perception and reasoning~\cite{hurst2024gpt,comanici2025gemini,bai2025qwen3,yang2025qwen3,wu2024deepseek}, bringing general-purpose embodied and wearable assistants closer to practical use. A useful assistant, however, should do more than react to explicit instructions~\cite{lu2024proactive}: it should infer a user's current intent~\cite{wen2025ai,peng2025eye} and anticipate likely needs from ongoing behavior~\cite{lee2025sensible,chu2025intention}. This requires reasoning beyond observable scenes, objects, and actions toward the latent goals that organize them across an ongoing procedure.

Egocentric video provides a natural testbed because it aligns the actor's actions, manipulated objects, and surrounding context in a first-person view~\cite{kulkarni2025egovita,vinod2025egovlm}. Recent benchmarks have begun to study intent-related reasoning in this setting~\cite{zhou2025x,peng2025eye,veerabadran2025benchmarking,chen2026egoplan,mangalam2023egoschema,li2026egocross,zhang2026egonight,zhu2026egosound}. For example, EgoGazeVQA~\cite{peng2025eye} studies gaze-grounded intent questions at the clip level, while WAGIBench~\cite{veerabadran2025benchmarking} evaluates episode-level goal inference for wearable assistants. In procedural activities, however, intent changes as an activity advances from one step to another~\cite{song2023ego4d}. Episode- or clip-level goals cannot fully characterize these fine-grained transitions~\cite{peirone2025hiero}: the same visible action may serve different immediate purposes, play different roles in a procedure, and support multiple plausible continuations.

\textbf{The key challenge is to infer what a person is trying to accomplish, why the current step matters, and what is likely to happen next before the outcome becomes visible.} To address this challenge, we introduce \textbf{EgoIntent}, a pre-outcome micro-step benchmark constructed from Ego4D source videos~\cite{grauman2022ego4d}. We manually identify and annotate 3,014 steps from 32 egocentric videos across 15 indoor and outdoor daily-life scenarios. Each step is described along three complementary dimensions: \textit{\localcolor{Local Intent (What)}}, the immediate goal within the current step; \textit{\proceduralcolor{Procedural Intent (Why)}}, the functional role of that step in the broader procedure; and \textit{\nextcolor{Next-Plan (Next)}}, the action most likely to follow. As summarized in \figref{fig:method}, models observe only the visible prefix ending immediately before the outcome-revealing tail; the tail and all subsequent frames remain hidden.

We evaluate 15 representative MLLMs with reference-based semantic scores and complementary reference-free diagnostics~\cite{liao2026benchmark}. Human baselines and Judge validation establish the answerability of the three dimensions and the reliability boundary of automatic evaluation. Controlled studies on four representative models further reveal a counterintuitive gap between intent prediction and temporal reasoning: only one model benefits significantly from correct frame order, a single boundary frame outperforms the full ordered clip for three models, and additional history often reduces performance. These results show that a model can produce plausible intent predictions without robustly integrating temporal evidence throughout a multistep procedure.

In summary, our main contributions are as follows:
\begin{itemize}
    \item We present EgoIntent, a manually constructed benchmark of 3,014 pre-outcome micro-steps for jointly understanding \localcolor{Local Intent}, \proceduralcolor{Procedural Intent}, and \nextcolor{Next-Plan} in egocentric videos of diverse procedural activities.
    
    \item We establish a validated open-ended evaluation protocol through human baselines, answerability and ambiguity analysis, judge--human agreement, and controlled tests of outcome and future-frame leakage.
    
    \item We benchmark 15 MLLMs and diagnose how representative models use temporal order, boundary frames, and historical context, revealing that high intent-prediction scores do not necessarily indicate genuine temporal reasoning across fine-grained procedural steps.
\end{itemize}

\section{Related Work}
\noindent\textbf{Egocentric video datasets and benchmarks.}
Video understanding research has progressed from action-centric perception to broader behavior understanding and, more recently, high-level reasoning and planning. Action-centric datasets such as Charades~\cite{sigurdsson2018charades} support activity understanding in everyday environments, while EPIC-KITCHENS~\cite{damen2018scaling} provides fine-grained egocentric action annotations and HD-EPIC~\cite{perrett2025hd} extends this kitchen domain with detailed multimodal signals. Broader egocentric resources include Ego4D~\cite{grauman2022ego4d}, which covers episodic memory, hand--object interaction, and future prediction, and Ego-Exo4D~\cite{grauman2024ego,fu2025objectrelator,mahdi2025exo2egosyn}, which adds synchronized first- and third-person observations of skilled activities. More recent benchmarks emphasize high-level reasoning and planning~\cite{he2026omnicot,zhang2026panoramic}: EgoSchema~\cite{mangalam2023egoschema} evaluates long-form video question answering; EgoThink~\cite{cheng2024egothink} and VidEgoThink~\cite{cheng2024videgothink} test egocentric reasoning in image and video settings; MM-Ego~\cite{ye2024mm} probes memory for fine-grained visual details; and EgoPlan-Bench~\cite{chen2026egoplan} evaluates next-action planning given observations and an explicit task goal. Together, these resources advance from recognizing visible activity to reasoning over extended behavior, but none jointly tests what an actor is trying to achieve now, why the current step is needed, and what will happen next before the current outcome is revealed. To fill this gap, EgoIntent shifts egocentric intent evaluation from coarse episode- or clip-level goals to manually segmented pre-outcome micro-steps.

\noindent\textbf{Intent reasoning and anticipation.}
Prior work on intent understanding spans intent-oriented question answering and clip- or episode-level goal inference. IntentQA~\cite{li2023intentqa} derives intent-oriented questions from the causal and temporal question types in NExT-QA~\cite{xiao2021next}. In egocentric video, EgoGazeVQA~\cite{peng2025eye} studies spatial, temporal, and causal intent questions with gaze as an additional signal, whereas WAGIBench~\cite{veerabadran2025benchmarking} infers wearable-assistant goals from video, audio, digital, and longitudinal context. These tasks support proactive assistance~\cite{lee2025sensible}, but procedural intent evolves at a finer step level: visually similar actions may serve different functions depending on their surrounding procedure~\cite{seminara2024differentiable,sener2022assembly101}. EgoIntent therefore targets micro-step transitions and jointly evaluates \textit{\localcolor{Local Intent}}, \textit{\proceduralcolor{Procedural Intent}}, and \textit{\nextcolor{Next-Plan}} from visual evidence available strictly before the step outcome becomes visually explicit.

\section{The EgoIntent Benchmark}

\subsection{Overview}

We present EgoIntent, an open-ended benchmark for understanding intent at pre-outcome micro-steps in egocentric procedural activities. Given only the visual evidence available before a step's key outcome, a model must infer what the actor is trying to accomplish now, why this step is needed in the broader procedure, and what action is most likely to happen next. This setting evaluates anticipatory understanding without exposing an explicit task goal, answer options, textual narrations, or future frames.

\begin{figure}[t]
    \centering
    \includegraphics[width=\linewidth]{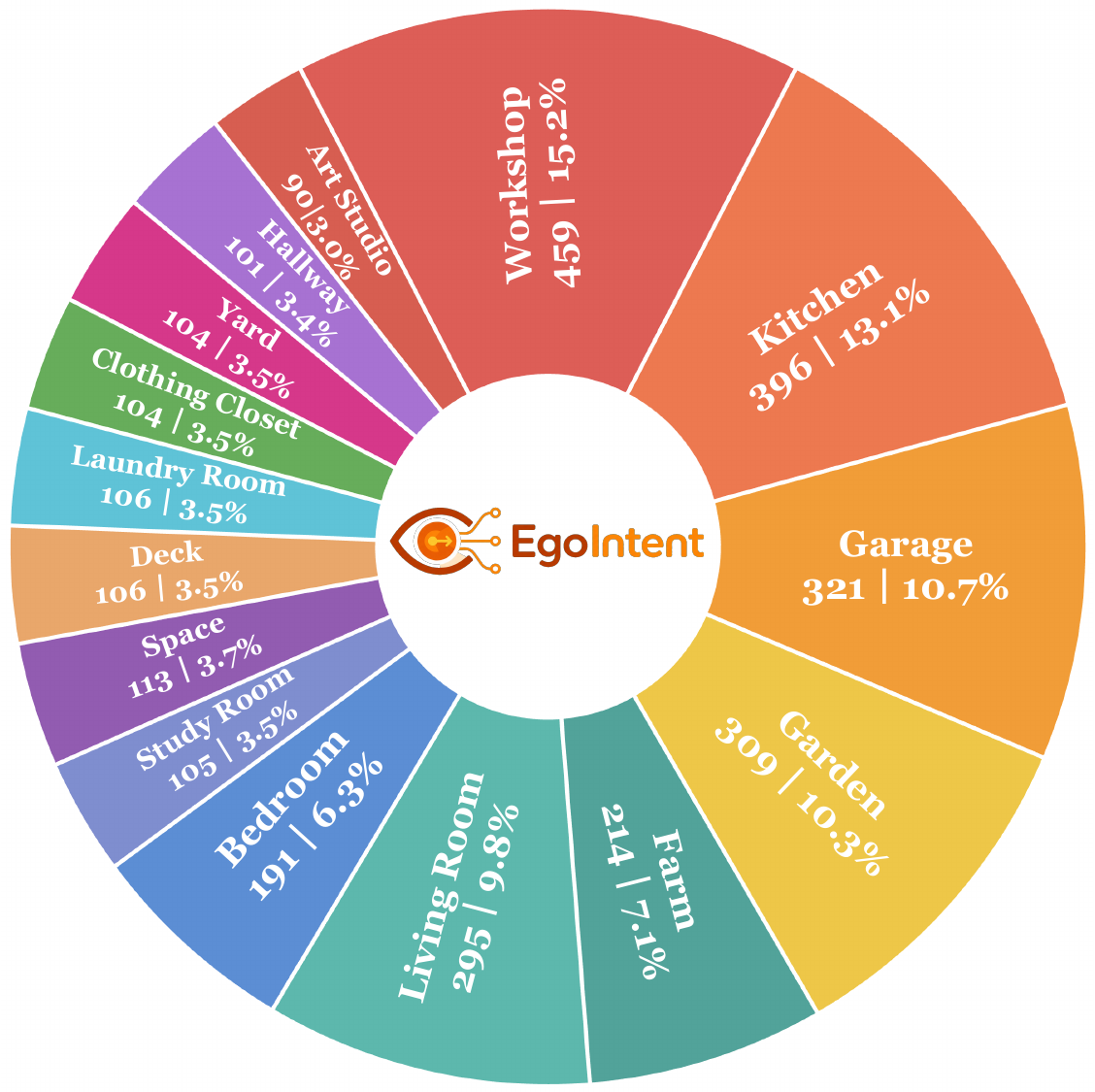}
    \caption{\textbf{Scenario-level composition of EgoIntent.} Each sector is proportional to the number of annotated micro-steps in one of 15 scenarios; labels report the count and share of all 3,014 micro-steps. Scenario size ranges from 90 steps (3.0\%) to 459 (15.2\%), and the four largest scenarios together comprise 49.3\% of the benchmark.}
    \label{fig:pie_chart}
\end{figure}

EgoIntent contains 3,014 manually constructed micro-steps from 32 Ego4D source videos across 15 indoor and outdoor daily-life scenarios. \figref{fig:pie_chart} quantifies the scenario-level composition: scenario size ranges from 90 micro-steps (3.0\%) for Art Studio to 459 (15.2\%) for Workshop. The four largest scenarios---Workshop, Kitchen, Garage, and Garden---jointly account for 49.3\% of the benchmark, while the other 11 contribute 50.7\%; thus, no single scenario dominates the collection. The benchmark covers varied forms of procedural behavior, including cooking, cleaning, organizing, repairing, painting, gardening, and outdoor manual work across diverse environments.

\subsection{Task Definition}

For the $i$-th micro-step, let $V_i=[t_i^{s},t_i^{o}]$ denote the egocentric observation from its manually determined start time $t_i^{s}$ to a pre-outcome cutoff $t_i^{o}$. The key outcome and every subsequent frame are excluded. From $V_i$ alone, a model produces an open-ended triplet $(l_i,p_i,n_i)$:

\noindent\textbf{\localcolor{Local Intent (What)}.}
$l_i$ is the actor's immediate goal within the current step. It must be achievable within that step and should express purpose rather than merely paraphrase a visible hand motion in the observed video.

\noindent\textbf{\proceduralcolor{Procedural Intent (Why)}.}
$p_i$ describes the functional role of the current step in the broader procedure. It connects the immediate goal to procedural progress without collapsing into either the \localcolor{Local Intent} or a generic activity label.

\noindent\textbf{\nextcolor{Next-Plan (Next)}.}
$n_i$ is the immediate action most likely to follow the current step. The reference records the observed continuation; when several continuations are reasonable from the pre-outcome evidence, plausible alternatives are considered separately in our evaluation diagnostics.

The three outputs are deliberately related but non-redundant: \localcolor{Local Intent} is anchored to the current micro-step, \proceduralcolor{Procedural Intent} explains its place in the larger procedure, and \nextcolor{Next-Plan} lies strictly after it. Because valid responses can differ in wording---and the future can admit more than one reasonable continuation---we evaluate semantic agreement rather than exact string matching.

\subsection{Benchmark Construction}

\subsubsection{Source Video Selection and Manual Segmentation.}

We curate raw Ego4D videos that contain multiple coherent procedural transitions and clear hand--object interactions. Annotators inspect each source video and manually divide it into micro-steps, where each step corresponds to one immediate, coherent goal and lasts no more than 10 seconds. They determine the step start, the outcome that marks completion of the immediate goal, and the corresponding temporal boundary.

\subsubsection{Pre-Outcome Observation Cutoff.}

For every micro-step, annotators place an observation cutoff immediately before the key outcome becomes visually explicit. The released model input ends at this cutoff: the outcome-revealing frames and the continuation into later steps are hidden. Consequently, a model must reason from action tendency, hand--object interaction, object state, and procedural context rather than recognize an already completed result. The same boundary rule is applied across all activities, followed by manual review for subtle or gradual outcomes.

\subsubsection{Intent Annotation.}

Annotators write the three intent labels from the video itself. \localcolor{Local Intent} states the immediate goal pursued before the cutoff, while \proceduralcolor{Procedural Intent} states why that goal advances the surrounding procedure. To annotate the observed \nextcolor{Next-Plan}, annotators inspect the immediate continuation after the current step; this continuation is retained only as the target label and is never included in the model input. When the visible prefix supports multiple reasonable futures, additional plausible next actions are recorded for ambiguity analysis instead of forcing every valid future into a single canonical wording.

\subsubsection{Quality Control.}

We conduct multiple rounds of human review over both temporal boundaries and language annotations. Reviewers check that each step contains one coherent immediate goal, respects the 10-second maximum, and ends before its key outcome. They also verify that \localcolor{Local Intent} and \proceduralcolor{Procedural Intent} are visually supportable from the released observation, that their abstraction levels remain distinct, and that \nextcolor{Next-Plan} is temporally subsequent. Cases involving gradual outcomes, preparatory actions, or disagreement between the current goal and the next action are re-examined and adjudicated. This process targets the two principal sources of ambiguity in the benchmark: hindsight leakage at the observation boundary and semantic overlap across the \localcolor{What}, \proceduralcolor{Why}, and \nextcolor{Next} annotations. To further assess annotation quality, our human auditors reviewed a stratified sample of 756 micro-steps. Of these, 724 (95.77\%) satisfied all audit criteria; the remaining 32 were corrected through annotation revision without removing any micro-steps. Full audit protocols, sampling details, agreement statistics, and error analyses are provided in the supplementary material.

\section{Experiments}

\begin{table*}[!t]
\centering
\small
\caption{\textbf{Full-benchmark performance of 15 MLLMs on EgoIntent ($N=3{,}014$).} The left block reports reference-based semantic scores, whereas the right block reports complementary reference-free diagnostics. The two blocks use different Judges and are not directly comparable. Overall averages \localcolor{Local}, \proceduralcolor{Procedural}, and \nextcolor{Next}. Bold marks within-group column maxima.}
\label{tab:main_results}
\setlength{\tabcolsep}{2.2pt}
\renewcommand{\arraystretch}{1.08}
\newcommand{\mainresultrow}{\rule[-2ex]{0pt}{1.5ex}}
\newlength{\mainmodelwidth}
\newlength{\mainscorewidth}
\setlength{\mainmodelwidth}{0.36\textwidth}
\setlength{\mainscorewidth}{0.064\textwidth}
\newcolumntype{Y}{>{\raggedright\arraybackslash}m{\mainmodelwidth}}
\newcolumntype{Z}{>{\centering\arraybackslash}m{\mainscorewidth}}
\newcommand{\bestscore}[1]{\textbf{#1}}
\newcommand{\worstscore}[1]{#1}
\begin{tabular*}{\textwidth}{@{\extracolsep{\fill}}Y|*{4}{Z}|*{4}{Z}@{}}
\toprule
\textbf{\textit{Model}} &
\multicolumn{4}{c|}{\textbf{\textit{Reference-Based Evaluation}}} &
\multicolumn{4}{c}{\textbf{\textit{Reference-Free Diagnostics}}} \\
\cmidrule(lr){2-5}\cmidrule(l){6-9}
& \textit{\localcolor{Local}} & \textit{\proceduralcolor{Procedural}} & \textit{\nextcolor{Next}} & \textit{Overall}
& \textit{GF} & \textit{HIC} & \textit{TPC} & \textit{NPF} \\
\midrule
\rowcolor{tablegroup}
\multicolumn{9}{c}{\textit{Closed-Source MLLMs}} \\
\midrule
\mainresultrow Doubao-Seed-2.1-Turbo~\cite{seed2025doubao}
& 57.98
& \bestscore{63.62}
& \bestscore{43.78}
& \bestscore{55.13}
& \bestscore{74.47}
& \bestscore{85.19}
& \bestscore{78.86}
& \bestscore{76.75} \\
\mainresultrow Qwen3.5-Plus~\cite{qwen3.5}
& \bestscore{57.99}
& 57.76 & 42.38 & 52.71
& 73.30 & 81.62 & 76.97 & 75.52 \\
\mainresultrow Gemini-3.5-Flash~\cite{gemini35flash}
& 43.34 & 48.31 & 32.72 & 41.45
& 58.79 & 70.24 & 62.66 & 60.66 \\
\mainresultrow\smash{\makecell[l]{Amazon-Nova-2-Lite-V1\\\cite{intelligence2025amazon}}}
& 37.98
& 42.96
& 22.84
& 34.59
& 55.05
& 63.75
& 56.77
& 54.93 \\
\midrule
\rowcolor{tablegroup}
\multicolumn{9}{c}{\textit{Open-Weight MLLMs}} \\
\midrule
\mainresultrow Qwen3-VL-32B-Instruct~\cite{bai2025qwen3}
& \textbf{45.14}
& \textbf{52.81}
& \textbf{30.44}
& \textbf{42.80}
& \textbf{69.84}
& \textbf{81.32}
& \textbf{75.12}
& \textbf{72.68} \\
\mainresultrow Qwen3-VL-8B-Instruct~\cite{bai2025qwen3}
& 40.86 & 45.03 & 26.59 & 37.49
& 61.18 & 71.50 & 64.75 & 62.34 \\
\mainresultrow Qwen2.5-VL-7B-Instruct~\cite{qwen2.5vl}
& 35.41 & 42.45 & 22.53 & 33.46
& 52.20 & 64.98 & 53.85 & 51.87 \\
\mainresultrow Qwen2-VL-7B-Instruct~\cite{wang2024qwen2}
& 37.42 & 39.57 & 22.53 & 33.18
& 51.26 & 60.19 & 48.67 & 46.81 \\
\mainresultrow Molmo2-8B~\cite{clark2026molmo2}
& 35.10 & 41.01 & 20.55 & 32.22
& 52.50 & 62.14 & 54.01 & 51.83 \\
\mainresultrow InternVL3-8B~\cite{zhu2025internvl3}
& 34.25 & 40.16 & 18.51 & 30.97
& 47.46 & 60.94 & 49.14 & 46.58 \\
\mainresultrow Molmo2-O-7B~\cite{clark2026molmo2}
& 32.36 & 37.32 & 19.02 & 29.57
& 49.29 & 57.77 & 48.42 & 46.33 \\
\mainresultrow LLaVA-Video-7B-Qwen2~\cite{zhang2024llava}
& 33.14 & 33.42 & 15.65 & 27.41
& 49.51 & 54.29 & 47.49 & 45.48 \\
\mainresultrow Kimi-VL-A3B-Thinking-2506~\cite{team2025kimi}
& 27.11 & 33.19 & 14.98 & 25.09
& 40.15 & 50.94 & 39.35 & 37.10 \\
\mainresultrow InternVL2-8B~\cite{chen2024far}
& 21.60 & 28.25 & 12.58 & 20.81
& 33.12 & 43.32 & 34.72 & 32.55 \\
\mainresultrow LLaVA-NeXT-Video-7B~\cite{zhang2024llavanextvideo}
& \worstscore{13.36}
& \worstscore{15.51}
& \worstscore{5.76}
& \worstscore{11.54}
& \worstscore{24.37}
& \worstscore{27.89}
& \worstscore{20.59}
& \worstscore{19.53} \\
\bottomrule
\end{tabular*}
\end{table*}

\subsection{Evaluation Protocol}

\noindent\textbf{Models and inputs.}
We evaluate 15 representative MLLMs, including four closed-source systems and eleven open-weight models. Each model receives only the pre-outcome visual observation and a common prompt requesting three fixed fields: \localcolor{Local Intent}, \proceduralcolor{Procedural Intent}, and \nextcolor{Next-Plan}. We exclude activity names, scene labels, narrations, reference annotations, and all future frames. The full-benchmark leaderboard uses documented model-specific media interfaces and should therefore be read as a comparison under those protocols, rather than as a strictly controlled comparison between heterogeneous native-video and multi-image interfaces. All controlled studies use the same prompt, temperature 0, and a fixed budget of 16 uniformly sampled frames.

\noindent\textbf{Reference-based scoring.}
Because EgoIntent requires open-ended generation, exact string matching would penalize semantically equivalent answers. DeepSeek-V4-Flash~\cite{deepseekai2026deepseekv4} therefore scores the semantic agreement between each prediction and its reference on a 0--100 scale for \localcolor{Local}, \proceduralcolor{Procedural}, and \nextcolor{Next} separately; Overall is their arithmetic mean. Missing predictions receive zero. For \nextcolor{Next-Plan}, the main score measures agreement with the observed continuation, while human and reference-free analyses separately quantify plausible alternatives for the same visual observation.

\noindent\textbf{Reference-free diagnostics.}
We additionally use a Kimi-K2.5 video Judge~\cite{kimi2026k25} that sees the observation video and an anonymized prediction, but not the reference answer. It scores four complementary properties on a 0--100 scale: \textit{Visual Grounding Faithfulness} (GF), whether claims are supported by visible evidence; \textit{Hierarchical Intent Consistency} (HIC), whether \localcolor{Local} and \proceduralcolor{Procedural} form a valid goal hierarchy; \textit{Temporal Progression Consistency} (TPC), whether current and future predictions follow a coherent order; and \textit{\nextcolor{Next-Plan} Feasibility} (NPF), whether the proposed next action is immediately executable from the observed state. These diagnostics explain prediction quality and do not replace the reference-based ranking. Detailed metric definitions and scoring procedures appear in the supplementary material.

\noindent\textbf{Statistical protocol.}
Controlled comparisons are paired within samples. We compute 95\% confidence intervals with 2,000 bootstrap resamples clustered by source video, and call a paired difference significant when its interval excludes zero. We always report the effective sample count and analyze only the common valid prediction set for the conditions being compared in each controlled analysis.

\subsection{Main Benchmark Results}

\tabref{tab:main_results} reports reference-based results on all 3,014 steps. Doubao-Seed-2.1-Turbo performs best overall at 55.13, while Qwen3-VL-32B-Instruct is the strongest open-weight model at 42.80. \nextcolor{Next-Plan} is the lowest-scoring dimension for every model, despite large differences in model scale and architecture. Thus, predicting the observed continuation remains harder than identifying either the immediate goal or its procedural role within the surrounding activity.

\begin{table*}[!t]
\centering
\small
\caption{\textbf{Pre-outcome leakage analysis on 508 human-validated paired samples.} Scores average four models. Panel A compares observation conditions; Panel B reports target-versus-nontarget gains (L/P/N: \localcolor{Local}/\proceduralcolor{Procedural}/\nextcolor{Next}). Bootstrap confidence intervals are clustered by source video.}
\label{tab:leakage_summary}
\renewcommand{\arraystretch}{1.0}
\begin{minipage}[t]{0.49\textwidth}
\vspace{0pt}
\centering
\setlength{\tabcolsep}{3pt}
\begin{tabular*}{\linewidth}{@{\extracolsep{\fill}}lcccc@{}}
\toprule
\rowcolor{tablegroup}
\multicolumn{5}{c}{\textit{A. Four observation conditions}} \\
\midrule
Condition & \localcolor{Local} & \proceduralcolor{Proc.} & \nextcolor{Next} & Overall \\
\midrule
Early & 45.59 & 51.86 & 35.55 & 44.33 \\
Official & 49.80 & 55.21 & 39.30 & 48.10 \\
Outcome Visible & \textbf{57.61} & 57.09 & 41.33 & 52.01 \\
\nextcolor{Next} Visible & 55.23 & \textbf{57.47} & \textbf{52.47} & \textbf{55.06} \\
\bottomrule
\end{tabular*}
\end{minipage}
\hfill
\begin{minipage}[t]{0.49\textwidth}
\vspace{0pt}
\centering
\setlength{\tabcolsep}{4pt}
\begin{tabular*}{\linewidth}{@{\extracolsep{\fill}}lcc@{}}
\toprule
\rowcolor{tablegroup}
\multicolumn{3}{c}{\textit{B. Dimension-specific gain tests}} \\
\midrule
Contrast & $\Delta$ & 95\% CI \\
\midrule
Outcome: L$-$P & $+5.93$ & $[+3.67,+8.14]$ \\
Outcome: L$-$N & $+5.78$ & $[+3.31,+8.17]$ \\
Future: N$-$L & $+7.75$ & $[+5.31,+10.08]$ \\
Future: N$-$P & $+10.91$ & $[+8.27,+13.63]$ \\
\bottomrule
\end{tabular*}
\end{minipage}
\end{table*}

\begin{table*}[!t]
\centering
\small
\caption{\textbf{Paired shortcut diagnostics on the 640-step subset.} Temporal, Motion, and Visual Evidence Gain denote Ordered$-$Shuffled, Ordered$-$Last Frame, and Ordered$-$Scene Only, respectively. Values include source-video-clustered 95\% confidence intervals; bold denotes intervals excluding zero (effective paired $N=639/639/598$).}
\label{tab:temporal_shortcuts}
\setlength{\tabcolsep}{4pt}
\renewcommand{\arraystretch}{0.9}
\begin{tabular*}{\textwidth}{@{\extracolsep{\fill}}lcccccc@{}}
\toprule
Model
& \multicolumn{2}{c}{Temporal}
& \multicolumn{2}{c}{Motion}
& \multicolumn{2}{c}{Visual Evidence} \\
\cmidrule(lr){2-3}\cmidrule(lr){4-5}\cmidrule(l){6-7}
& Gain & 95\% CI
& Gain & 95\% CI
& Gain & 95\% CI \\
\midrule
Doubao-Seed-2.1-Turbo
& $-0.31$ & $[-2.34,+1.69]$
& $-1.36$ & $[-4.53,+2.00]$
& \textbf{$+40.96$} & $[+36.63,+44.66]$ \\
Qwen3-VL-32B
& $+0.56$ & $[-1.28,+2.46]$
& \textbf{$-7.41$} & $[-9.45,-5.28]$
& \textbf{$+33.52$} & $[+27.60,+39.50]$ \\
Molmo2-8B
& \textbf{$+2.38$} & $[+0.75,+4.05]$
& \textbf{$-2.70$} & $[-4.95,-0.13]$
& \textbf{$+34.73$} & $[+29.71,+39.03]$ \\
Qwen3.5-Plus
& $+1.43$ & $[-0.68,+3.39]$
& \textbf{$-3.08$} & $[-5.44,-0.62]$
& \textbf{$+35.75$} & $[+29.94,+40.80]$ \\
\bottomrule
\end{tabular*}
\end{table*}

\figref{fig:scores} juxtaposes the OS ranking with the four reference-free diagnostics. The three highest-OS models also lie near the upper end of the diagnostic curves, and Doubao leads all five displayed measures. Across models, HIC exceeds GF by 9.57 points on average, while NPF exceeds reference-based \nextcolor{Next} by 28.67 points. These cross-metric gaps are descriptive: HIC and GF reflect distinct rubric dimensions, whereas NPF and reference-based \nextcolor{Next} additionally use different Judges and evaluation targets. For example, Qwen3-VL-32B scores 30.44 on observed \nextcolor{Next} matching and 72.68 on feasibility, suggesting that some predictions that do not match the observed continuation may nevertheless represent feasible next actions.

\subsection{Human Baseline and Task Validity}

We evaluate three independent human participants on a balanced human-evaluation subset of 320 steps covering all scenarios and source videos. The same subset is used for the two model baselines in \tabref{tab:human_validity}; these numbers must therefore not be compared directly with the full-benchmark scores in \tabref{tab:main_results}. Humans reach 70.7 Overall, exceeding the strongest closed and open baselines on this subset by 16.9 and 27.5 points, respectively. The paired 95\% CIs for these gaps are $[13.8,20.0]$ and $[24.0,31.1]$ in magnitude.

\begin{table}[!t]
\centering
\small
\caption{\textbf{Human performance and task validity on the balanced 320-step subset.} Panel A compares the three-human mean with the strongest closed- and open-model baselines on identical steps. Panel B reports answerability (Ans.), chance-corrected Fleiss $\kappa$, mean confidence (Conf.; 1--5), and pairwise semantic agreement (Sem.; 0--1); higher is better.}
\label{tab:human_validity}
\setlength{\tabcolsep}{1.5pt}
\renewcommand{\arraystretch}{0.9}
\begin{tabular*}{\columnwidth}{@{\extracolsep{\fill}}lccccc@{}}
\toprule
\rowcolor{tablegroup}
\multicolumn{6}{c}{\textit{A. Human--model performance}} \\
\midrule
Evaluator & \localcolor{Local} & \proceduralcolor{Proc.} & \nextcolor{Next} & Overall & Gap \\
\midrule
Human mean & 78.3 & 73.1 & 60.8 & \textbf{70.7} & -- \\
Doubao & 59.7 & 56.2 & 45.5 & 53.8 & $-16.9$ \\
Qwen3-VL-32B & 48.8 & 44.3 & 36.5 & 43.2 & $-27.5$ \\
\bottomrule
\end{tabular*}
\smallskip

\setlength{\tabcolsep}{2pt}
\begin{tabular*}{\columnwidth}{@{\extracolsep{\fill}}lcccc@{}}
\toprule
\rowcolor{tablegroup}
\multicolumn{5}{c}{\textit{B. Answerability and agreement}} \\
\midrule
Dimension & Ans. & Fleiss $\kappa$ & Conf. & Sem. \\
\midrule
\localcolor{Local} & 91.6\% & 0.68 & 4.18 & 0.79 \\
\proceduralcolor{Procedural} & 84.9\% & 0.58 & 3.86 & 0.70 \\
\nextcolor{Next} & 70.8\% & 0.44 & 3.39 & 0.56 \\
\bottomrule
\end{tabular*}
\end{table}

Answerability falls monotonically from \localcolor{Local} (91.6\%) to \proceduralcolor{Procedural} (84.9\%) and \nextcolor{Next} (70.8\%), as do inter-rater agreement and confidence. For \nextcolor{Next-Plan}, humans identify multiple reasonable answers in 32.8\% of samples and judge another 29.1\% not reliably predictable from the visible prefix. Thus, \localcolor{Local} and \proceduralcolor{Procedural} are usually answerable, whereas \nextcolor{Next} must be assessed jointly using observed-match scores, feasibility, and plausible-future analyses for each sample.

\subsection{Does Pre-Outcome Truncation Prevent Leakage?}

We test four visual conditions on the same diagnostic samples: \textit{Early} ends 0.5 seconds before the official cutoff; \textit{Official} uses the benchmark input; \textit{Outcome Visible} extends the observation to expose the current result; and \textit{\nextcolor{Next} Visible} exposes early evidence from the following micro-step. Human boundary auditing finds the official pre-outcome boundary valid for 603 of 640 inspected samples (94.2\%). The strictly paired analysis uses the 508 samples for which all four conditions are human-valid and all four evaluated models have complete predictions and Judge scores.

Official improves Overall over Early by 3.77 points, showing that the selected prefix retains useful action-development cues. Yet revealing the outcome adds another 3.91 points, and revealing the following step adds 6.96. Crucially, the gains align with the information injected: Outcome Visible improves \localcolor{Local} by 7.81 points on average, whereas \nextcolor{Next} Visible improves \nextcolor{Next} by 13.17. Both interventions benefit their target dimension significantly more than the non-target dimensions (\tabrefpanel{tab:leakage_summary}{B}). The official cutoff therefore occupies an informative middle ground: it is not arbitrarily early, but it withholds the visual evidence that would directly disclose the current result or following action.

\subsection{What Evidence Do Models Actually Use?}

To separate temporal reasoning from static shortcuts, we construct a 640-step diagnostic subset and compare ordered video, the identical frames in shuffled order, the last boundary frame alone, and a scene-name-only condition. Ordered versus shuffled frames isolates temporal-order gain; ordered versus last frame measures the value of multi-frame dynamics over the boundary state; and ordered versus scene name measures the value of concrete visual evidence.

The result is counterintuitive (\tabref{tab:temporal_shortcuts}). Only Molmo2-8B obtains a significantly positive temporal-order gain: $+2.38$ with a 95\% CI of $[+0.75,+4.05]$. For Doubao, Qwen3-VL-32B, and Qwen3.5-Plus, the intervals include zero. Moreover, the last frame significantly outperforms the full ordered clip for Qwen3-VL-32B ($-7.41$), Molmo2-8B ($-2.70$), and Qwen3.5-Plus ($-3.08$), where negative values denote ordered-minus-last. This is not mere scene guessing: ordered video exceeds scene-only input by 33.52--40.96 points across all four models. Current models use concrete visual evidence, but much of their predictive power comes from the static state near the decision boundary rather than robust use of temporal order. Consequently, a high benchmark score alone should not be read as evidence of strong temporal reasoning.

\subsection{Does More History Help?}

 We compare the current step alone with 5 seconds of history, 15 seconds of history, and the preceding annotated step. Every condition ends at the same official cutoff and receives the same 16-frame budget.

\begin{figure}[!t]
\centering
\includegraphics[width=\columnwidth]{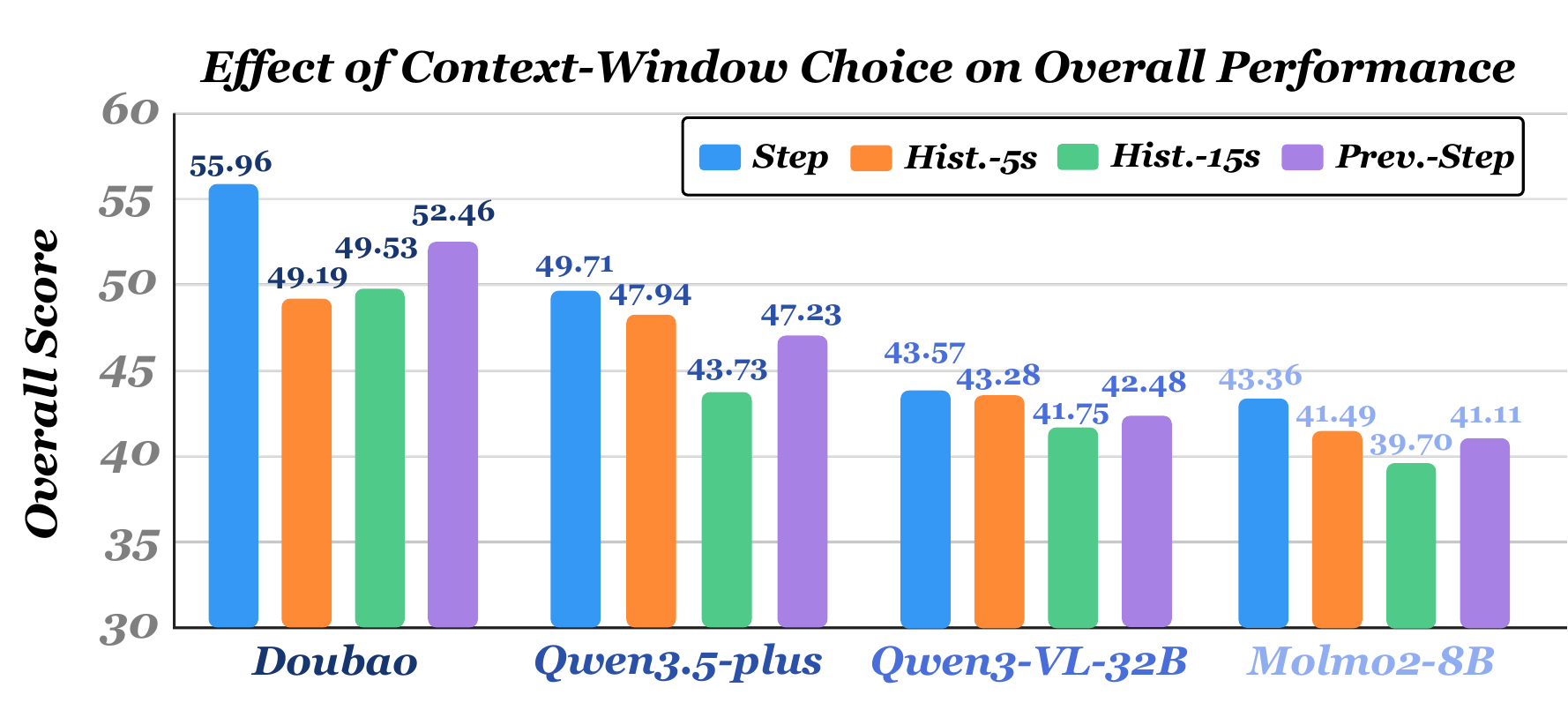}
\caption{\textbf{Context-window effects on the 640-step diagnostic subset.} Overall scores for four MLLMs under Step, Hist.-5s, Hist.-15s, and Prev.-Step inputs with a shared observation endpoint.}
\label{fig:context_history}
\end{figure}

As shown in \figref{fig:context_history}, Step-only achieves the highest Overall for all four models. On paired common samples, 15 seconds of history significantly reduces Overall relative to Step-only for Doubao ($-6.36$), Molmo2-8B ($-3.60$), and Qwen3.5-Plus ($-5.92$); Qwen3-VL-32B also decreases by 1.75 points, although its interval includes zero. The loss is largest for the shortest steps: averaged across models, the 15-second condition drops by 10.49 points for $<1$ second and 6.49 points for 1--2 seconds.

This result does not imply that procedural history is intrinsically unhelpful. Under a fixed frame budget, adding history reduces the sampling density of the current micro-step. The experiment instead shows that current models do not reliably select and integrate useful history when earlier context competes with the visually salient boundary state.

\FloatBarrier
\subsection{Judge Validation}

We validate automatic scoring using 300 anonymized prediction--reference triplets from 150 unique steps, each rated by three humans. \tabref{tab:judge_validation} reports human inter-rater reliability and the agreement of DeepSeek-V4-Flash and GLM-5.1~\cite{glm5team2026glm5} with human consensus scores across the evaluated triplets.

\begin{table}[!t]
\centering
\small
\caption{\textbf{Human and automatic-Judge validation on 150 steps (300 triplets; three raters each).} Panel A reports ICC$(2,k)$ and Krippendorff's $\alpha$; Panel B reports Spearman $\rho$, MAE, and Judge-minus-human bias against human consensus. Overall is computed per triplet.}
\label{tab:judge_validation}
\setlength{\tabcolsep}{2.5pt}
\renewcommand{\arraystretch}{0.9}
\begin{tabular*}{\columnwidth}{@{\extracolsep{\fill}}lccc@{}}
\toprule
\rowcolor{tablegroup}
\multicolumn{4}{c}{\textit{A. Human inter-rater reliability}} \\
\midrule
Dimension & ICC$(2,k)$ & 95\% CI & $\alpha$ \\
\midrule
\localcolor{Local} & 0.87 & $[0.84,0.90]$ & 0.78 \\
\proceduralcolor{Procedural} & 0.83 & $[0.79,0.87]$ & 0.73 \\
\nextcolor{Next} & 0.78 & $[0.72,0.83]$ & 0.69 \\
Mean & \textbf{0.83} & -- & \textbf{0.73} \\
\bottomrule
\end{tabular*}
\smallskip

\setlength{\tabcolsep}{1.5pt}
\begin{tabular*}{\columnwidth}{@{\extracolsep{\fill}}llccc@{}}
\toprule
\rowcolor{tablegroup}
\multicolumn{5}{c}{\textit{B. Automatic Judge vs.\ human consensus}} \\
\midrule
Judge & Dim. & $\rho$ & MAE & Bias \\
\midrule
DeepSeek-V4 & \localcolor{Local} & 0.76 & 11.1 & $-1.0$ \\
& \proceduralcolor{Procedural} & 0.72 & 12.5 & $-1.1$ \\
& \nextcolor{Next} & 0.67 & 15.6 & $+1.0$ \\
& Overall & \textbf{0.78} & \textbf{9.8} & $\mathbf{-0.4}$ \\
\midrule
GLM-5.1 & \localcolor{Local} & 0.71 & 12.0 & $-0.8$ \\
& \proceduralcolor{Procedural} & 0.68 & 13.0 & $-1.5$ \\
& \nextcolor{Next} & 0.62 & 15.1 & $+0.3$ \\
& Overall & \textbf{0.73} & \textbf{11.4} & $\mathbf{-0.7}$ \\
\bottomrule
\end{tabular*}
\end{table}

Mean human inter-rater reliability across the three dimensions is ICC$(2,k)=0.83$. For Overall, DeepSeek-V4-Flash attains Spearman $\rho=0.78$ with the human consensus, a mean absolute error of 9.8, and a mean bias of $-0.4$. Although GLM-5.1 has lower sample-level agreement (Spearman $\rho=0.73$, MAE 11.4), it exhibits the same dimension-wise trend and exactly preserves the human Overall ranking of the six evaluated models. This consistency across Judges supports the robustness of our Judge-based evaluation protocol. For the primary Judge, \nextcolor{Next-Plan} has the weakest agreement and largest error (Spearman $\rho=0.67$, MAE 15.6), consistent with its greater ambiguity. The primary Judge is useful for aggregate evaluation, but individual ambiguous futures still require human judgment.\textbf{All experimental details can be found in the supplementary materials.}
\FloatBarrier

\section{Conclusion}
We introduced \textbf{EgoIntent}, a manually constructed benchmark of 3,014 pre-outcome micro-steps from 32 egocentric videos across 15 daily-life scenarios. EgoIntent separates anticipatory understanding into \localcolor{Local Intent}, \proceduralcolor{Procedural Intent}, and \nextcolor{Next-Plan}. Across 15 evaluated MLLMs, Doubao-Seed-2.1-Turbo achieves the highest Overall score of 55.13, while Qwen3-VL-32B-Instruct is the strongest open-weight model at 42.80.

Humans reach 70.7 Overall on the balanced subset, exceeding the strongest model by 16.9 points. Answerability, agreement, and confidence decrease from \localcolor{Local} to \proceduralcolor{Procedural} to \nextcolor{Next}; multiple futures are reasonable for 32.8\% of samples, while another 29.1\% are not reliably predictable. Boundary auditing and controlled leakage tests further confirm that the official cutoff preserves useful action-development cues while withholding the current outcome and following action.

Only one of four representative models benefits significantly from correct temporal order, while the last boundary frame outperforms the full ordered clip for three models and added history often reduces performance. Current MLLMs therefore rely disproportionately on static boundary states rather than integrating motion, order, and procedural history. Judge validation supports aggregate automatic evaluation, although ambiguous futures still require human review.

\bibliography{references}

\clearpage
\appendix
\onecolumn
\setcounter{secnumdepth}{3}
\makeatletter
\renewcommand\subsubsection{%
  \@startsection{subsubsection}{3}{\z@}%
    {-6pt plus -2pt minus -1pt}{-1em}{\normalsize\bf}%
}
\makeatother
\setlength{\emergencystretch}{3em}
\newenvironment{promptblock}{%
  \par\hrule height 1.0pt\smallskip
  \begingroup\ttfamily\raggedright\sloppy
}{%
  \par\endgroup\smallskip\hrule height 1.0pt\medskip
}
\lstdefinestyle{promptcode}{%
  basicstyle=\footnotesize\ttfamily,
  numbers=none,
  frame=single,
  framerule=0.8pt,
  framesep=4pt,
  aboveskip=4pt,
  belowskip=8pt,
  showstringspaces=false,
  breaklines=true,
  breakatwhitespace=false,
  columns=fullflexible,
  keepspaces=true,
  tabsize=2
}
\newcommand{\promptheading}[1]{%
  \par\medskip
  \noindent\textbf{#1}\par\nobreak\smallskip
}
\newcommand{\checksumline}[2]{%
  \par\smallskip
  \begingroup\raggedright
  \noindent\textbf{#1}\par\nobreak
  \noindent\texttt{\detokenize{#2}}\par
  \endgroup
}
\newcommand{\sceneexample}[4]{%
  \par\noindent
  \begin{minipage}{\textwidth}
  \centering
  \includegraphics[width=0.94\linewidth]{#1}
  \captionsetup{skip=2pt}
  \captionof{figure}{\textbf{#2.} #3}
  \label{#4}
  \end{minipage}
  \par\vspace{4pt}
}
\newenvironment{modeltable}{%
  \par\noindent\begin{minipage}{\textwidth}\centering
}{%
  \par\end{minipage}\par
}
\section*{EgoIntent Supplementary Material}

\section{Data Visualization}

\subsection{Dataset composition}

Figure~\ref{fig:app_event_distribution} provides an event-level view of the EgoIntent benchmark. The 32 bars correspond to the 32 source videos/events, and the colored group labels indicate the 15 scenes represented in the dataset. This view complements the scenario-level composition in Figure~\ref{fig:pie_chart} by showing how the 3,014 annotated micro-steps are distributed within and across scenes.

The event-level breakdown also exposes variation hidden by scene totals. Scenes represented by several source events contribute different numbers of steps from each recording, while single-event scenes remain directly identifiable. Showing both granularities makes the benchmark composition transparent and separates broad scenario coverage from concentration within individual recordings.

\begin{center}
\includegraphics[width=\textwidth]{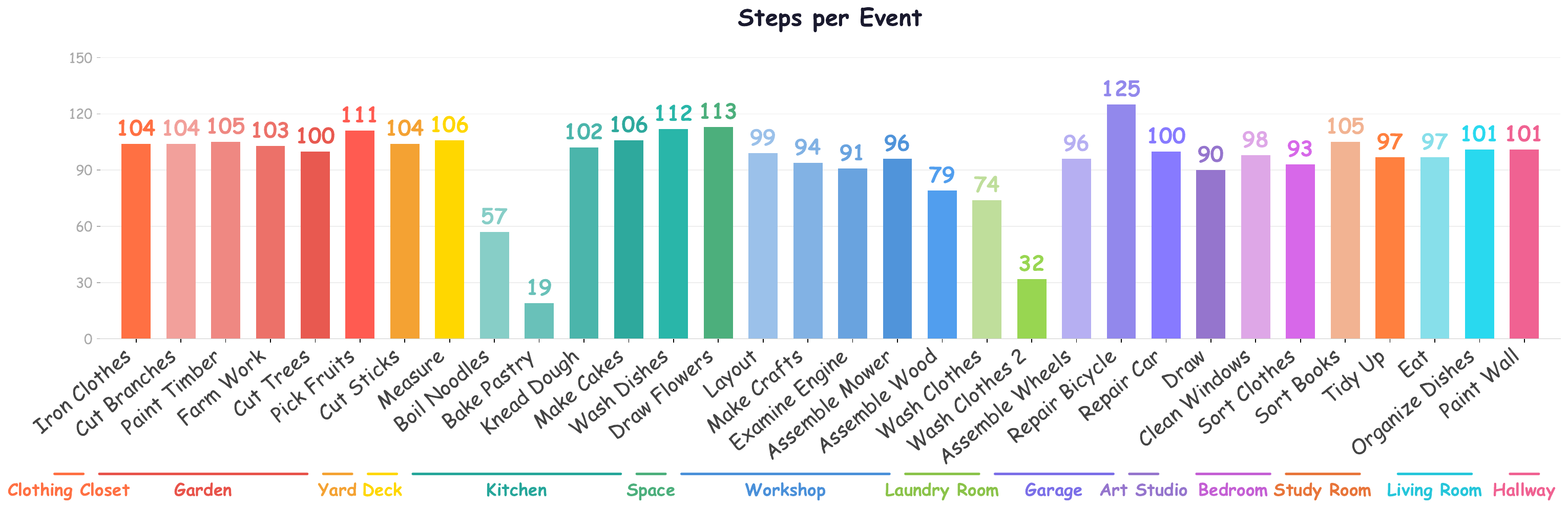}
\captionof{figure}{\textbf{Event-level composition of EgoIntent.} Each bar shows the number of annotated micro-steps from one source video/event. Colors and group labels organize the 32 events into the 15 EgoIntent scenes, and the values above the bars report the corresponding micro-step counts.}
\label{fig:app_event_distribution}
\end{center}

\subsection{Qualitative examples across scenes}

\raggedbottom
Figures~\ref{fig:app_scene_art_studio}--\ref{fig:app_scene_yard} show representative micro-steps from all 15 EgoIntent scenes. The left filmstrip contains the released pre-outcome observation, while the faded frames on the right show the withheld continuation. The annotations distinguish \localcolor{Local Intent}, \proceduralcolor{Procedural Intent}, \nextcolor{Next Plan}, and plausible alternatives.

These panels are compact task visualizations rather than complete video summaries. Each observed prefix stops while the current manipulation is still in progress, providing evidence for an immediate goal without revealing its outcome. The faded frames show the continuation realized in the source video, whereas the plausible-plan list records other compatible futures. Across scenes, the examples vary in objects, hand-object interactions, viewpoints, and motion patterns. \nextcolor{Next-plan} prediction must therefore connect visible manipulation to both local purpose and broader procedure while preserving uncertainty among feasible continuations.

\sceneexample{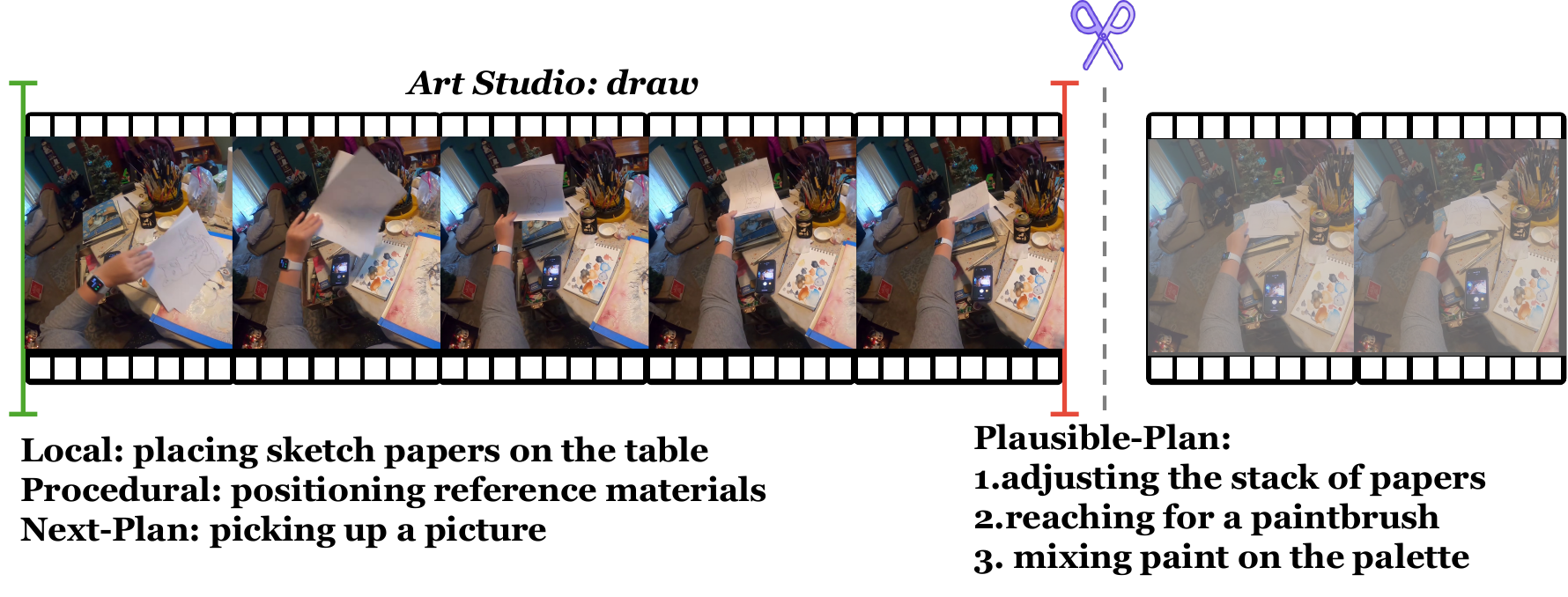}{Art Studio}{Positioning sketch papers as visual references.}{fig:app_scene_art_studio}

\sceneexample{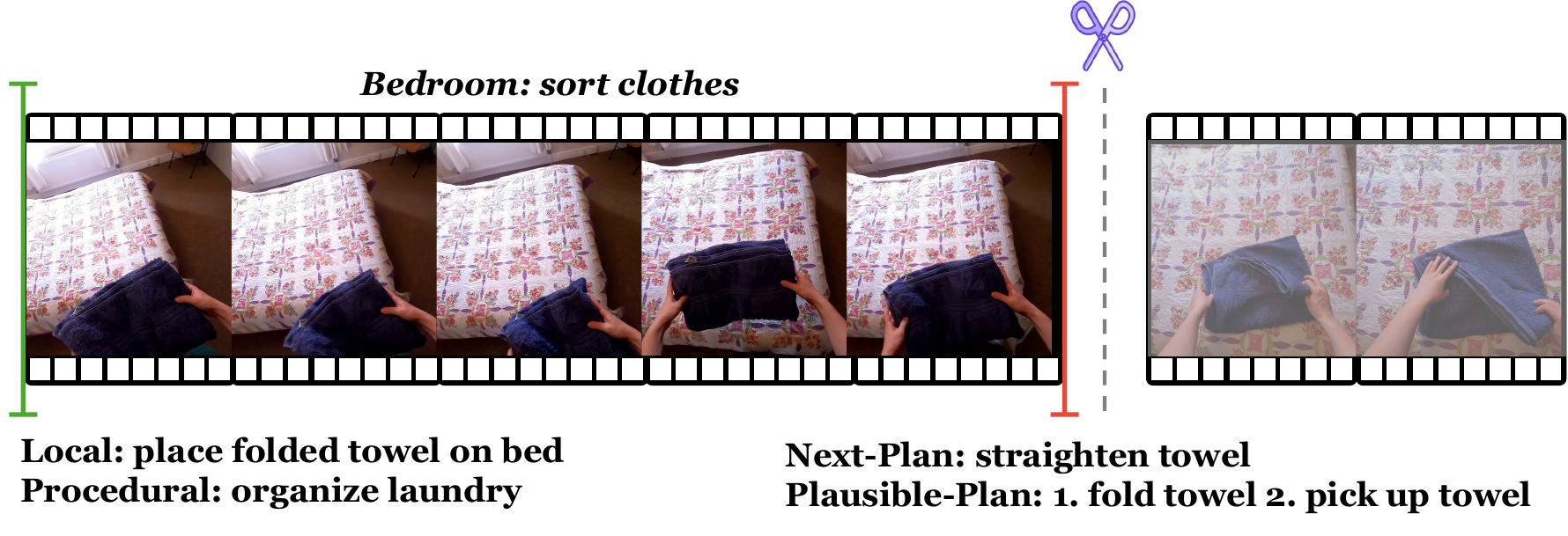}{Bedroom}{Placing a folded towel on the bed.}{fig:app_scene_bedroom}

\sceneexample{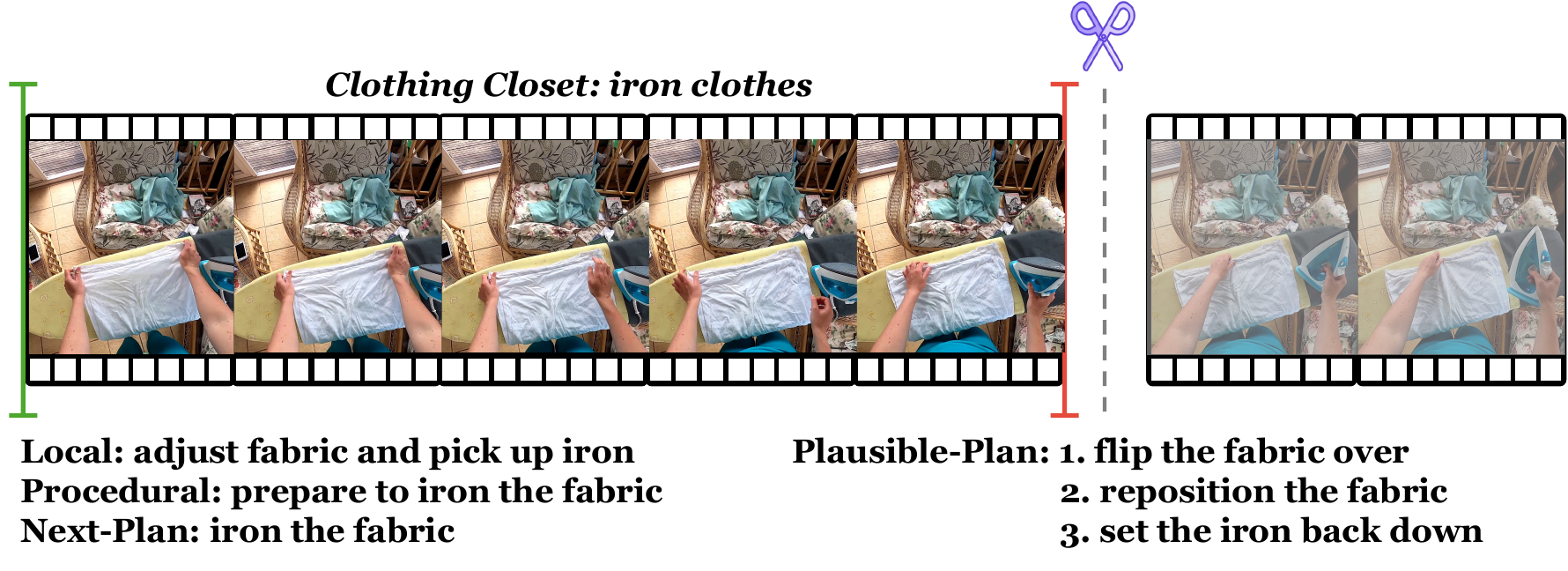}{Clothing Closet}{Preparing fabric and an iron for pressing.}{fig:app_scene_clothing_closet}

\sceneexample{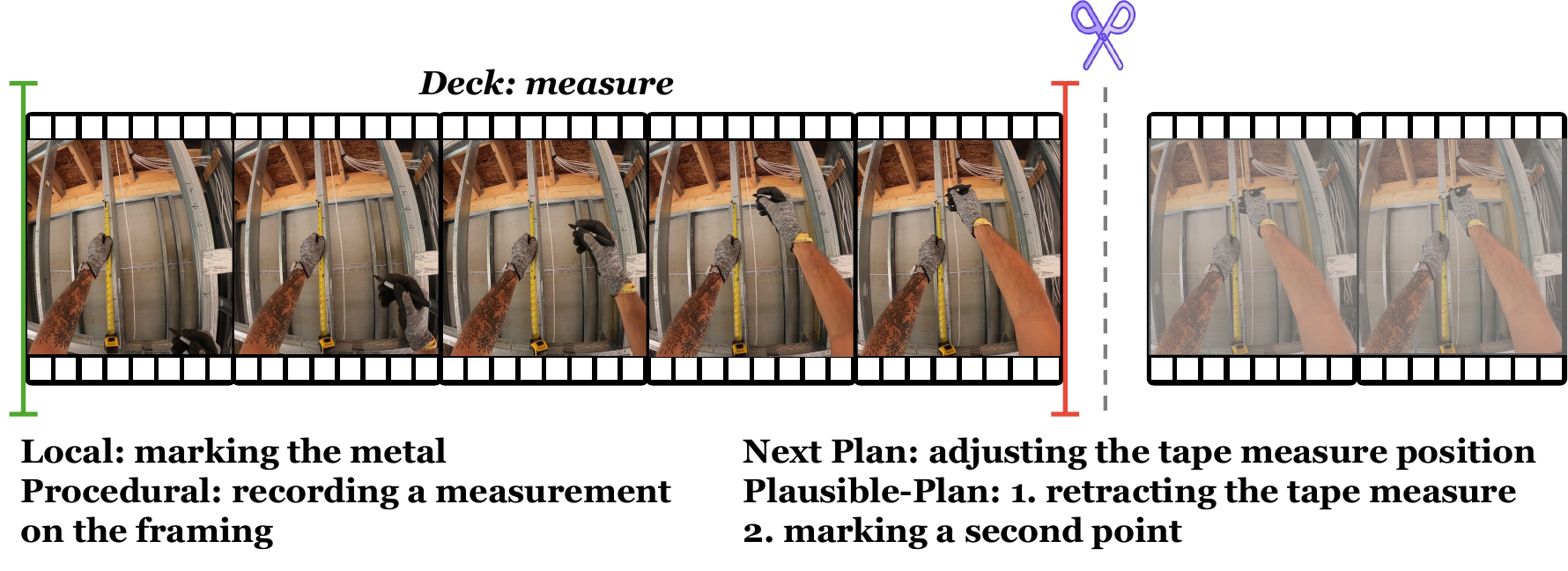}{Deck}{Marking a measured point on the framing.}{fig:app_scene_deck}

\sceneexample{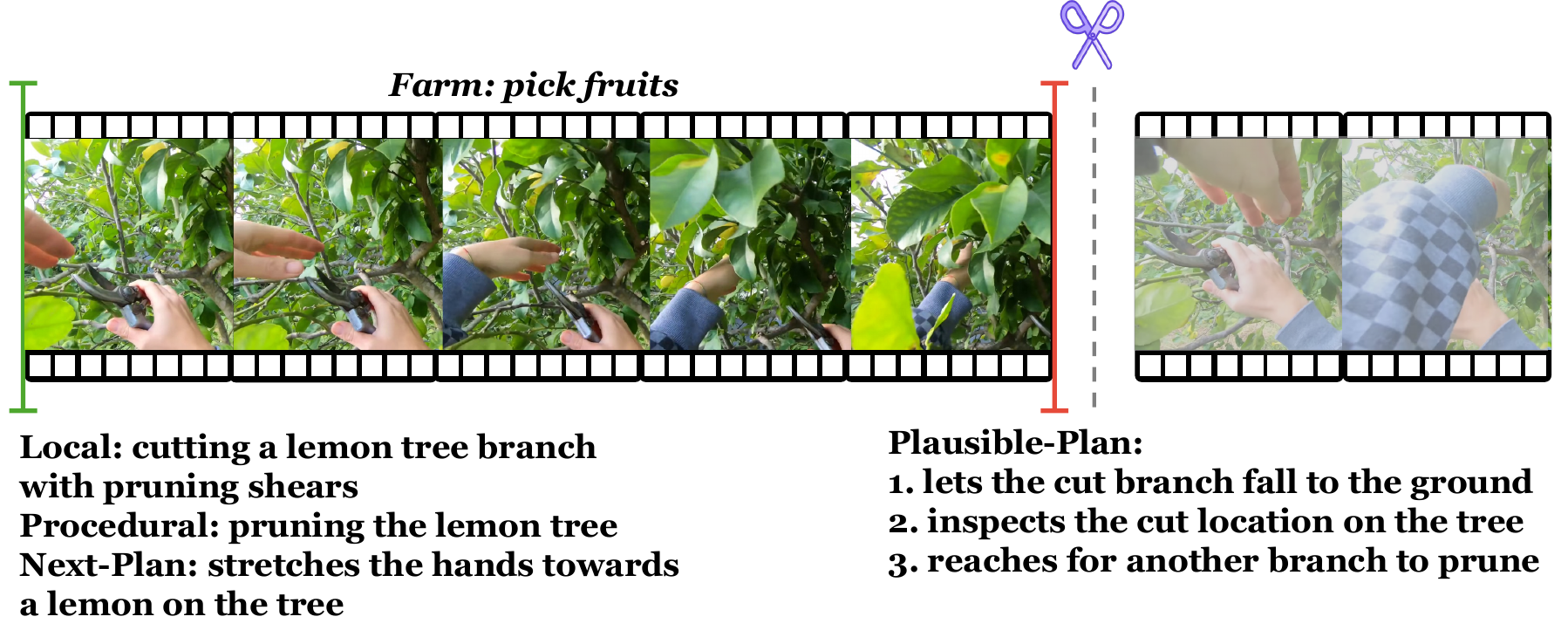}{Farm}{Pruning a branch on a lemon tree.}{fig:app_scene_farm}

\sceneexample{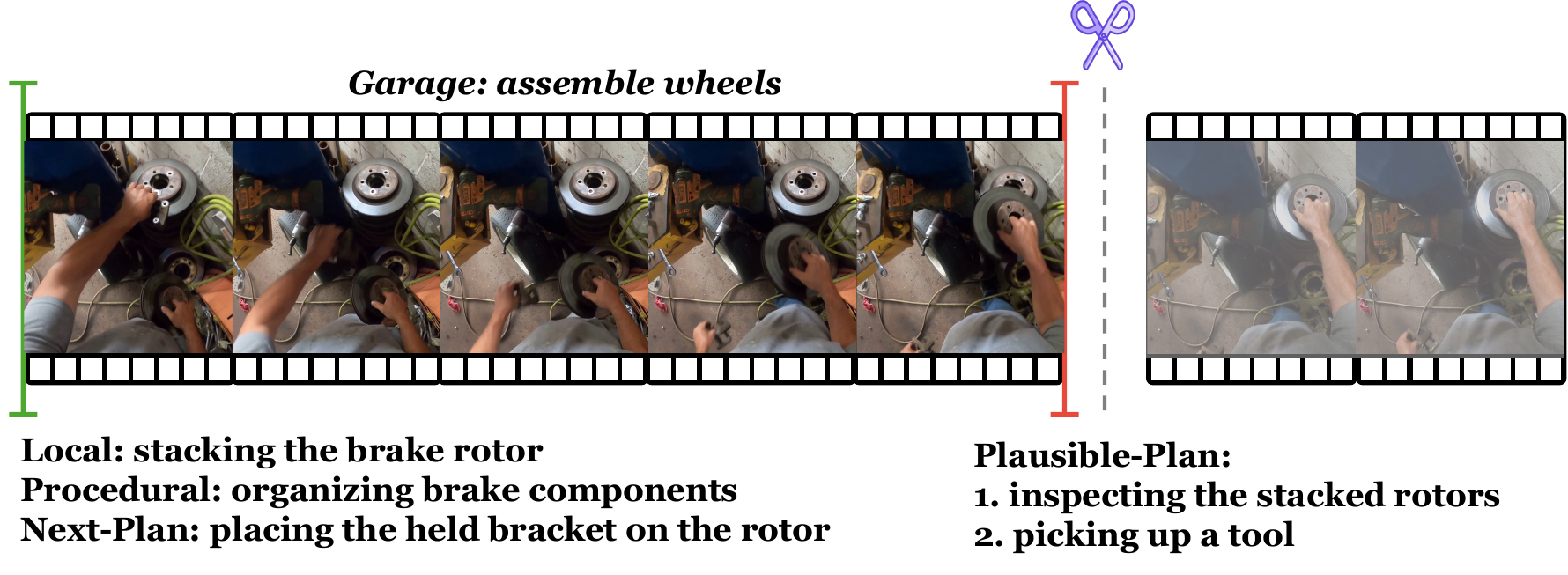}{Garage}{Stacking brake rotors while organizing components.}{fig:app_scene_garage}

\sceneexample{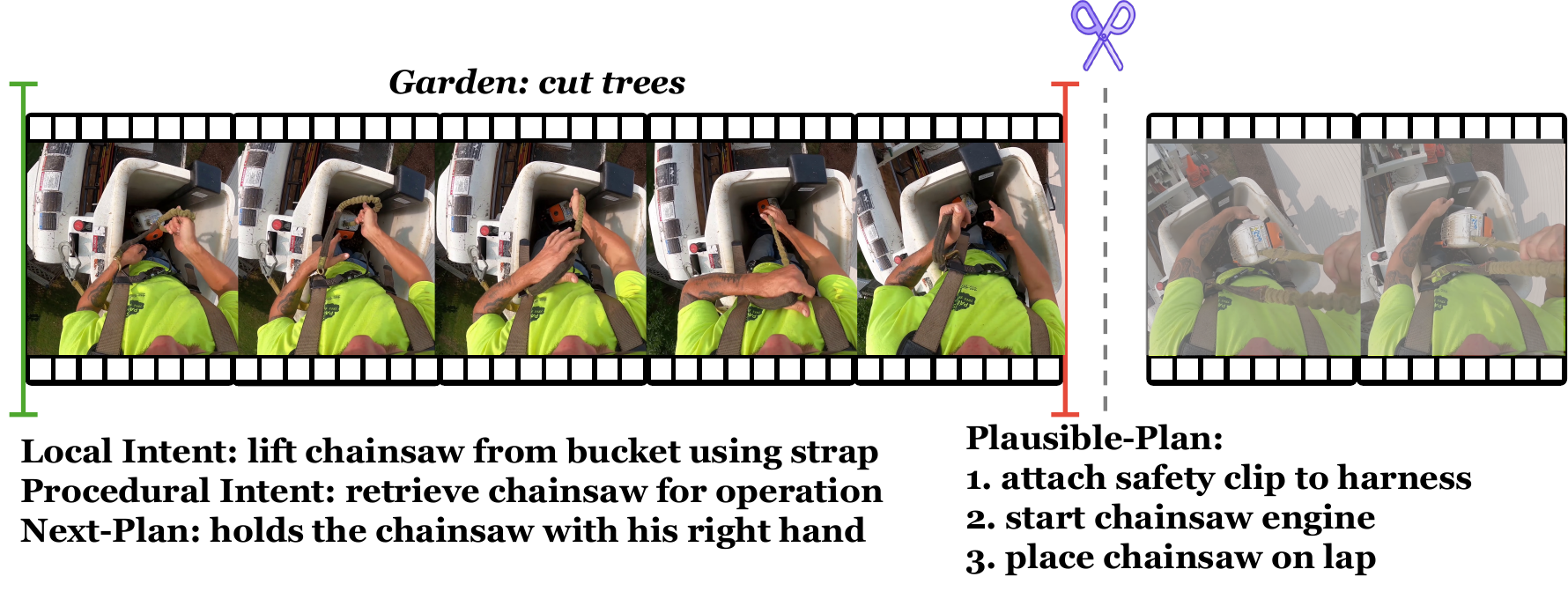}{Garden}{Lifting a chainsaw in preparation for use.}{fig:app_scene_garden}

\sceneexample{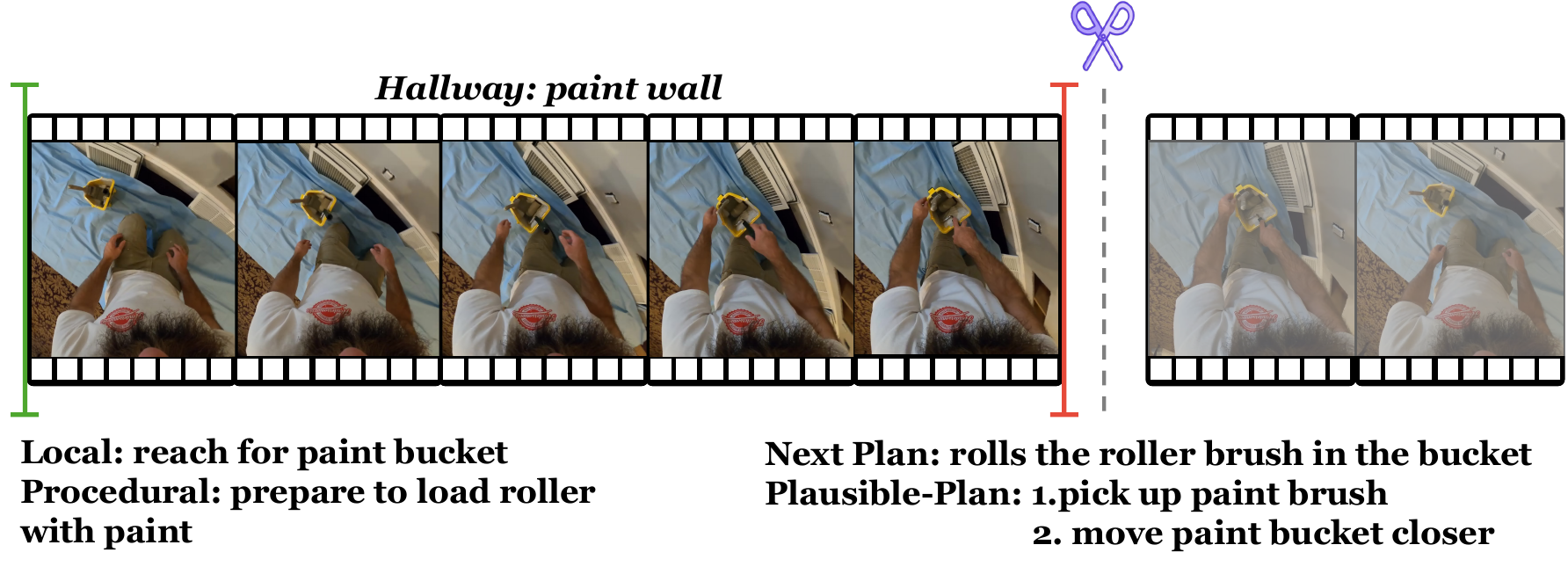}{Hallway}{Retrieving paint while preparing the roller.}{fig:app_scene_hallway}

\sceneexample{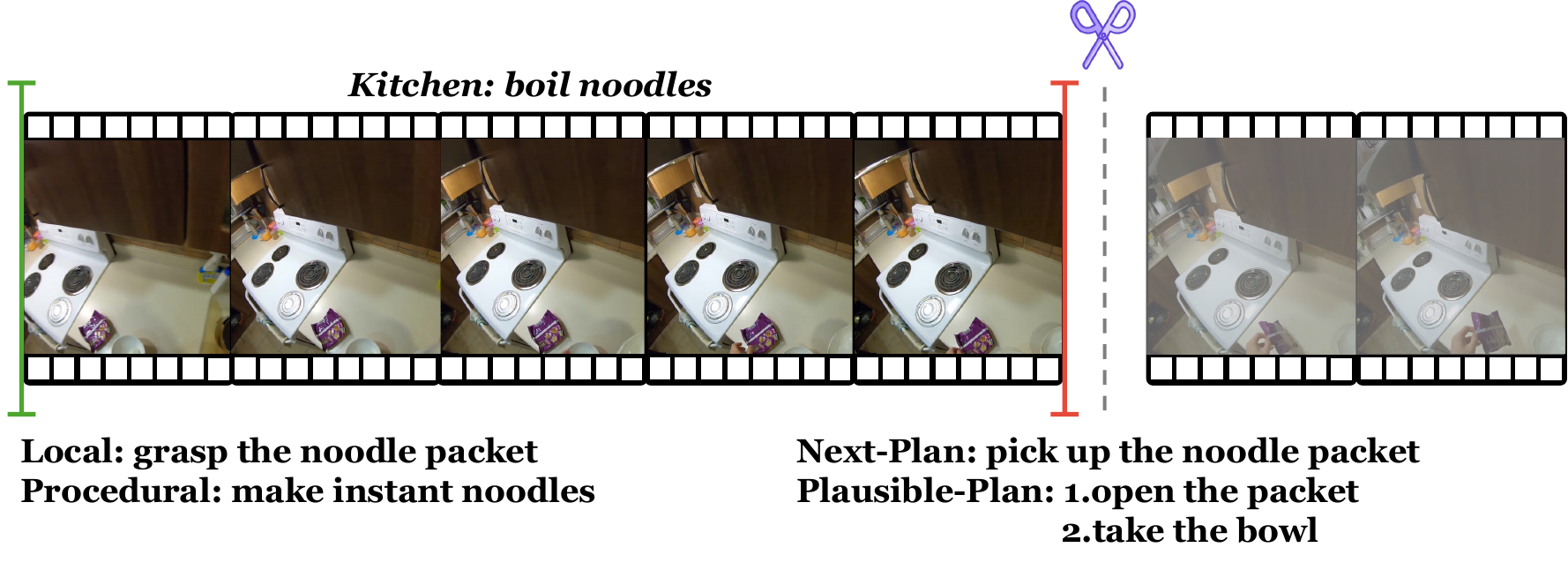}{Kitchen}{Grasping a noodle packet during meal preparation.}{fig:app_scene_kitchen}

\sceneexample{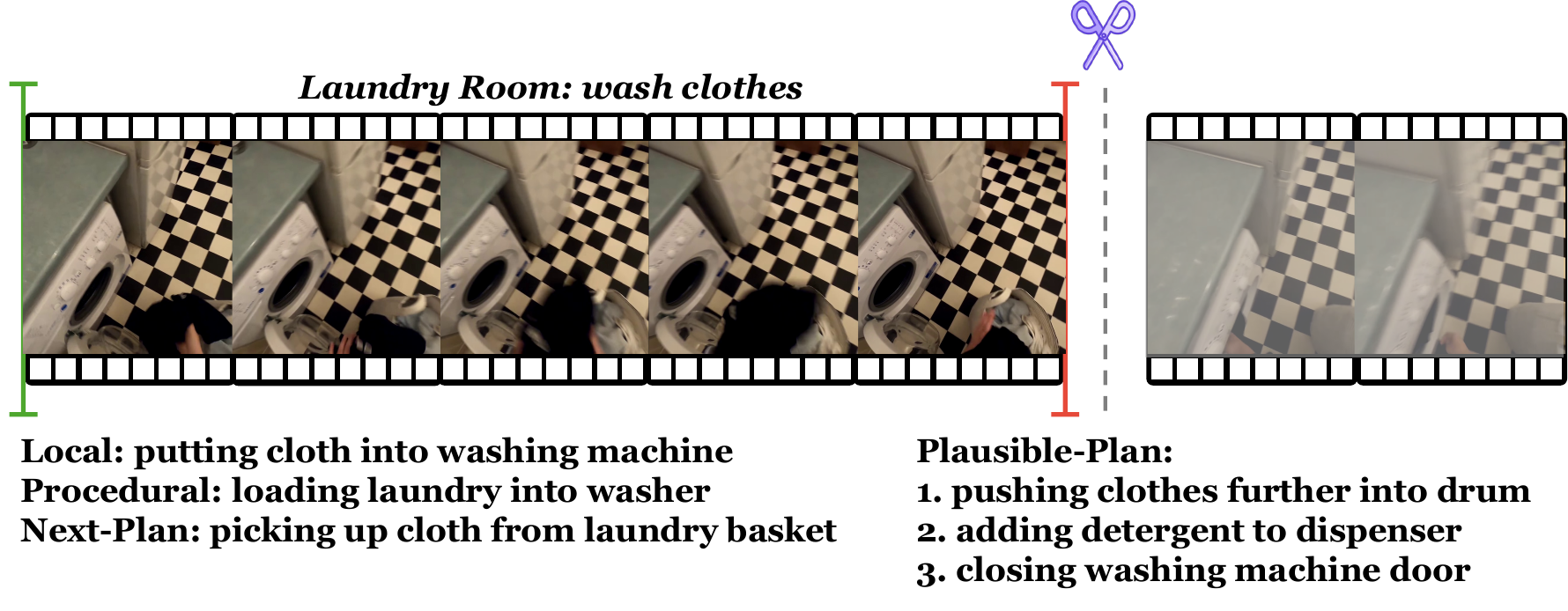}{Laundry Room}{Loading a cloth into the washing machine.}{fig:app_scene_laundry_room}

\sceneexample{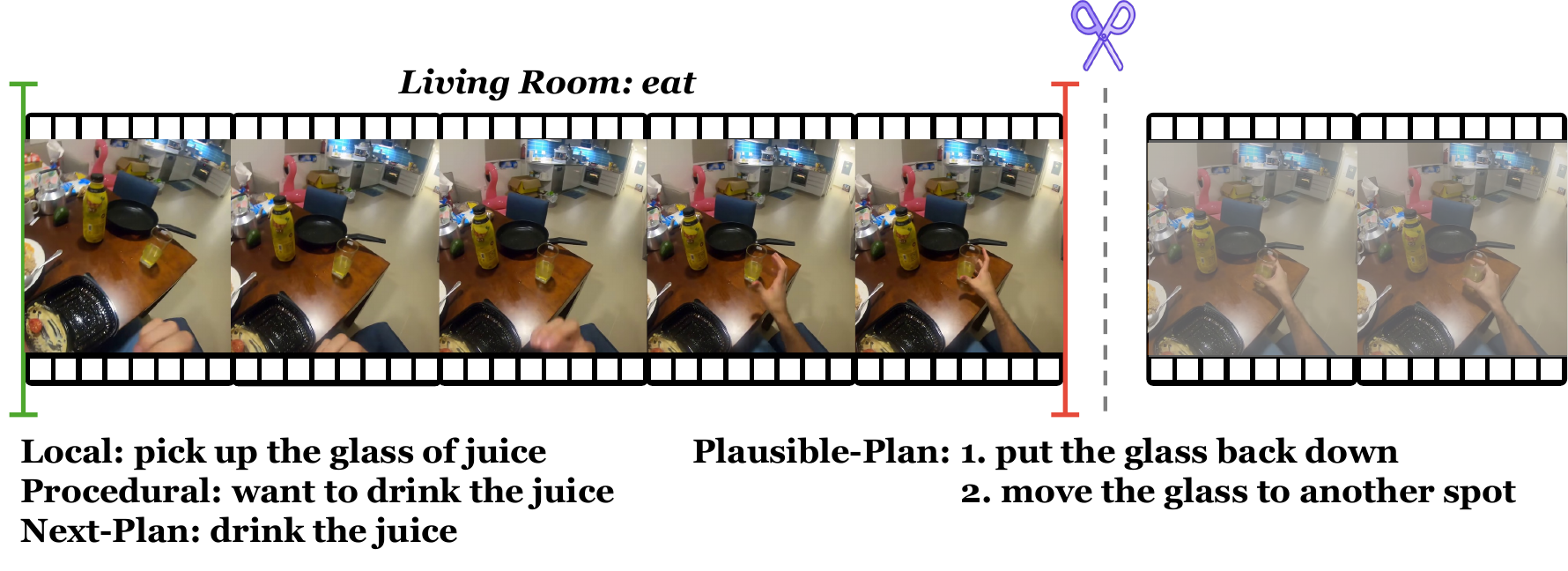}{Living Room}{Picking up a glass while drinking juice.}{fig:app_scene_living_room}

\sceneexample{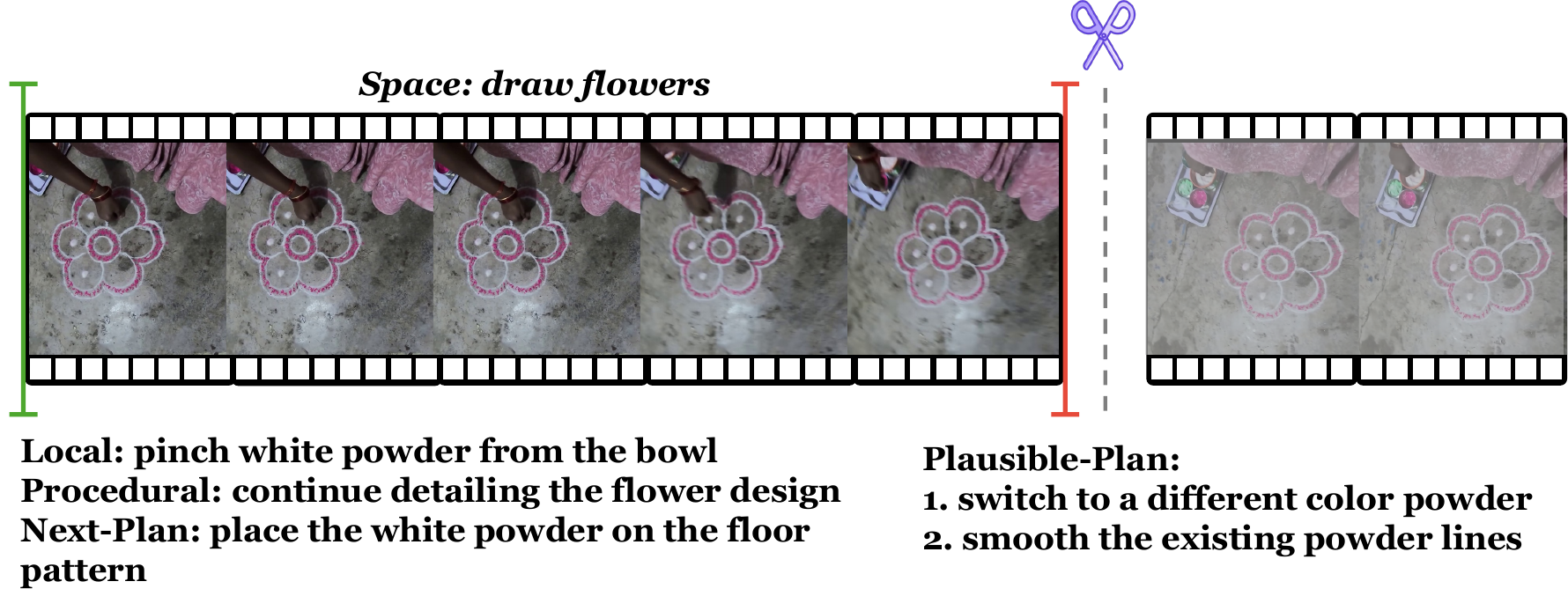}{Space}{Applying white powder to a decorative flower pattern.}{fig:app_scene_space}

\sceneexample{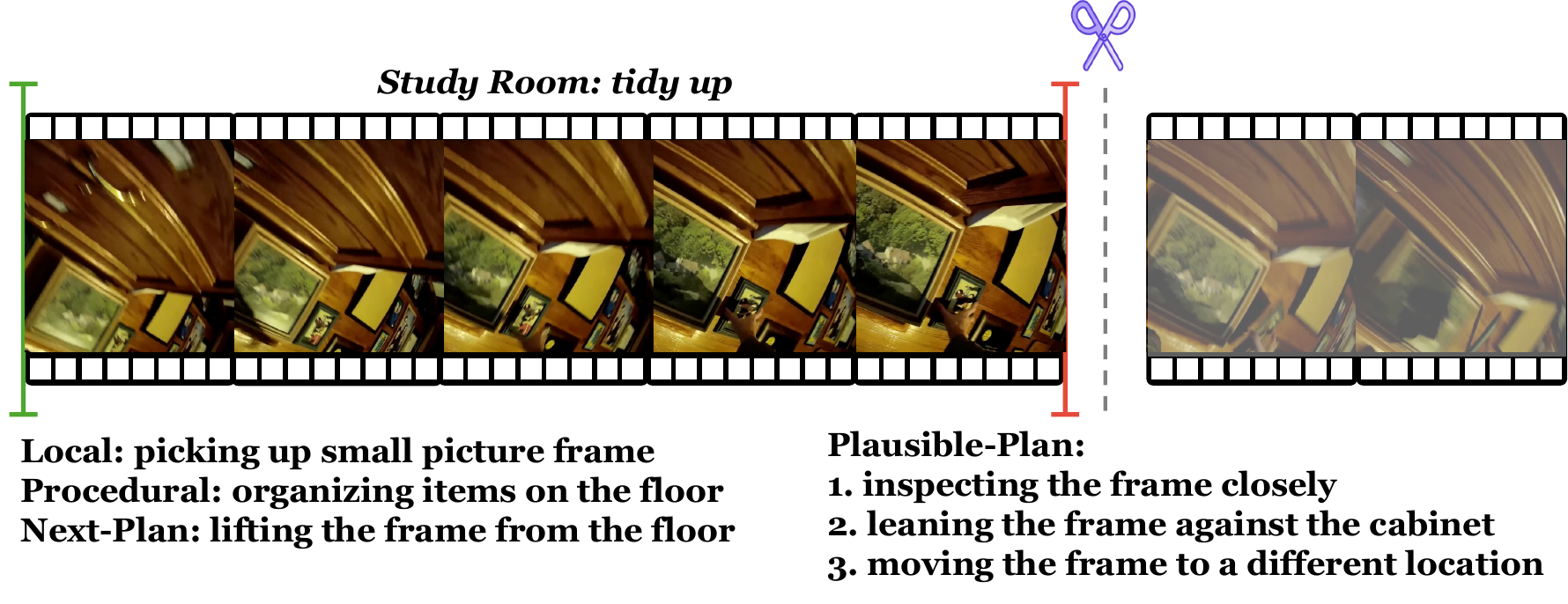}{Study Room}{Picking up a picture frame while tidying.}{fig:app_scene_study_room}

\sceneexample{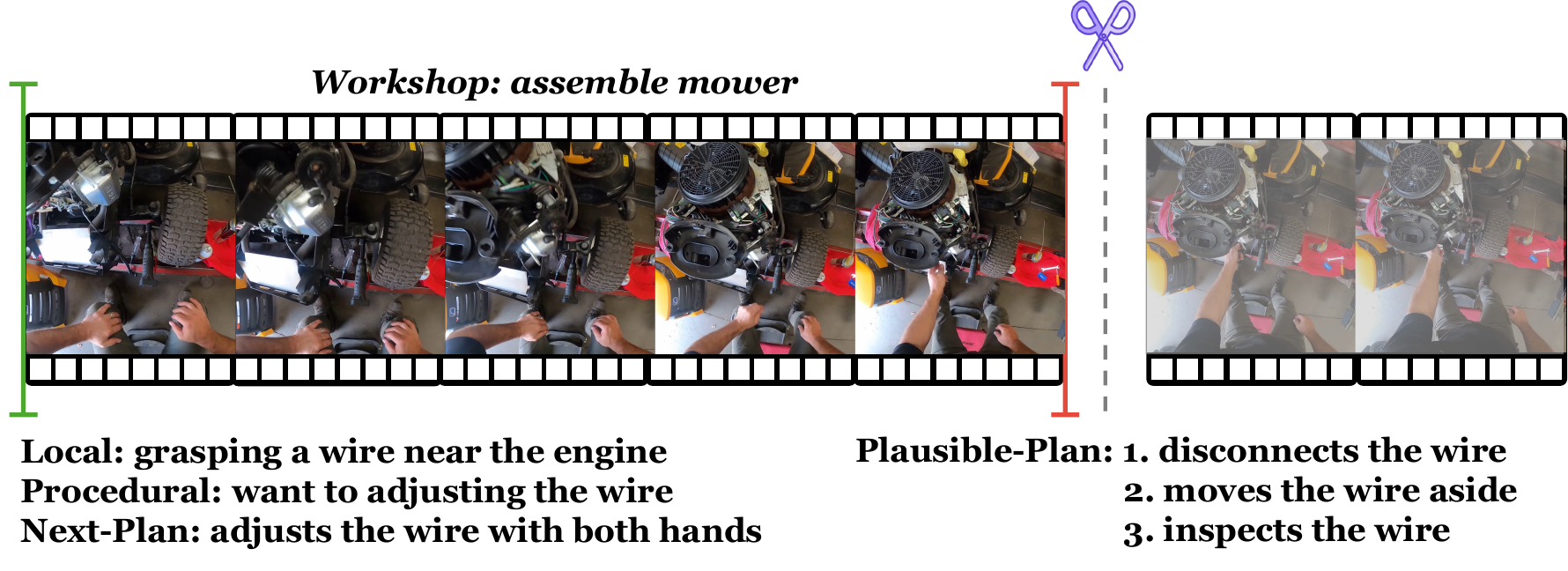}{Workshop}{Grasping a wire near the mower engine.}{fig:app_scene_workshop}

\sceneexample{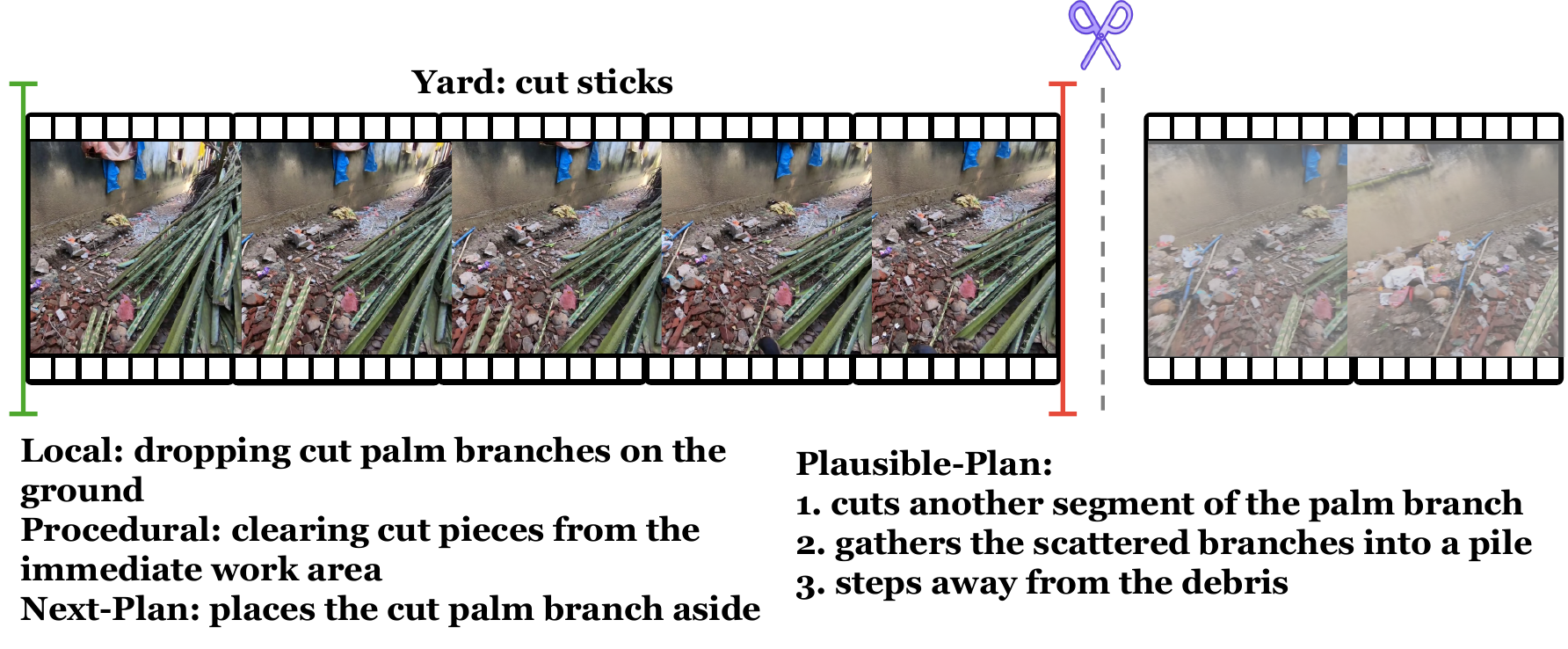}{Yard}{Dropping cut palm branches while clearing debris.}{fig:app_scene_yard}

\flushbottom
\section{Dataset Construction and Annotation Quality}

\subsection{Micro-step annotation and temporal boundaries}

EgoIntent decomposes each source video into temporally ordered micro-steps. Each micro-step contains a single coherent immediate goal and is associated with three textual targets:

\begin{itemize}
\item \textbf{\localcolor{Local Intent}}, the immediate purpose of the observed action;
\item \textbf{\proceduralcolor{Procedural Intent}}, the role of the local action in a larger procedure;
\item \textbf{\nextcolor{Next Plan}}, one or more actions that may directly follow the current step.

\end{itemize}

The released observation ends at an official pre-outcome cutoff. Annotators inspect the surrounding source-video context when defining the cutoff but only the frames at or before the cutoff are released as model input. The cutoff is intended to preserve ongoing-action evidence while hiding the decisive outcome and the next micro-step.

The following hierarchy guides the distinction between the two intent levels:

\begin{equation}
\mbox{visible action}
\rightarrow
\mbox{\localcolor{Local Intent}}
\rightarrow
\mbox{\proceduralcolor{Procedural Intent}}.
\label{eq:app_intent_hierarchy}
\end{equation}

Equation~(\ref{eq:app_intent_hierarchy}) makes explicit that \localcolor{Local Intent} mediates between the visible action and the more abstract \proceduralcolor{Procedural Intent}.

\localcolor{Local Intent} should be supported by the observed manipulation and object state. \proceduralcolor{Procedural Intent} should state why the local goal matters in the larger task, rather than paraphrasing the \localcolor{Local Intent} or naming the entire event. \nextcolor{Next Plan} should begin after the current micro-step and should not merely restate its completion.

\subsection{Independent annotation audit}

We conducted a separate audit on the frozen benchmark before releasing the corrected annotations. The audit subset, \texttt{\detokenize{audit756_v1}}, contains 756 unique steps sampled from all 3,014 benchmark steps. Sampling was completed before any audit outcome was observed, using seed \texttt{\detokenize{20260713}}. The allocation approximately preserved the benchmark distribution while covering all 15 scenes, all 32 events/source videos, and all eight duration intervals. Within each event-by-duration stratum, deterministic seeded sampling was used subject to the event and global duration quotas. The audit artifacts were frozen with the following checksums:

\checksumline{Canonical single-column sample-list SHA-256}{2a6af658f159136f85c3ff1736e21791829c1e240c69fc31eaa574e741333384}
\checksumline{Complete frozen sampling-manifest SHA-256}{2a64a6ed7e95e9f32cfb809954dfdc35feba3a346bba76d90acba8336c0c4c69}

\begin{center}
\captionof{table}{Audit-subset allocation across the 15 EgoIntent scenes.}
\label{tab:app_audit_scene_allocation}
{
\small
\setlength{\tabcolsep}{3.5pt}
\renewcommand{\arraystretch}{1.08}
\begin{tabular}{ccc}
\toprule
\multicolumn{1}{c}{\textbf{Scene}} & \textbf{Benchmark N} & \textbf{Audit N} \\
\midrule
Workshop & 459 & 116 \\
Kitchen & 396 & 100 \\
Garage & 321 & 80 \\
Garden & 309 & 77 \\
Living Room & 198 & 49 \\
Farm & 214 & 54 \\
Bedroom & 191 & 48 \\
Space & 113 & 28 \\
Deck & 106 & 27 \\
Laundry Room & 106 & 27 \\
Study Room & 202 & 50 \\
Clothing Closet & 104 & 26 \\
Yard & 104 & 26 \\
Hallway & 101 & 25 \\
Art Studio & 90 & 23 \\
Total & 3,014 & 756 \\
\bottomrule
\end{tabular}
}
\end{center}

Three domain-expert reviewers who did not participate in the original annotation independently examined the same 756 samples. Reviewers were blind to model predictions, automatic Judge scores, original annotator identity, and the other reviewers' decisions. For every sample, they first inspected the released observation together with its \localcolor{Local} and \proceduralcolor{Procedural} annotations, and then inspected the surrounding source-video context and the \nextcolor{Next Plan} annotation. Across the human-baseline study, Judge-validation study, and annotation-quality audit (P0-1, P0-2, and P0-6), the human evaluation and audit work required approximately 300 person-hours in total.

Each reviewer assigned \texttt{\detokenize{Valid}}, \texttt{\detokenize{Invalid}}, or \texttt{\detokenize{Uncertain}} to seven criteria:

\begin{center}
\captionof{table}{Annotation-audit criteria and validity requirements.}
\label{tab:app_audit_criteria}
{
\small
\setlength{\tabcolsep}{3.5pt}
\renewcommand{\arraystretch}{1.08}
\newcommand{\auditcriteriarow}{\rule[-2.2ex]{0pt}{5.2ex}}
\renewcommand{\tabularxcolumn}[1]{m{#1}}
\begin{tabularx}{0.90\textwidth}{>{\centering\arraybackslash}m{0.05\textwidth}@{\hspace{2pt}}>{\centering\arraybackslash}m{0.27\textwidth}>{\centering\arraybackslash}X}
\toprule
\textbf{ID} & \textbf{Criterion} & \textbf{Validity requirement} \\
\midrule
\auditcriteriarow C1 & Step coherence & The step contains one coherent immediate goal. \\
\auditcriteriarow C2 & Duration limit & The step duration does not exceed 10 seconds. \\
\auditcriteriarow C3 & Pre-outcome boundary & The observation ends before the decisive outcome becomes visible. \\
\auditcriteriarow C4 & \localcolor{Local} validity & \localcolor{Local Intent} agrees with the visible action and object state. \\
\auditcriteriarow C5 & \proceduralcolor{Procedural} validity & \proceduralcolor{Procedural Intent} correctly describes the local goal's role in the larger procedure. \\
\auditcriteriarow C6 & Hierarchical distinction & \localcolor{Local} and \proceduralcolor{Procedural} intents occupy distinguishable abstraction levels. \\
\auditcriteriarow C7 & \nextcolor{Next Plan} validity & \nextcolor{Next Plan} describes an action that directly follows the current step. \\
\bottomrule
\end{tabularx}
}
\end{center}

A sample was considered overall valid only when all seven criteria passed. Majority decisions were used when at least two reviewers agreed; unresolved cases were jointly adjudicated. Inter-rater statistics were computed from the independent pre-adjudication labels. Overall validity was computed from the adjudicated labels before correction.

\subsection{Audit statistics}

For a binary validity criterion, let $n$ be the number of audited samples, $x$ the number judged valid, and $\hat p=x/n$. We report the Wilson interval

\begin{equation}
\frac{
\hat p+\frac{z^2}{2n}
\pm
z\sqrt{\frac{\hat p(1-\hat p)}{n}+\frac{z^2}{4n^2}}
}{
1+\frac{z^2}{n}
},
\qquad z=1.96.
\label{eq:app_wilson_interval}
\end{equation}

In Eq.~(\ref{eq:app_wilson_interval}), setting $z=1.96$ gives the reported 95\% Wilson interval for each binary audit criterion.

The interval is descriptive because it does not account for within-video dependence. A source-video-clustered interval should be used if the individual audit records are included in the final release.

Before correction, 724 of 756 audited samples passed all criteria, giving an overall validity rate of 95.77\% (Wilson 95\% CI: [94.09\%, 96.99\%]). All individual criteria exceeded 96\%, and every sampled step satisfied the 10-second duration limit.

\begin{center}
\captionof{table}{Pre-correction annotation validity on the independently audited subset.}
\label{tab:app_audit_validity}
{
\small
\setlength{\tabcolsep}{3.5pt}
\renewcommand{\arraystretch}{1.08}
\begin{tabular}{crcr}
\toprule
\textbf{Audit criterion} & \textbf{Valid / N} & \textbf{Validity rate} & \textbf{Wilson 95\% CI} \\
\midrule
Step coherence & 746 / 756 & 98.68\% & [97.58\%, 99.28\%] \\
Duration limit & 756 / 756 & 100.00\% & [99.49\%, 100.00\%] \\
Pre-outcome boundary & 726 / 756 & 96.03\% & [94.39\%, 97.21\%] \\
\localcolor{Local} validity & 740 / 756 & 97.88\% & [96.59\%, 98.69\%] \\
\proceduralcolor{Procedural} validity & 733 / 756 & 96.96\% & [95.48\%, 97.96\%] \\
Hierarchical distinction & 730 / 756 & 96.56\% & [95.01\%, 97.64\%] \\
\nextcolor{Next Plan} validity & 727 / 756 & 96.16\% & [94.55\%, 97.32\%] \\
Overall validity before correction & 724 / 756 & 95.77\% & [94.09\%, 96.99\%] \\
\bottomrule
\end{tabular}
}
\end{center}

Across C1 and C3--C7, the macro-average raw agreement was 94.3\% and the macro-average Fleiss $\kappa$ was 0.67. C2 was excluded from this average because every sample passed, leaving no category variation. Agreement was lowest for \nextcolor{Next Plan} validity ($\kappa=0.58$), consistent with the existence of multiple plausible continuations.

\subsection{Error taxonomy and correction}

The 32 invalid samples could contain more than one error. The most common issue was that the annotated \nextcolor{Next Plan} was not the immediate continuation (29 samples). Other recurring errors were semantic overlap between \localcolor{Local} and \proceduralcolor{Procedural} intents (26), an inaccurate \proceduralcolor{Procedural Intent} (23), and an observation that already exposed the decisive outcome (23). Thirteen samples supported an additional plausible next-action branch that had not been recorded.

\begin{center}
\captionof{table}{Adjudicated corrections applied after the annotation audit.}
\label{tab:app_audit_corrections}
{
\small
\setlength{\tabcolsep}{3.5pt}
\renewcommand{\arraystretch}{1.08}
\begin{tabular}{cc}
\toprule
\multicolumn{1}{c}{\textbf{Correction}} & \textbf{Number of affected samples} \\
\midrule
Move the observation cutoff & 30 \\
Redefine a multi-goal micro-step & 10 \\
Revise \localcolor{Local Intent} & 16 \\
Revise \proceduralcolor{Procedural Intent} & 23 \\
Revise \nextcolor{Next Plan} & 29 \\
Add a plausible next action & 13 \\
At least one completed correction & 32 \\
Deleted steps & 0 \\
\bottomrule
\end{tabular}
}
\end{center}

All 32 adjudicated problems were corrected. No micro-step was deleted, so the corrected benchmark remains at 3,014 steps. The audit statistics describe the pre-correction state, while the released annotations contain the adjudicated corrections.

\section{Evaluation Protocol and Additional Full-Benchmark Diagnostics}

\subsection{Official prediction and reference-based task scores}

The exact benchmark-prediction prompt package is reproduced in Section~\ref{sec:app_prompt_official_prediction}.

For sample $i$, model $k$, and target dimension $d\in\{L,P,N\}$, let $q_{ikd}\in[0,100]$ be the semantic correctness score assigned to the prediction relative to the frozen reference set. A missing or invalid prediction receives zero. The dimension-level micro average is

\begin{equation}
M_{k,d}=\frac{1}{N}\sum_{i=1}^{N}q_{ikd},
\qquad N=3014.
\label{eq:app_task_dimension_micro}
\end{equation}

The benchmark Overall score is the unweighted mean of the three targets:

\begin{equation}
M_{k,\mathrm{overall}}
=
\frac{M_{k,L}+M_{k,P}+M_{k,N}}{3}.
\label{eq:app_task_overall}
\end{equation}

Equation~(\ref{eq:app_task_dimension_micro}) computes each reference-based task score over all benchmark steps, while Eq.~(\ref{eq:app_task_overall}) defines Overall as their unweighted mean.

For \nextcolor{Next Plan}, the comparison uses all valid frozen references rather than requiring a single surface form. This reference-based score asks whether a prediction matches an annotated target; it should not be interpreted as a direct measure of whether an unmatched continuation is physically plausible.

The frozen evaluation protocol uses \texttt{\detokenize{deepseek-v4-flash}} for the three reference-based dimensions. Predictions are anonymized before scoring, and the same scoring protocol is applied across models.

The exact reference-based semantic Judge package is reproduced in Section~\ref{sec:app_prompt_semantic_judge}.

\subsection{Reference-free diagnostic scores}

We complement reference matching with four video-grounded diagnostics. The video Judge receives the pre-outcome observation and an anonymized model prediction, but not the reference answer. It assigns an ordinal score

\begin{equation}
s_{i,m}^{(k)}\in\{0,1,2,3,4\},
\qquad
m\in\{\mathrm{GF,HIC,TPC,NPF}\}.
\label{eq:app_diagnostic_ordinal}
\end{equation}

The score is converted to a 0--100 scale by

\begin{equation}
x_{i,m}^{(k)}=25s_{i,m}^{(k)}.
\label{eq:app_diagnostic_scaled}
\end{equation}

Model-level scores are step-level micro averages:

\begin{equation}
M_m^{(k)}
=
\frac{1}{N}\sum_{i=1}^{N}x_{i,m}^{(k)}
=
\frac{25}{N}\sum_{i=1}^{N}s_{i,m}^{(k)}.
\label{eq:app_diagnostic_micro}
\end{equation}

The four-diagnostic average is

\begin{equation}
M_4^{(k)}
=
\frac{M_{\mathrm{GF}}^{(k)}+
M_{\mathrm{HIC}}^{(k)}+
M_{\mathrm{TPC}}^{(k)}+
M_{\mathrm{NPF}}^{(k)}}{4}.
\label{eq:app_diagnostic_average}
\end{equation}

Equation~(\ref{eq:app_diagnostic_ordinal}) defines the Judge's ordinal output, Eq.~(\ref{eq:app_diagnostic_scaled}) maps it to the 0--100 scale, Eq.~(\ref{eq:app_diagnostic_micro}) aggregates it over steps, and Eq.~(\ref{eq:app_diagnostic_average}) averages the four diagnostics.

Missing predictions receive zero for all four diagnostics. The four scores are explanatory measurements and do not replace the reference-based benchmark ranking.

The frozen evaluation protocol uses a Kimi-K2.5 video Judge for these diagnostics. For each step, the 15 anonymous candidates are deterministically shuffled before being independently scored.

The exact reference-free video diagnostic Judge package is reproduced in Section~\ref{sec:app_prompt_diagnostic_judge}.

\begingroup
\makeatletter
\renewcommand{\subsubsection}{%
  \@startsection{subsubsection}{3}{\z@}{-1pt}{-1em}{\normalsize\bf}%
}
\makeatother

\subsubsection{Grounding Faithfulness (GF)}

GF measures whether the three outputs are supported by visible evidence and avoid hallucinated objects, unobserved future actions, and contradictions.

\begingroup
\centering
\captionof{table}{Operational rubric for Grounding Faithfulness (GF).}
\label{tab:app_gf_rubric}
{
\small
\setlength{\tabcolsep}{3.5pt}
\renewcommand{\arraystretch}{1.08}
\begin{tabular}{cc}
\toprule
\textbf{Raw score} & \multicolumn{1}{c}{\textbf{Operational interpretation}} \\
\midrule
4 & All critical claims are clearly supported by visible evidence. \\
3 & The prediction is visually supported overall, with minor reasonable inference. \\
2 & Only part of the prediction is supported; obvious guessing remains. \\
1 & The prediction mainly relies on weak cues or scene priors. \\
0 & The prediction hallucinates, leaks future information, or contradicts the video. \\
\bottomrule
\end{tabular}
}
\par
\endgroup

\subsubsection{Hierarchical Intent Consistency (HIC)}

HIC measures whether \localcolor{Local} and \proceduralcolor{Procedural} intents form a meaningful abstraction hierarchy.

\begingroup
\centering
\captionof{table}{Operational rubric for Hierarchical Intent Consistency (HIC).}
\label{tab:app_hic_rubric}
{
\small
\setlength{\tabcolsep}{3.5pt}
\renewcommand{\arraystretch}{1.08}
\begin{tabular}{cc}
\toprule
\textbf{Raw score} & \multicolumn{1}{c}{\textbf{Operational interpretation}} \\
\midrule
4 & The hierarchy is clear and \proceduralcolor{Procedural} explains the function of \localcolor{Local}. \\
3 & The relation is correct but somewhat broad or overlapping. \\
2 & The two statements are related but the hierarchy is unclear. \\
1 & They are near-paraphrases, or \proceduralcolor{Procedural} is excessively broad. \\
0 & The two intents are contradictory or unrelated. \\
\bottomrule
\end{tabular}
}
\par
\endgroup

\subsubsection{Temporal Progression Consistency (TPC)}

TPC measures whether \localcolor{Local}, \proceduralcolor{Procedural}, and \nextcolor{Next Plan} form a correctly ordered progression across the observation boundary.

\begingroup
\centering
\captionof{table}{Operational rubric for Temporal Progression Consistency (TPC).}
\label{tab:app_tpc_rubric}
{
\small
\setlength{\tabcolsep}{3.5pt}
\renewcommand{\arraystretch}{1.08}
\begin{tabular}{cc}
\toprule
\textbf{Raw score} & \multicolumn{1}{c}{\textbf{Operational interpretation}} \\
\midrule
4 & Current and next-step boundaries are clear and the progression is natural. \\
3 & The order is correct with minor boundary ambiguity. \\
2 & The progression is broadly plausible but current/next membership is unclear. \\
1 & \nextcolor{Next} overlaps the current action or jumps too far ahead. \\
0 & The temporal order is wrong or the predicted next action already occurred. \\
\bottomrule
\end{tabular}
}
\par
\endgroup

\subsubsection{\nextcolor{Next Plan} Feasibility (NPF)}

NPF measures whether the predicted continuation is physically and procedurally executable from the observed state, even when it does not exactly match the reference.

\begingroup
\centering
\captionof{table}{Operational rubric for \nextcolor{Next Plan} Feasibility (NPF).}
\label{tab:app_npf_rubric}
{
\small
\setlength{\tabcolsep}{3.5pt}
\renewcommand{\arraystretch}{1.08}
\begin{tabular}{cc}
\toprule
\textbf{Raw score} & \multicolumn{1}{c}{\textbf{Operational interpretation}} \\
\midrule
4 & Immediately executable and strongly compatible with the current state. \\
3 & Plausible but broad or indirect. \\
2 & Possible but missing an important precondition. \\
1 & Physically possible but procedurally unnatural. \\
0 & Impossible, contradictory, or dependent on absent objects. \\
\bottomrule
\end{tabular}
}
\par
\endgroup

\endgroup

\subsection{Detailed Per-Model Scene-Level Scores}

The following 15 tables report each model's micro-average scores in every scene. Overall is the arithmetic mean of \localcolor{Local}, \proceduralcolor{Procedural}, and \nextcolor{Next}, while GF, HIC, TPC, and NPF are reported individually.

\subsubsection{Closed-Source Models}\mbox{}\par

\begin{modeltable}
\captionof{table}{Scene-level performance of \textbf{Doubao-Seed-2.1-Turbo} on the full EgoIntent benchmark.}
\label{tab:app_scene_doubao}
\nopagebreak[4]
{
\small
\renewcommand{\arraystretch}{1.00}
\begin{tabular*}{\textwidth}{@{\extracolsep{\fill}}lcccccccc@{}}
\toprule
\textbf{Scene} & \textbf{\localcolor{Local}} & \textbf{\proceduralcolor{Procedural}} & \textbf{\nextcolor{Next}} & \textbf{Overall} & \textbf{GF} & \textbf{HIC} & \textbf{TPC} & \textbf{NPF} \\
\midrule
indoor/art\_studio & 65.11 & 65.17 & 50.03 & 60.10 & 81.50 & 89.44 & 85.00 & 83.00 \\
indoor/bedroom & 63.12 & 69.37 & 42.71 & 58.40 & 74.52 & 86.52 & 80.89 & 79.02 \\
indoor/clothing\_closet & 63.94 & 71.68 & 51.54 & 62.39 & 73.52 & 84.62 & 81.01 & 80.21 \\
indoor/garage & 44.56 & 49.91 & 33.58 & 42.68 & 67.54 & 80.84 & 73.60 & 71.13 \\
indoor/hallway & 46.32 & 58.32 & 40.56 & 48.40 & 79.20 & 89.60 & 86.88 & 84.88 \\
indoor/kitchen & 65.05 & 71.35 & 45.18 & 60.53 & 80.31 & 87.94 & 81.25 & 80.01 \\
indoor/laundry\_room & 58.41 & 64.67 & 50.83 & 57.97 & 67.63 & 80.19 & 73.11 & 71.58 \\
indoor/living\_room & 51.82 & 57.17 & 39.93 & 49.64 & 65.41 & 75.13 & 67.30 & 64.67 \\
indoor/study\_room & 55.59 & 59.11 & 43.59 & 52.76 & 70.41 & 85.02 & 74.13 & 70.77 \\
indoor/workshop & 57.42 & 61.56 & 41.42 & 53.47 & 78.85 & 88.78 & 82.90 & 80.63 \\
outdoor/deck & 72.03 & 70.50 & 47.76 & 63.43 & 75.89 & 85.14 & 80.19 & 77.25 \\
outdoor/farm & 64.17 & 71.86 & 48.22 & 61.42 & 77.15 & 87.85 & 81.31 & 78.84 \\
outdoor/garden & 55.99 & 63.83 & 48.09 & 55.97 & 78.21 & 87.14 & 82.85 & 81.33 \\
outdoor/space & 66.55 & 75.58 & 51.35 & 64.49 & 75.55 & 88.72 & 80.97 & 78.97 \\
outdoor/yard & 50.10 & 53.27 & 38.56 & 47.31 & 57.41 & 71.88 & 65.62 & 62.66 \\
\bottomrule
\end{tabular*}
}
\end{modeltable}

\begin{modeltable}
\captionof{table}{Scene-level performance of \textbf{Qwen3.5-Plus} on the full EgoIntent benchmark.}
\label{tab:app_scene_qwen35}
\nopagebreak[4]
{
\small
\renewcommand{\arraystretch}{1.00}
\begin{tabular*}{\textwidth}{@{\extracolsep{\fill}}lcccccccc@{}}
\toprule
\textbf{Scene} & \textbf{\localcolor{Local}} & \textbf{\proceduralcolor{Procedural}} & \textbf{\nextcolor{Next}} & \textbf{Overall} & \textbf{GF} & \textbf{HIC} & \textbf{TPC} & \textbf{NPF} \\
\midrule
indoor/art\_studio & 55.33 & 60.61 & 39.71 & 51.89 & 65.67 & 77.78 & 69.72 & 68.56 \\
indoor/bedroom & 62.88 & 66.54 & 48.91 & 59.44 & 75.57 & 82.46 & 79.84 & 78.37 \\
indoor/clothing\_closet & 68.37 & 61.88 & 48.45 & 59.56 & 81.93 & 76.20 & 82.45 & 81.41 \\
indoor/garage & 47.63 & 45.64 & 30.65 & 41.30 & 66.83 & 78.74 & 72.27 & 69.81 \\
indoor/hallway & 50.25 & 47.57 & 40.88 & 46.23 & 69.54 & 76.73 & 74.01 & 73.50 \\
indoor/kitchen & 67.25 & 67.63 & 46.73 & 60.54 & 75.83 & 79.42 & 76.52 & 76.03 \\
indoor/laundry\_room & 62.22 & 66.93 & 57.47 & 62.21 & 72.35 & 79.72 & 74.53 & 73.24 \\
indoor/living\_room & 58.47 & 58.59 & 43.75 & 53.60 & 73.12 & 81.31 & 77.15 & 76.16 \\
indoor/study\_room & 52.95 & 53.19 & 40.54 & 48.89 & 72.02 & 83.04 & 75.12 & 73.12 \\
indoor/workshop & 54.77 & 53.31 & 37.92 & 48.67 & 73.84 & 84.04 & 78.70 & 77.30 \\
outdoor/deck & 60.33 & 53.25 & 42.15 & 51.91 & 75.65 & 85.38 & 82.08 & 79.60 \\
outdoor/farm & 62.50 & 64.09 & 43.89 & 56.83 & 76.57 & 87.03 & 79.44 & 77.32 \\
outdoor/garden & 54.69 & 57.07 & 46.86 & 52.87 & 75.62 & 83.66 & 79.13 & 78.02 \\
outdoor/space & 72.57 & 70.80 & 50.62 & 64.66 & 77.98 & 86.95 & 81.64 & 79.86 \\
outdoor/yard & 46.35 & 41.83 & 27.78 & 38.65 & 61.50 & 73.08 & 67.79 & 66.03 \\
\bottomrule
\end{tabular*}
}
\end{modeltable}

\begin{modeltable}
\captionof{table}{Scene-level performance of \textbf{Gemini 3.5 Flash} on the full EgoIntent benchmark.}
\label{tab:app_scene_gemini35}
\nopagebreak[4]
{
\small
\renewcommand{\arraystretch}{1.00}
\begin{tabular*}{\textwidth}{@{\extracolsep{\fill}}lcccccccc@{}}
\toprule
\textbf{Scene} & \textbf{\localcolor{Local}} & \textbf{\proceduralcolor{Procedural}} & \textbf{\nextcolor{Next}} & \textbf{Overall} & \textbf{GF} & \textbf{HIC} & \textbf{TPC} & \textbf{NPF} \\
\midrule
indoor/art\_studio & 39.89 & 46.44 & 32.10 & 39.48 & 56.94 & 71.67 & 61.11 & 59.94 \\
indoor/bedroom & 44.84 & 52.17 & 33.62 & 43.55 & 61.91 & 72.51 & 65.97 & 64.36 \\
indoor/clothing\_closet & 55.38 & 54.76 & 44.81 & 51.65 & 71.15 & 74.76 & 71.63 & 70.12 \\
indoor/garage & 32.15 & 34.13 & 22.29 & 29.52 & 45.72 & 61.99 & 50.55 & 47.92 \\
indoor/hallway & 35.25 & 39.65 & 32.72 & 35.87 & 60.40 & 69.31 & 64.85 & 63.35 \\
indoor/kitchen & 51.73 & 56.78 & 36.40 & 48.30 & 63.45 & 70.64 & 64.39 & 63.91 \\
indoor/laundry\_room & 45.19 & 52.08 & 43.25 & 46.84 & 54.01 & 62.26 & 56.84 & 54.84 \\
indoor/living\_room & 49.80 & 55.13 & 36.83 & 47.25 & 59.47 & 67.80 & 61.99 & 60.25 \\
indoor/study\_room & 36.11 & 45.77 & 31.21 & 37.70 & 50.00 & 65.10 & 54.46 & 51.47 \\
indoor/workshop & 40.60 & 44.68 & 27.40 & 37.56 & 59.37 & 72.22 & 65.52 & 62.49 \\
outdoor/deck & 50.09 & 44.15 & 39.62 & 44.62 & 55.42 & 68.16 & 59.91 & 58.61 \\
outdoor/farm & 49.50 & 59.21 & 35.83 & 48.18 & 66.59 & 78.39 & 69.28 & 67.28 \\
outdoor/garden & 41.65 & 48.88 & 36.11 & 42.21 & 62.70 & 73.06 & 66.42 & 64.67 \\
outdoor/space & 54.34 & 57.70 & 34.25 & 48.76 & 63.50 & 78.32 & 67.26 & 65.04 \\
outdoor/yard & 27.45 & 30.29 & 20.28 & 26.01 & 52.16 & 67.31 & 61.06 & 57.62 \\
\bottomrule
\end{tabular*}
}
\end{modeltable}

\begin{modeltable}
\captionof{table}{Scene-level performance of \textbf{Amazon-Nova-2-Lite-V1} on the full EgoIntent benchmark.}
\label{tab:app_scene_amazon_nova}
\nopagebreak[4]
{
\small
\renewcommand{\arraystretch}{1.00}
\begin{tabular*}{\textwidth}{@{\extracolsep{\fill}}lcccccccc@{}}
\toprule
\textbf{Scene} & \textbf{\localcolor{Local}} & \textbf{\proceduralcolor{Procedural}} & \textbf{\nextcolor{Next}} & \textbf{Overall} & \textbf{GF} & \textbf{HIC} & \textbf{TPC} & \textbf{NPF} \\
\midrule
indoor/art\_studio & 28.78 & 26.00 & 15.12 & 23.30 & 48.89 & 57.78 & 49.72 & 49.39 \\
indoor/bedroom & 40.39 & 45.50 & 26.74 & 37.54 & 59.82 & 61.26 & 62.96 & 62.01 \\
indoor/clothing\_closet & 47.31 & 49.42 & 34.90 & 43.88 & 61.30 & 57.21 & 59.62 & 59.78 \\
indoor/garage & 23.36 & 28.05 & 15.26 & 22.23 & 42.76 & 56.31 & 46.57 & 43.79 \\
indoor/hallway & 29.85 & 38.27 & 21.88 & 30.00 & 54.21 & 62.87 & 55.94 & 54.93 \\
indoor/kitchen & 41.35 & 45.96 & 21.67 & 36.33 & 54.67 & 60.80 & 52.90 & 51.41 \\
indoor/laundry\_room & 35.61 & 35.00 & 27.16 & 32.59 & 50.94 & 57.78 & 54.48 & 52.48 \\
indoor/living\_room & 38.48 & 44.72 & 27.85 & 37.02 & 49.49 & 58.08 & 49.12 & 47.75 \\
indoor/study\_room & 38.00 & 44.28 & 23.71 & 35.33 & 58.54 & 67.08 & 59.90 & 57.78 \\
indoor/workshop & 33.98 & 39.56 & 17.44 & 30.33 & 53.21 & 63.29 & 53.49 & 51.00 \\
outdoor/deck & 44.62 & 44.95 & 25.14 & 38.24 & 57.55 & 67.22 & 66.04 & 62.62 \\
outdoor/farm & 48.34 & 57.20 & 29.08 & 44.87 & 64.25 & 76.05 & 68.34 & 66.11 \\
outdoor/garden & 42.60 & 48.79 & 25.87 & 39.09 & 61.97 & 70.47 & 63.67 & 62.56 \\
outdoor/space & 53.89 & 61.33 & 28.76 & 47.99 & 64.38 & 77.43 & 65.49 & 63.71 \\
outdoor/yard & 32.21 & 34.23 & 16.62 & 27.69 & 49.52 & 63.94 & 57.69 & 54.49 \\
\bottomrule
\end{tabular*}
}
\end{modeltable}

\clearpage
\subsubsection{Open-Weight Models}\mbox{}\par

\begin{modeltable}
\captionof{table}{Scene-level performance of \textbf{Qwen3-VL-32B-Instruct} on the full EgoIntent benchmark.}
\label{tab:app_scene_qwen3vl32}
\nopagebreak[4]
{
\small
\renewcommand{\arraystretch}{1.00}
\begin{tabular*}{\textwidth}{@{\extracolsep{\fill}}lcccccccc@{}}
\toprule
\textbf{Scene} & \textbf{\localcolor{Local}} & \textbf{\proceduralcolor{Procedural}} & \textbf{\nextcolor{Next}} & \textbf{Overall} & \textbf{GF} & \textbf{HIC} & \textbf{TPC} & \textbf{NPF} \\
\midrule
indoor/art\_studio & 34.53 & 44.11 & 23.50 & 34.05 & 51.39 & 66.11 & 57.78 & 56.06 \\
indoor/bedroom & 48.46 & 57.07 & 32.96 & 46.16 & 68.19 & 78.53 & 71.34 & 69.20 \\
indoor/clothing\_closet & 59.95 & 71.35 & 40.10 & 57.13 & 88.46 & 92.31 & 90.87 & 88.87 \\
indoor/garage & 28.71 & 36.67 & 21.73 & 29.04 & 58.41 & 74.84 & 65.81 & 62.95 \\
indoor/hallway & 28.71 & 35.30 & 20.97 & 28.33 & 57.92 & 69.55 & 66.34 & 63.10 \\
indoor/kitchen & 53.33 & 61.89 & 32.63 & 49.28 & 73.99 & 82.45 & 77.21 & 75.53 \\
indoor/laundry\_room & 42.59 & 50.47 & 33.27 & 42.11 & 61.79 & 73.35 & 65.57 & 63.80 \\
indoor/living\_room & 41.52 & 45.35 & 28.29 & 38.39 & 66.92 & 78.28 & 70.83 & 68.08 \\
indoor/study\_room & 45.36 & 53.02 & 36.26 & 44.88 & 74.63 & 87.25 & 80.82 & 78.07 \\
indoor/workshop & 49.89 & 58.13 & 30.69 & 46.24 & 73.47 & 85.40 & 79.47 & 76.65 \\
outdoor/deck & 50.55 & 53.38 & 37.26 & 47.06 & 79.95 & 88.44 & 85.38 & 82.20 \\
outdoor/farm & 47.87 & 57.76 & 28.64 & 44.76 & 71.14 & 83.88 & 76.40 & 74.05 \\
outdoor/garden & 44.73 & 53.73 & 33.16 & 43.88 & 71.60 & 81.63 & 77.27 & 74.94 \\
outdoor/space & 67.52 & 73.89 & 43.70 & 61.71 & 71.46 & 85.40 & 76.11 & 73.88 \\
outdoor/yard & 22.69 & 22.64 & 13.00 & 19.45 & 69.71 & 82.21 & 77.88 & 75.16 \\
\bottomrule
\end{tabular*}
}
\end{modeltable}

\begin{modeltable}
\captionof{table}{Scene-level performance of \textbf{Qwen3-VL-8B-Instruct} on the full EgoIntent benchmark.}
\label{tab:app_scene_qwen3vl8}
\nopagebreak[4]
{
\small
\renewcommand{\arraystretch}{1.00}
\begin{tabular*}{\textwidth}{@{\extracolsep{\fill}}lcccccccc@{}}
\toprule
\textbf{Scene} & \textbf{\localcolor{Local}} & \textbf{\proceduralcolor{Procedural}} & \textbf{\nextcolor{Next}} & \textbf{Overall} & \textbf{GF} & \textbf{HIC} & \textbf{TPC} & \textbf{NPF} \\
\midrule
indoor/art\_studio & 38.22 & 41.28 & 23.72 & 34.41 & 54.17 & 62.50 & 54.17 & 52.72 \\
indoor/bedroom & 46.07 & 54.55 & 26.41 & 42.35 & 60.34 & 71.47 & 62.70 & 59.78 \\
indoor/clothing\_closet & 50.24 & 59.71 & 43.82 & 51.26 & 64.66 & 74.28 & 74.04 & 72.76 \\
indoor/garage & 28.88 & 30.84 & 17.85 & 25.86 & 51.48 & 64.33 & 56.00 & 53.37 \\
indoor/hallway & 30.15 & 37.92 & 26.98 & 31.68 & 53.96 & 60.15 & 58.17 & 55.67 \\
indoor/kitchen & 53.32 & 60.18 & 28.50 & 47.33 & 68.56 & 77.34 & 66.67 & 64.92 \\
indoor/laundry\_room & 45.94 & 52.31 & 33.27 & 43.84 & 59.20 & 70.99 & 62.97 & 60.74 \\
indoor/living\_room & 36.11 & 41.74 & 24.41 & 34.09 & 58.71 & 70.45 & 62.37 & 59.49 \\
indoor/study\_room & 30.40 & 37.90 & 26.87 & 31.72 & 56.31 & 70.92 & 63.49 & 60.00 \\
indoor/workshop & 39.81 & 41.86 & 25.87 & 35.85 & 64.60 & 72.44 & 69.12 & 66.90 \\
outdoor/deck & 42.74 & 34.34 & 28.92 & 35.33 & 63.68 & 71.70 & 74.29 & 71.35 \\
outdoor/farm & 46.21 & 52.71 & 28.33 & 42.42 & 63.79 & 75.35 & 65.77 & 63.54 \\
outdoor/garden & 40.52 & 43.04 & 26.86 & 36.81 & 62.62 & 71.93 & 66.34 & 64.59 \\
outdoor/space & 58.19 & 58.58 & 39.46 & 52.08 & 69.47 & 80.31 & 69.03 & 65.70 \\
outdoor/yard & 22.16 & 22.16 & 10.74 & 18.36 & 54.09 & 68.03 & 62.74 & 58.82 \\
\bottomrule
\end{tabular*}
}
\end{modeltable}

\begin{modeltable}
\captionof{table}{Scene-level performance of \textbf{Qwen2.5-VL-7B-Instruct} on the full EgoIntent benchmark.}
\label{tab:app_scene_qwen25vl7}
\nopagebreak[4]
{
\small
\renewcommand{\arraystretch}{1.00}
\begin{tabular*}{\textwidth}{@{\extracolsep{\fill}}lcccccccc@{}}
\toprule
\textbf{Scene} & \textbf{\localcolor{Local}} & \textbf{\proceduralcolor{Procedural}} & \textbf{\nextcolor{Next}} & \textbf{Overall} & \textbf{GF} & \textbf{HIC} & \textbf{TPC} & \textbf{NPF} \\
\midrule
indoor/art\_studio & 33.11 & 29.89 & 20.94 & 27.98 & 40.83 & 49.72 & 41.67 & 40.50 \\
indoor/bedroom & 37.43 & 48.53 & 22.97 & 36.31 & 50.13 & 65.18 & 50.00 & 48.26 \\
indoor/clothing\_closet & 45.19 & 52.60 & 30.75 & 42.85 & 63.46 & 71.63 & 63.70 & 61.22 \\
indoor/garage & 27.73 & 28.88 & 17.23 & 24.61 & 47.20 & 61.45 & 52.18 & 48.78 \\
indoor/hallway & 31.24 & 38.22 & 24.90 & 31.45 & 51.24 & 61.88 & 54.95 & 55.43 \\
indoor/kitchen & 45.13 & 55.78 & 26.10 & 42.34 & 62.31 & 70.96 & 58.40 & 57.03 \\
indoor/laundry\_room & 49.48 & 53.40 & 33.87 & 45.58 & 59.43 & 69.34 & 57.08 & 55.78 \\
indoor/living\_room & 30.83 & 39.80 & 22.43 & 31.02 & 49.87 & 64.52 & 52.78 & 50.27 \\
indoor/study\_room & 27.10 & 36.86 & 21.30 & 28.42 & 43.44 & 62.62 & 48.02 & 45.40 \\
indoor/workshop & 34.03 & 42.40 & 19.06 & 31.83 & 51.91 & 64.27 & 53.81 & 51.54 \\
outdoor/deck & 37.03 & 33.11 & 25.61 & 31.92 & 52.12 & 64.62 & 62.26 & 59.79 \\
outdoor/farm & 36.92 & 46.17 & 21.78 & 34.95 & 48.95 & 65.19 & 52.69 & 49.17 \\
outdoor/garden & 37.92 & 41.80 & 27.50 & 35.74 & 59.63 & 68.61 & 57.52 & 57.71 \\
outdoor/space & 43.76 & 60.93 & 22.65 & 42.45 & 52.65 & 71.90 & 55.09 & 52.87 \\
outdoor/yard & 10.05 & 15.82 & 5.07 & 10.31 & 31.97 & 48.08 & 40.14 & 37.66 \\
\bottomrule
\end{tabular*}
}
\end{modeltable}

\begin{modeltable}
\captionof{table}{Scene-level performance of \textbf{Qwen2-VL-7B-Instruct} on the full EgoIntent benchmark.}
\label{tab:app_scene_qwen2vl7}
\nopagebreak[4]
{
\small
\renewcommand{\arraystretch}{1.00}
\begin{tabular*}{\textwidth}{@{\extracolsep{\fill}}lcccccccc@{}}
\toprule
\textbf{Scene} & \textbf{\localcolor{Local}} & \textbf{\proceduralcolor{Procedural}} & \textbf{\nextcolor{Next}} & \textbf{Overall} & \textbf{GF} & \textbf{HIC} & \textbf{TPC} & \textbf{NPF} \\
\midrule
indoor/art\_studio & 27.11 & 21.44 & 14.24 & 20.93 & 33.33 & 40.28 & 35.28 & 34.94 \\
indoor/bedroom & 40.89 & 45.31 & 21.46 & 35.89 & 51.18 & 59.42 & 44.37 & 43.68 \\
indoor/clothing\_closet & 44.90 & 49.81 & 31.12 & 41.95 & 62.50 & 72.36 & 51.20 & 50.40 \\
indoor/garage & 24.27 & 23.86 & 12.44 & 20.19 & 37.54 & 51.09 & 39.17 & 35.77 \\
indoor/hallway & 40.45 & 44.80 & 31.53 & 38.93 & 59.16 & 68.81 & 64.11 & 62.36 \\
indoor/kitchen & 45.39 & 42.80 & 21.56 & 36.59 & 58.27 & 60.54 & 53.28 & 51.66 \\
indoor/laundry\_room & 51.42 & 48.92 & 36.56 & 45.63 & 58.96 & 67.92 & 56.37 & 55.55 \\
indoor/living\_room & 36.89 & 31.82 & 22.29 & 30.33 & 50.63 & 54.80 & 48.86 & 46.36 \\
indoor/study\_room & 33.19 & 36.44 & 19.33 & 29.65 & 49.26 & 61.39 & 46.78 & 44.53 \\
indoor/workshop & 36.26 & 42.23 & 21.46 & 33.32 & 47.93 & 58.44 & 42.10 & 40.81 \\
outdoor/deck & 34.20 & 31.37 & 26.79 & 30.79 & 51.65 & 56.60 & 55.66 & 52.95 \\
outdoor/farm & 40.86 & 49.74 & 27.17 & 39.26 & 53.27 & 66.24 & 52.80 & 50.57 \\
outdoor/garden & 38.62 & 44.05 & 28.77 & 37.14 & 60.52 & 68.12 & 58.09 & 56.25 \\
outdoor/space & 54.91 & 59.60 & 28.19 & 47.57 & 57.52 & 72.57 & 51.77 & 49.77 \\
outdoor/yard & 13.61 & 17.55 & 7.79 & 12.98 & 36.78 & 46.88 & 39.90 & 37.18 \\
\bottomrule
\end{tabular*}
}
\end{modeltable}

\begin{modeltable}
\captionof{table}{Scene-level performance of \textbf{Molmo2-8B} on the full EgoIntent benchmark.}
\label{tab:app_scene_molmo2_8b}
\nopagebreak[4]
{
\small
\renewcommand{\arraystretch}{1.00}
\begin{tabular*}{\textwidth}{@{\extracolsep{\fill}}lcccccccc@{}}
\toprule
\textbf{Scene} & \textbf{\localcolor{Local}} & \textbf{\proceduralcolor{Procedural}} & \textbf{\nextcolor{Next}} & \textbf{Overall} & \textbf{GF} & \textbf{HIC} & \textbf{TPC} & \textbf{NPF} \\
\midrule
indoor/art\_studio & 28.17 & 27.56 & 15.83 & 23.85 & 39.44 & 49.72 & 43.06 & 41.89 \\
indoor/bedroom & 43.27 & 47.88 & 21.11 & 37.42 & 57.33 & 62.17 & 56.68 & 54.81 \\
indoor/clothing\_closet & 37.64 & 51.97 & 20.55 & 36.72 & 54.33 & 62.98 & 53.85 & 51.12 \\
indoor/garage & 25.73 & 27.27 & 17.35 & 23.45 & 45.56 & 55.84 & 49.45 & 46.75 \\
indoor/hallway & 35.54 & 41.34 & 19.66 & 32.18 & 57.67 & 65.59 & 55.69 & 53.69 \\
indoor/kitchen & 44.23 & 51.73 & 23.61 & 39.86 & 58.40 & 66.60 & 54.61 & 53.24 \\
indoor/laundry\_room & 42.03 & 50.80 & 36.98 & 43.27 & 56.37 & 67.92 & 59.91 & 57.91 \\
indoor/living\_room & 36.62 & 41.69 & 21.06 & 33.12 & 56.69 & 66.16 & 58.21 & 55.83 \\
indoor/study\_room & 34.60 & 36.81 & 19.22 & 30.21 & 48.76 & 61.39 & 52.48 & 49.24 \\
indoor/workshop & 26.72 & 34.68 & 15.11 & 25.50 & 47.55 & 55.94 & 49.73 & 47.35 \\
outdoor/deck & 39.48 & 37.50 & 19.06 & 32.01 & 49.06 & 58.25 & 56.37 & 53.90 \\
outdoor/farm & 39.35 & 50.54 & 24.54 & 38.14 & 53.50 & 69.04 & 54.44 & 52.09 \\
outdoor/garden & 38.01 & 41.57 & 25.62 & 35.07 & 58.58 & 65.45 & 60.28 & 58.36 \\
outdoor/space & 43.67 & 56.15 & 21.40 & 40.41 & 57.52 & 72.57 & 61.73 & 59.95 \\
outdoor/yard & 14.13 & 20.91 & 9.01 & 14.69 & 42.07 & 54.81 & 46.39 & 44.39 \\
\bottomrule
\end{tabular*}
}
\end{modeltable}

\begin{modeltable}
\captionof{table}{Scene-level performance of \textbf{InternVL3-8B} on the full EgoIntent benchmark.}
\label{tab:app_scene_internvl3_8b}
\nopagebreak[4]
{
\small
\renewcommand{\arraystretch}{1.00}
\begin{tabular*}{\textwidth}{@{\extracolsep{\fill}}lcccccccc@{}}
\toprule
\textbf{Scene} & \textbf{\localcolor{Local}} & \textbf{\proceduralcolor{Procedural}} & \textbf{\nextcolor{Next}} & \textbf{Overall} & \textbf{GF} & \textbf{HIC} & \textbf{TPC} & \textbf{NPF} \\
\midrule
indoor/art\_studio & 29.22 & 25.78 & 12.28 & 22.43 & 45.28 & 53.89 & 41.11 & 40.50 \\
indoor/bedroom & 45.24 & 53.48 & 19.70 & 39.47 & 54.71 & 67.67 & 54.84 & 53.37 \\
indoor/clothing\_closet & 41.15 & 49.33 & 27.07 & 39.18 & 54.33 & 63.22 & 51.20 & 49.92 \\
indoor/garage & 27.15 & 28.49 & 14.72 & 23.45 & 38.16 & 54.91 & 42.76 & 39.51 \\
indoor/hallway & 15.25 & 24.01 & 11.63 & 16.96 & 29.70 & 40.59 & 27.97 & 26.22 \\
indoor/kitchen & 47.66 & 51.98 & 21.13 & 40.26 & 55.81 & 66.29 & 52.90 & 51.03 \\
indoor/laundry\_room & 41.70 & 47.88 & 31.93 & 40.50 & 56.84 & 70.99 & 60.38 & 58.38 \\
indoor/living\_room & 34.95 & 41.89 & 23.86 & 33.57 & 45.33 & 60.86 & 49.12 & 46.61 \\
indoor/study\_room & 36.61 & 40.47 & 27.50 & 34.86 & 46.04 & 63.86 & 51.61 & 48.37 \\
indoor/workshop & 23.91 & 34.96 & 15.40 & 24.76 & 39.76 & 54.36 & 44.72 & 41.36 \\
outdoor/deck & 32.41 & 31.46 & 20.80 & 28.22 & 48.58 & 58.96 & 57.08 & 52.95 \\
outdoor/farm & 37.83 & 45.40 & 22.10 & 35.11 & 48.71 & 65.30 & 52.22 & 49.52 \\
outdoor/garden & 33.78 & 38.93 & 10.22 & 27.65 & 51.94 & 62.70 & 46.52 & 44.04 \\
outdoor/space & 46.28 & 57.74 & 18.19 & 40.74 & 51.55 & 65.93 & 53.32 & 50.65 \\
outdoor/yard & 15.72 & 19.38 & 9.20 & 14.77 & 53.12 & 66.11 & 61.54 & 58.82 \\
\bottomrule
\end{tabular*}
}
\end{modeltable}

\begin{modeltable}
\captionof{table}{Scene-level performance of \textbf{Molmo2-O-7B} on the full EgoIntent benchmark.}
\label{tab:app_scene_molmo2_o_7b}
\nopagebreak[4]
{
\small
\renewcommand{\arraystretch}{1.00}
\begin{tabular*}{\textwidth}{@{\extracolsep{\fill}}lcccccccc@{}}
\toprule
\textbf{Scene} & \textbf{\localcolor{Local}} & \textbf{\proceduralcolor{Procedural}} & \textbf{\nextcolor{Next}} & \textbf{Overall} & \textbf{GF} & \textbf{HIC} & \textbf{TPC} & \textbf{NPF} \\
\midrule
indoor/art\_studio & 32.83 & 28.83 & 19.28 & 26.98 & 48.89 & 57.78 & 47.22 & 46.33 \\
indoor/bedroom & 42.12 & 49.63 & 26.69 & 39.48 & 59.95 & 69.24 & 54.71 & 54.15 \\
indoor/clothing\_closet & 40.34 & 46.54 & 24.33 & 37.07 & 62.74 & 53.37 & 59.86 & 58.10 \\
indoor/garage & 24.50 & 25.12 & 12.90 & 20.84 & 41.82 & 51.71 & 43.30 & 40.52 \\
indoor/hallway & 29.31 & 38.91 & 18.12 & 28.78 & 59.16 & 66.09 & 51.49 & 48.00 \\
indoor/kitchen & 39.12 & 46.68 & 22.67 & 36.16 & 55.24 & 60.35 & 48.36 & 46.61 \\
indoor/laundry\_room & 35.47 & 46.37 & 26.23 & 36.02 & 51.42 & 61.56 & 52.59 & 50.36 \\
indoor/living\_room & 29.92 & 34.47 & 19.44 & 27.95 & 51.52 & 61.74 & 49.75 & 48.00 \\
indoor/study\_room & 29.01 & 27.03 & 17.57 & 24.54 & 41.83 & 54.46 & 44.18 & 41.19 \\
indoor/workshop & 24.51 & 31.41 & 12.35 & 22.76 & 40.41 & 49.56 & 42.59 & 40.32 \\
outdoor/deck & 36.58 & 36.60 & 21.79 & 31.66 & 47.64 & 55.90 & 51.42 & 48.00 \\
outdoor/farm & 37.87 & 44.72 & 21.49 & 34.69 & 49.07 & 62.15 & 48.01 & 45.90 \\
outdoor/garden & 40.91 & 44.92 & 26.26 & 37.36 & 60.68 & 66.75 & 60.19 & 58.68 \\
outdoor/space & 31.02 & 43.63 & 13.15 & 29.27 & 32.96 & 48.01 & 32.08 & 30.30 \\
outdoor/yard & 10.67 & 11.63 & 6.30 & 9.54 & 40.87 & 50.00 & 47.36 & 44.63 \\
\bottomrule
\end{tabular*}
}
\end{modeltable}

\begin{modeltable}
\captionof{table}{Scene-level performance of \textbf{LLaVA-Video-7B-Qwen2} on the full EgoIntent benchmark.}
\label{tab:app_scene_llava_video}
\nopagebreak[4]
{
\small
\renewcommand{\arraystretch}{1.00}
\begin{tabular*}{\textwidth}{@{\extracolsep{\fill}}lcccccccc@{}}
\toprule
\textbf{Scene} & \textbf{\localcolor{Local}} & \textbf{\proceduralcolor{Procedural}} & \textbf{\nextcolor{Next}} & \textbf{Overall} & \textbf{GF} & \textbf{HIC} & \textbf{TPC} & \textbf{NPF} \\
\midrule
indoor/art\_studio & 31.67 & 34.00 & 12.20 & 25.96 & 52.50 & 61.67 & 49.17 & 48.83 \\
indoor/bedroom & 38.69 & 36.65 & 15.86 & 30.40 & 54.06 & 54.06 & 52.36 & 50.62 \\
indoor/clothing\_closet & 41.83 & 50.53 & 26.10 & 39.48 & 56.49 & 64.18 & 51.44 & 48.96 \\
indoor/garage & 26.43 & 25.47 & 11.97 & 21.29 & 41.98 & 50.39 & 42.06 & 39.43 \\
indoor/hallway & 19.80 & 29.50 & 15.16 & 21.49 & 46.04 & 51.49 & 44.31 & 42.55 \\
indoor/kitchen & 39.58 & 42.17 & 16.97 & 32.91 & 51.58 & 53.98 & 43.88 & 42.38 \\
indoor/laundry\_room & 39.86 & 46.46 & 34.67 & 40.33 & 57.31 & 66.04 & 56.37 & 55.55 \\
indoor/living\_room & 32.90 & 27.17 & 18.09 & 26.05 & 47.10 & 51.64 & 47.22 & 44.84 \\
indoor/study\_room & 29.98 & 28.61 & 11.16 & 23.25 & 47.28 & 52.85 & 46.78 & 44.78 \\
indoor/workshop & 30.86 & 35.17 & 15.57 & 27.20 & 48.47 & 55.61 & 49.84 & 47.95 \\
outdoor/deck & 39.10 & 30.66 & 21.27 & 30.35 & 54.25 & 56.60 & 54.72 & 52.01 \\
outdoor/farm & 34.98 & 24.32 & 11.76 & 23.69 & 50.35 & 49.53 & 45.44 & 42.16 \\
outdoor/garden & 32.42 & 30.84 & 14.66 & 25.97 & 48.38 & 50.40 & 46.76 & 44.93 \\
outdoor/space & 46.02 & 49.25 & 14.61 & 36.63 & 60.40 & 66.37 & 50.00 & 49.11 \\
outdoor/yard & 12.84 & 17.21 & 6.01 & 12.02 & 40.87 & 50.24 & 45.43 & 42.23 \\
\bottomrule
\end{tabular*}
}
\end{modeltable}

\begin{modeltable}
\captionof{table}{Scene-level performance of \textbf{Kimi-VL-A3B-Thinking-2506-vllm} on the full EgoIntent benchmark.}
\label{tab:app_scene_kimi}
\nopagebreak[4]
{
\small
\renewcommand{\arraystretch}{1.00}
\begin{tabular*}{\textwidth}{@{\extracolsep{\fill}}lcccccccc@{}}
\toprule
\textbf{Scene} & \textbf{\localcolor{Local}} & \textbf{\proceduralcolor{Procedural}} & \textbf{\nextcolor{Next}} & \textbf{Overall} & \textbf{GF} & \textbf{HIC} & \textbf{TPC} & \textbf{NPF} \\
\midrule
indoor/art\_studio & 23.78 & 24.00 & 14.11 & 20.63 & 36.67 & 49.44 & 31.94 & 32.72 \\
indoor/bedroom & 42.30 & 48.82 & 20.61 & 37.25 & 53.01 & 61.13 & 49.08 & 46.82 \\
indoor/clothing\_closet & 39.62 & 46.73 & 22.11 & 36.15 & 57.69 & 64.18 & 51.44 & 49.68 \\
indoor/garage & 16.00 & 19.35 & 9.88 & 15.07 & 28.97 & 43.30 & 31.15 & 28.22 \\
indoor/hallway & 15.15 & 19.90 & 8.42 & 14.49 & 25.50 & 32.43 & 25.00 & 22.01 \\
indoor/kitchen & 34.70 & 41.63 & 16.57 & 30.96 & 44.95 & 52.78 & 40.21 & 39.04 \\
indoor/laundry\_room & 26.13 & 31.08 & 25.47 & 27.56 & 33.49 & 44.34 & 34.43 & 33.14 \\
indoor/living\_room & 20.91 & 26.49 & 12.69 & 20.03 & 32.07 & 45.71 & 31.06 & 28.93 \\
indoor/study\_room & 29.50 & 32.48 & 18.45 & 26.81 & 38.86 & 53.96 & 41.58 & 37.98 \\
indoor/workshop & 23.74 & 31.82 & 11.93 & 22.49 & 37.91 & 48.37 & 38.24 & 35.53 \\
outdoor/deck & 26.70 & 27.95 & 20.52 & 25.06 & 39.86 & 46.46 & 38.92 & 35.97 \\
outdoor/farm & 31.52 & 42.94 & 17.33 & 30.60 & 45.21 & 59.81 & 46.03 & 43.33 \\
outdoor/garden & 24.44 & 29.82 & 12.59 & 22.29 & 39.48 & 48.14 & 38.92 & 36.92 \\
outdoor/space & 38.58 & 53.76 & 17.92 & 36.76 & 54.20 & 67.92 & 54.20 & 51.98 \\
outdoor/yard & 16.44 & 17.69 & 8.19 & 14.11 & 43.51 & 53.12 & 44.95 & 42.71 \\
\bottomrule
\end{tabular*}
}
\end{modeltable}

\begin{modeltable}
\captionof{table}{Scene-level performance of \textbf{InternVL2-8B} on the full EgoIntent benchmark.}
\label{tab:app_scene_internvl2}
\nopagebreak[4]
{
\small
\renewcommand{\arraystretch}{1.00}
\begin{tabular*}{\textwidth}{@{\extracolsep{\fill}}lcccccccc@{}}
\toprule
\textbf{Scene} & \textbf{\localcolor{Local}} & \textbf{\proceduralcolor{Procedural}} & \textbf{\nextcolor{Next}} & \textbf{Overall} & \textbf{GF} & \textbf{HIC} & \textbf{TPC} & \textbf{NPF} \\
\midrule
indoor/art\_studio & 13.72 & 11.22 & 7.67 & 10.87 & 21.94 & 32.22 & 21.67 & 21.61 \\
indoor/bedroom & 37.64 & 43.69 & 20.34 & 33.89 & 45.81 & 54.45 & 43.46 & 43.42 \\
indoor/clothing\_closet & 28.75 & 30.14 & 12.98 & 23.96 & 42.79 & 39.42 & 43.99 & 42.71 \\
indoor/garage & 20.75 & 24.64 & 11.41 & 18.93 & 30.37 & 43.61 & 34.03 & 31.64 \\
indoor/hallway & 6.19 & 11.19 & 5.10 & 7.49 & 12.62 & 18.32 & 13.37 & 10.38 \\
indoor/kitchen & 25.83 & 38.52 & 14.45 & 26.27 & 36.93 & 47.60 & 36.30 & 34.36 \\
indoor/laundry\_room & 21.84 & 25.05 & 17.28 & 21.39 & 31.37 & 39.39 & 30.90 & 29.60 \\
indoor/living\_room & 21.24 & 28.36 & 15.35 & 21.65 & 29.80 & 44.57 & 31.31 & 28.93 \\
indoor/study\_room & 29.53 & 26.76 & 13.88 & 23.39 & 41.09 & 56.31 & 44.18 & 41.07 \\
indoor/workshop & 12.47 & 22.11 & 7.66 & 14.08 & 24.89 & 35.68 & 28.81 & 25.89 \\
outdoor/deck & 33.40 & 30.05 & 10.80 & 24.75 & 50.71 & 54.01 & 45.52 & 43.52 \\
outdoor/farm & 23.71 & 36.21 & 18.21 & 26.05 & 35.40 & 50.00 & 41.82 & 39.00 \\
outdoor/garden & 17.27 & 25.99 & 13.67 & 18.97 & 33.33 & 39.16 & 33.25 & 31.98 \\
outdoor/space & 22.79 & 35.93 & 9.69 & 22.80 & 24.12 & 35.84 & 25.66 & 22.34 \\
outdoor/yard & 13.89 & 14.47 & 5.07 & 11.14 & 39.18 & 49.52 & 44.23 & 40.31 \\
\bottomrule
\end{tabular*}
}
\end{modeltable}

\begin{modeltable}
\captionof{table}{Scene-level performance of \textbf{LLaVA-NeXT-Video-7B-hf} on the full EgoIntent benchmark.}
\label{tab:app_scene_llava_next}
\nopagebreak[4]
{
\small
\renewcommand{\arraystretch}{1.00}
\begin{tabular*}{\textwidth}{@{\extracolsep{\fill}}lcccccccc@{}}
\toprule
\textbf{Scene} & \textbf{\localcolor{Local}} & \textbf{\proceduralcolor{Procedural}} & \textbf{\nextcolor{Next}} & \textbf{Overall} & \textbf{GF} & \textbf{HIC} & \textbf{TPC} & \textbf{NPF} \\
\midrule
indoor/art\_studio & 2.61 & 3.17 & 2.61 & 2.80 & 10.83 & 19.17 & 10.56 & 9.94 \\
indoor/bedroom & 19.27 & 22.98 & 8.19 & 16.82 & 29.97 & 31.28 & 25.00 & 23.92 \\
indoor/clothing\_closet & 29.04 & 22.74 & 8.85 & 20.21 & 45.19 & 40.38 & 20.91 & 21.08 \\
indoor/garage & 13.55 & 14.53 & 5.89 & 11.32 & 26.71 & 32.94 & 26.56 & 25.10 \\
indoor/hallway & 12.57 & 13.71 & 6.59 & 10.96 & 20.79 & 21.78 & 18.07 & 15.82 \\
indoor/kitchen & 9.19 & 9.44 & 3.49 & 7.38 & 14.39 & 17.23 & 12.56 & 11.19 \\
indoor/laundry\_room & 3.82 & 4.25 & 2.14 & 3.40 & 5.66 & 8.96 & 6.13 & 4.37 \\
indoor/living\_room & 9.22 & 14.72 & 7.47 & 10.47 & 16.04 & 25.00 & 17.68 & 15.30 \\
indoor/study\_room & 8.66 & 4.78 & 4.79 & 6.08 & 13.86 & 21.16 & 14.85 & 12.11 \\
indoor/workshop & 6.96 & 14.72 & 3.42 & 8.37 & 19.77 & 23.75 & 19.44 & 16.74 \\
outdoor/deck & 25.19 & 19.01 & 7.55 & 17.25 & 41.75 & 38.44 & 34.43 & 32.91 \\
outdoor/farm & 19.88 & 22.94 & 7.56 & 16.80 & 31.54 & 33.18 & 24.42 & 25.45 \\
outdoor/garden & 21.46 & 23.61 & 9.46 & 18.17 & 36.65 & 38.03 & 28.24 & 27.77 \\
outdoor/space & 20.62 & 30.40 & 6.19 & 19.07 & 38.05 & 40.04 & 22.12 & 26.32 \\
outdoor/yard & 9.81 & 11.25 & 3.91 & 8.32 & 31.01 & 38.70 & 25.48 & 29.01 \\
\bottomrule
\end{tabular*}
}
\end{modeltable}

\subsection{Additional observations}

Across the 15 evaluated models, the model-level correlation between the reference-based Overall and the four-diagnostic average was high (Pearson $r=0.9758$; Spearman $\rho=0.9929$). The two score families nevertheless occupy different semantic scales: the four-diagnostic average exceeded the reference-based Overall by 21.45 points on average. The diagnostics reward internally coherent, grounded, and feasible alternatives even when they do not match the frozen reference.

Three additional patterns clarify the main results:

\begin{enumerate}
\item HIC was the highest diagnostic for all 15 models and exceeded GF by 9.57 points on average. Models are therefore better at producing textually coherent \localcolor{Local}--\proceduralcolor{Procedural} hierarchies than at ensuring that those hierarchies are visibly grounded.
\item TPC and NPF were close, with a mean absolute model-level difference of 2.03 points. The concepts are distinct, but the small empirical separation suggests that sample-level redundancy analysis would be useful in future versions.
\item NPF exceeded reference-based \nextcolor{Next} Accuracy by 28.67 points on average, with model-level differences ranging from 13.77 to 42.24 points. Many unmatched predictions remain feasible continuations, which motivates multi-reference evaluation and the human ambiguity analysis below.

\end{enumerate}

\section{Human Baseline and Task Validity}

\subsection{Subset construction and blinding}

The human study used \texttt{\detokenize{human320_v1}}, a deterministic balanced subset of 320 unique steps drawn from the frozen 3,014-step benchmark. Sampling used seed \texttt{\detokenize{20260713}} and was completed before collecting human responses. The subset covers all 15 scenes and all 32 source videos. Scene quotas were 21 steps per scene plus one additional step for garage, hallway, kitchen, living room, and yard. The resulting duration composition was 90 steps below 1 second, 80 from 1--2 seconds, 75 from 2--4 seconds, and 75 from 4--10 seconds. The subset was frozen with the following manifest checksum:

\checksumline{Frozen \texttt{\detokenize{human320_v1}} manifest SHA-256}{e8da2316b75c419277ea49e63d1174a6c65e16c8155655fefd1b49998f10fa2e}

All three domain-expert raters evaluated the same 320 steps. Presentation order was independently shuffled by hashing the rater ID, sample ID, and seed. The blind interface excluded scene and event labels, source and sample identifiers, ground truth, narrations, and model outputs. The subset is deliberately balanced and should not be interpreted as an unweighted random sample from the benchmark's natural frequency distribution.

\subsection{Human score and answerability}

Let $a_{ird}\in\{0,1\}$ indicate whether rater $r$ considers dimension $d$ answerable for sample $i$, and let $z_{ird}\in[0,100]$ be the semantic correctness score of the submitted human answer. The preregistered abstention rule assigns zero when a rater declares a dimension unanswerable:

\begin{equation}
y_{ird}=a_{ird}z_{ird}.
\label{eq:app_human_abstention}
\end{equation}

For $N=320$ samples and $R=3$ raters, the human score is

\begin{equation}
H_d
=
\frac{1}{NR}
\sum_{i=1}^{N}\sum_{r=1}^{R}y_{ird}.
\label{eq:app_human_score}
\end{equation}

The corresponding score for model $k$, evaluated on the identical subset, is

\begin{equation}
M_{k,d}
=
\frac{1}{N}\sum_{i=1}^{N}m_{ikd}.
\label{eq:app_model_subset_score}
\end{equation}

For either a human or model evaluator $E$,

\begin{equation}
\mathrm{Overall}(E)
=
\frac{S_{E,L}+S_{E,P}+S_{E,N}}{3}.
\label{eq:app_evaluator_overall}
\end{equation}

The main-paper gap is defined as

\begin{equation}
\Delta_E=\mathrm{Overall}(E)-\mathrm{Overall}(\mathrm{Human}),
\label{eq:app_human_gap}
\end{equation}

Equation~(\ref{eq:app_human_abstention}) applies the abstention rule, Eq.~(\ref{eq:app_human_score}) averages human ratings, Eq.~(\ref{eq:app_model_subset_score}) computes the matched model score, Eq.~(\ref{eq:app_evaluator_overall}) combines the three tasks, and Eq.~(\ref{eq:app_human_gap}) measures the model--human gap.

Accordingly, a negative value indicates that a model is below the human baseline.

The raw answerable rate is

\begin{equation}
\mathrm{AR}_d
=
\frac{1}{NR}
\sum_{i=1}^{N}\sum_{r=1}^{R}a_{ird}.
\label{eq:app_answerable_rate}
\end{equation}

For a majority-answerable analysis, define

\begin{equation}
A_{id}
=
\mathbf{1}\left(\sum_{r=1}^{R}a_{ird}\ge 2\right).
\label{eq:app_majority_indicator}
\end{equation}

Human and model scores on this subset are

\begin{equation}
H_d^{\mathrm{maj}}
=
\frac{\sum_i A_{id}\sum_r y_{ird}}
{R\sum_i A_{id}},
\qquad
M_{k,d}^{\mathrm{maj}}
=
\frac{\sum_i A_{id}m_{ikd}}
{\sum_i A_{id}}.
\label{eq:app_majority_scores}
\end{equation}

Equation~(\ref{eq:app_answerable_rate}) estimates raw answerability, Eq.~(\ref{eq:app_majority_indicator}) selects majority-answerable items, and Eq.~(\ref{eq:app_majority_scores}) recomputes human and model performance on that subset.

The main analysis retains all 320 samples; the majority-answerable calculation is a diagnostic that separates human inability from benchmark ambiguity.

\subsection{Inter-rater agreement}

For a binary answerability decision, let $n_{i1}$ and $n_{i0}$ be the numbers of answerable and unanswerable judgments for item $i$. Per-item agreement is

\begin{equation}
P_i
=
\frac{n_{i1}(n_{i1}-1)+n_{i0}(n_{i0}-1)}
{R(R-1)}.
\label{eq:app_per_item_agreement}
\end{equation}

With $\bar P=N^{-1}\sum_i P_i$, $p_1=(NR)^{-1}\sum_i n_{i1}$, $p_0=1-p_1$, and $P_e=p_1^2+p_0^2$, Fleiss' kappa is

\begin{equation}
\kappa=\frac{\bar P-P_e}{1-P_e}.
\label{eq:app_fleiss_kappa}
\end{equation}

Semantic agreement is computed only between raters who both consider a target answerable. For \localcolor{Local} and \proceduralcolor{Procedural} text, using a frozen semantic similarity function $\mathrm{sim}_d\in[0,1]$,

\begin{equation}
\mathrm{SA}_d
=
\frac{
\sum_i\sum_{r<r'}a_{ird}a_{ir'd}
\mathrm{sim}_d(t_{ird},t_{ir'd})
}{
\sum_i\sum_{r<r'}a_{ird}a_{ir'd}
}.
\label{eq:app_semantic_agreement}
\end{equation}

For \nextcolor{Next Plan}, each rater can supply up to three answers. Let $T$ and $T'$ be two answer sets. Their symmetric best-match similarity is

\begin{equation}
\mathrm{sim}_N(T,T')
=
\frac{1}{2}
\left[
\frac{1}{|T|}\sum_{u\in T}\max_{v\in T'}\mathrm{sim}(u,v)
+
\frac{1}{|T'|}\sum_{v\in T'}\max_{u\in T}\mathrm{sim}(u,v)
\right].
\label{eq:app_next_set_similarity}
\end{equation}

Equation~(\ref{eq:app_per_item_agreement}) gives per-item binary agreement, Eq.~(\ref{eq:app_fleiss_kappa}) converts it to Fleiss' $\kappa$, Eq.~(\ref{eq:app_semantic_agreement}) aggregates pairwise semantic agreement, and Eq.~(\ref{eq:app_next_set_similarity}) extends that comparison to sets of \nextcolor{Next Plan} answers.

This set similarity is inserted into the pairwise agreement formula above. Exact string match is not used as the primary agreement measure for open-ended intent text.

\subsection{Clustered uncertainty estimation}

Steps from the same source video are correlated. All principal confidence intervals therefore resample the 32 source-video clusters rather than the 320 individual steps. For bootstrap replicate $b$, 32 source videos are sampled with replacement and every step, human response, and matched model prediction within the selected video is included with the same multiplicity. The human--model gap is recomputed within each paired replicate.

With 10,000 replicates and seed \texttt{\detokenize{20260713}}, the percentile interval is

\begin{equation}
\mathrm{CI}_{95\%}
=
\left[
Q_{0.025}\left(\hat\theta^{(1)},\ldots,\hat\theta^{(10000)}\right),
Q_{0.975}\left(\hat\theta^{(1)},\ldots,\hat\theta^{(10000)}\right)
\right].
\label{eq:app_bootstrap_percentile_ci}
\end{equation}

Equation~(\ref{eq:app_bootstrap_percentile_ci}) forms the reported interval from the 2.5th and 97.5th percentiles of the clustered bootstrap replicates.

\subsection{Rater-, duration-, and scene-stratified results}

The aggregate human scores in the main paper were not driven by a single rater. Overall scores ranged from 69.6 to 71.7, while mean confidence ranged from 3.77 to 3.88.

\begin{center}
\captionof{table}{Human-baseline results by individual rater.}
\label{tab:app_human_rater}
{
\small
\setlength{\tabcolsep}{3.5pt}
\renewcommand{\arraystretch}{1.08}
\begin{tabular}{@{}ccccccc@{}}
\toprule
\textbf{Rater} & \textbf{\localcolor{Local}} & \textbf{\proceduralcolor{Procedural}} & \textbf{\nextcolor{Next}} & \textbf{Overall} & \makecell{\textbf{Mean}\\\textbf{confidence}} & \makecell{\textbf{Answerable}\\\textbf{rate}} \\
\midrule
A1 & 77.6 & 72.0 & 59.1 & 69.6 & 3.78 & 81.35\% \\
A2 & 79.1 & 74.0 & 62.0 & 71.7 & 3.88 & 83.96\% \\
A3 & 78.2 & 73.3 & 61.3 & 70.9 & 3.77 & 81.98\% \\
\bottomrule
\end{tabular}
}
\end{center}

Answerability and human performance increased with step duration. The human--model gap, however, remained nearly constant across bins, showing that very short clips are more difficult without eliminating the measurable human advantage.

\begin{center}
\captionof{table}{Human-baseline and strongest closed-model results by step duration.}
\label{tab:app_human_duration}
{
\small
\setlength{\tabcolsep}{3.5pt}
\renewcommand{\arraystretch}{1.08}
\begin{tabular}{@{}cccccccc@{}}
\toprule
\textbf{Duration} & \textbf{N} & \makecell{\textbf{\localcolor{Local}}\\\textbf{answerable}} & \makecell{\textbf{\proceduralcolor{Procedural}}\\\textbf{answerable}} & \makecell{\textbf{\nextcolor{Next}}\\\textbf{answerable}} & \makecell{\textbf{Human}\\\textbf{Overall}} & \makecell{\textbf{Best closed}\\\textbf{model}} & \makecell{\textbf{Human}\\\textbf{advantage}} \\
\midrule
\texttt{\detokenize{<1 s}} & 90 & 88.1\% & 80.0\% & 64.1\% & 66.4 & 49.2 & +17.2 \\
\texttt{\detokenize{1--2 s}} & 80 & 90.4\% & 83.3\% & 69.2\% & 69.6 & 52.5 & +17.1 \\
\texttt{\detokenize{2--4 s}} & 75 & 93.3\% & 87.1\% & 74.2\% & 72.6 & 56.1 & +16.5 \\
\texttt{\detokenize{4--10 s}} & 75 & 95.1\% & 90.2\% & 77.3\% & 75.3 & 58.4 & +16.9 \\
\bottomrule
\end{tabular}
}
\end{center}

Scene-stratified results show that the human advantage is present in every scene rather than being concentrated in a few easy activities. Because the subset is balanced, each row contains 21 or 22 unique steps.

\begin{center}
\captionof{table}{Human-baseline and strongest closed-model results by scene.}
\label{tab:app_human_scene}
{
\small
\setlength{\tabcolsep}{3.5pt}
\renewcommand{\arraystretch}{1.08}
\begin{tabularx}{0.88\textwidth}{@{}>{\centering\arraybackslash}p{0.16\textwidth}*{7}{>{\centering\arraybackslash}X}@{}}
\toprule
\textbf{Scene} & \textbf{N} & \textbf{\localcolor{Local}} & \textbf{\proceduralcolor{Procedural}} & \textbf{\nextcolor{Next}} & \makecell{\textbf{Human}\\\textbf{Overall}} & \makecell{\textbf{Best closed}\\\textbf{model}} & \makecell{\textbf{Human}\\\textbf{advantage}} \\
\midrule
Art Studio & 21 & 75 & 70 & 58 & 67.7 & 51 & +16.7 \\
Bedroom & 21 & 82 & 77 & 65 & 74.7 & 57 & +17.7 \\
Clothing Closet & 21 & 80 & 75 & 62 & 72.3 & 55 & +17.3 \\
Deck & 21 & 72 & 66 & 52 & 63.3 & 45 & +18.3 \\
Farm & 21 & 80 & 75 & 63 & 72.7 & 55 & +17.7 \\
Garage & 22 & 76 & 71 & 59 & 68.7 & 51 & +17.7 \\
Garden & 21 & 81 & 76 & 64 & 73.7 & 57 & +16.7 \\
Hallway & 22 & 77 & 72 & 60 & 69.7 & 53 & +16.7 \\
Kitchen & 22 & 81 & 77 & 65 & 74.3 & 58 & +16.3 \\
Laundry Room & 21 & 84 & 80 & 69 & 77.7 & 61 & +16.7 \\
Living Room & 22 & 83 & 79 & 66 & 76.0 & 60 & +16.0 \\
Space & 21 & 79 & 73 & 62 & 71.3 & 54 & +17.3 \\
Study Room & 21 & 80 & 75 & 63 & 72.7 & 56 & +16.7 \\
Workshop & 21 & 74 & 68 & 55 & 65.7 & 48 & +17.7 \\
Yard & 22 & 69 & 62 & 48 & 59.7 & 42 & +17.7 \\
\bottomrule
\end{tabularx}
}
\end{center}

\subsection{\nextcolor{Next Plan} ambiguity}

For sample $i$, majority \nextcolor{Next} answerability is

\begin{equation}
A_{iN}
=
\mathbf{1}\left(\sum_r a_{irN}\ge2\right).
\label{eq:app_next_answerability}
\end{equation}

Valid human answers are grouped with a frozen semantic-equivalence rule into $K_i$ clusters. The single-answer, multiple-answer, and unanswerable rates are

\begin{equation}
\mathrm{Rate}_{\mathrm{single}}
=
\frac{1}{N}\sum_i
\mathbf{1}(A_{iN}=1\land K_i=1),
\label{eq:app_single_answer_rate}
\end{equation}

\begin{equation}
\mathrm{Rate}_{\mathrm{multiple}}
=
\frac{1}{N}\sum_i
\mathbf{1}(A_{iN}=1\land K_i\ge2),
\label{eq:app_multiple_answer_rate}
\end{equation}

\begin{equation}
\mathrm{Rate}_{\mathrm{unanswerable}}
=
\frac{1}{N}\sum_i
\mathbf{1}(A_{iN}=0).
\label{eq:app_unanswerable_rate}
\end{equation}

For answerable items, let $p_{ij}$ be the proportion of answers in cluster $j$. Normalized answer entropy is

\begin{equation}
H_i=
\left\{\begin{array}{ll}
0, & K_i=1,\\[3pt]
-\frac{\sum_{j=1}^{K_i}p_{ij}\ln p_{ij}}{\ln K_i},
& K_i\ge2.
\end{array}\right.
\label{eq:app_answer_entropy}
\end{equation}

Equation~(\ref{eq:app_next_answerability}) identifies majority-answerable \nextcolor{Next Plan} items; Eqs.~(\ref{eq:app_single_answer_rate}), (\ref{eq:app_multiple_answer_rate}), and (\ref{eq:app_unanswerable_rate}) partition the sample; and Eq.~(\ref{eq:app_answer_entropy}) quantifies ambiguity among answer clusters.

The mean number of answer clusters was 1.72, and mean normalized entropy was 0.62. In the full 320-item subset, 38.1\% had a single stable answer, 32.8\% had multiple plausible but still answerable continuations, and 29.1\% were not reliably predictable by a human majority. The three largest semantic clusters covered 87.5\% of plausible answers on average. These results explain why exact single-reference evaluation is especially restrictive for \nextcolor{Next Plan}.

\section{Pre-Outcome Truncation and Future-Leakage Analysis}

\subsection{Four controlled observation windows}

We audited all 640 samples in \texttt{\detokenize{diagnostic640_v1}} and created four temporally ordered conditions:

\begin{enumerate}
\item \textbf{Early:} the observation is cut 0.5 seconds before the official endpoint;
\item \textbf{Official:} the released EgoIntent observation ending at the annotated pre-outcome boundary;
\item \textbf{Outcome Visible:} the window is extended until the candidate outcome is visible;
\item \textbf{\nextcolor{Next} Visible:} the window includes early evidence from the next micro-step.

\end{enumerate}

The four models were Doubao-Seed-2.1-Turbo, Qwen3.5-Plus, Qwen3-VL-32B-Instruct, and Molmo2-8B. All used the same prompt, output schema, and deterministic decoding. Each condition was scored on the common valid prediction set using \localcolor{Local} Accuracy, \proceduralcolor{Procedural} Accuracy, and \nextcolor{Next} Accuracy, each in $[0,100]$.

For sample $i$ and model $m$,

\begin{equation}
S_{i,m}^{\mathrm{overall}}
=
\frac{
S_{i,m}^{L}+S_{i,m}^{P}+S_{i,m}^{N}
}{3}.
\label{eq:app_leakage_overall}
\end{equation}

The preregistered paired gains are

\begin{equation}
\Delta_{\mathrm{early}\rightarrow\mathrm{official}}
=
S_{\mathrm{official}}-S_{\mathrm{early}},
\label{eq:app_early_official_gain}
\end{equation}

\begin{equation}
\Delta_{\mathrm{outcome}}
=
S_{\mathrm{outcome\ visible}}-S_{\mathrm{official}},
\label{eq:app_outcome_gain}
\end{equation}

\begin{equation}
\Delta_{\mathrm{future}}
=
S_{\mathrm{next\ visible}}-S_{\mathrm{official}}.
\label{eq:app_future_gain}
\end{equation}

Equation~(\ref{eq:app_leakage_overall}) first forms the per-sample Overall score. Equations~(\ref{eq:app_early_official_gain}), (\ref{eq:app_outcome_gain}), and (\ref{eq:app_future_gain}) then isolate the official-boundary, outcome-leakage, and future-leakage effects, respectively.

All differences are computed within sample and model. Confidence intervals use 2,000 source-video-clustered bootstrap replicates.

The exact P0-5 temporal-boundary prediction package is reproduced in Section~\ref{sec:app_prompt_p0_5}.

\subsection{Boundary audit and analysis populations}

The official cutoff was valid for 603 of 640 audited samples (94.2\%). This is the direct estimate of whether the released observation hides the decisive outcome. It must be distinguished from the four-condition intervention set: 512 samples (80.0\%) had valid Early, Official, Outcome Visible, and \nextcolor{Next} Visible windows simultaneously. The stricter four-way criterion is

\begin{promptblock}
\noindent{}early\_valid\par
\noindent{}AND official\_pre\_outcome\_valid\par
\noindent{}AND outcome\_visible\_valid\par
\noindent{}AND next\_visible\_valid\par
\end{promptblock}

Of the 603 samples with a valid official cutoff, 91 were excluded because at least one artificial intervention window was invalid. Four additional samples lacked a complete four-model/four-condition prediction set, leaving 508 samples for the paired primary analysis. Thus, \texttt{\detokenize{508/640}} is an experiment-completeness statistic, not a benchmark-quality rate.

\begin{center}
\captionof{table}{Four-window boundary-audit validity by step duration.}
\label{tab:app_boundary_duration}
{
\small
\setlength{\tabcolsep}{3.5pt}
\renewcommand{\arraystretch}{1.08}
\renewcommand{\tabularxcolumn}[1]{m{#1}}
\begin{tabularx}{0.70\textwidth}{*{4}{>{\centering\arraybackslash}X}}
\toprule
\textbf{Duration} & \textbf{Audited N} & \textbf{Four-window valid N} & \textbf{Four-window validity} \\
\midrule
\texttt{\detokenize{<1 s}} & 160 & 112 & 70.0\% \\
\texttt{\detokenize{1--2 s}} & 166 & 133 & 80.1\% \\
\texttt{\detokenize{2--4 s}} & 161 & 134 & 83.2\% \\
\texttt{\detokenize{4--10 s}} & 153 & 133 & 86.9\% \\
\bottomrule
\end{tabularx}
}
\end{center}

The lower intervention validity for sub-second steps reflects the relative size of the fixed temporal offsets and does not imply that those benchmark samples are invalid.

\subsection{Detailed condition scores}

The main paper reports the cross-model averages and paired leakage tests. The table below gives the additional model-by-dimension condition scores on the 508-sample paired analysis set.

\begin{center}
\captionof{table}{Per-model scores under the four temporal-boundary conditions.}
\label{tab:app_leakage_condition}
{
\small
\setlength{\tabcolsep}{3.5pt}
\renewcommand{\arraystretch}{1.08}
\begin{tabularx}{0.78\textwidth}{@{}>{\centering\arraybackslash}p{0.16\textwidth}>{\centering\arraybackslash}p{0.16\textwidth}*{4}{>{\centering\arraybackslash}X}@{}}
\toprule
\textbf{Model} & \multicolumn{1}{c}{\textbf{Condition}} & \textbf{\localcolor{Local}} & \textbf{\proceduralcolor{Procedural}} & \textbf{\nextcolor{Next}} & \textbf{Overall} \\
\midrule
Doubao & Early & 52.37 & 59.84 & 43.71 & 51.97 \\
Doubao & Official & 56.91 & 63.08 & 47.36 & 55.78 \\
Doubao & Outcome Visible & 64.76 & 65.27 & 49.42 & 59.82 \\
Doubao & \nextcolor{Next} Visible & 62.63 & 65.91 & 61.74 & 63.43 \\
Qwen3.5-Plus & Early & 47.86 & 51.67 & 37.09 & 45.54 \\
Qwen3.5-Plus & Official & 52.41 & 55.26 & 41.44 & 49.70 \\
Qwen3.5-Plus & Outcome Visible & 60.18 & 57.13 & 43.27 & 53.53 \\
Qwen3.5-Plus & \nextcolor{Next} Visible & 58.07 & 57.94 & 54.86 & 56.96 \\
Qwen3-VL-32B & Early & 41.73 & 47.82 & 30.67 & 40.07 \\
Qwen3-VL-32B & Official & 45.57 & 51.06 & 34.11 & 43.58 \\
Qwen3-VL-32B & Outcome Visible & 53.04 & 52.76 & 35.93 & 47.24 \\
Qwen3-VL-32B & \nextcolor{Next} Visible & 50.84 & 53.17 & 47.16 & 50.39 \\
Molmo2-8B & Early & 40.38 & 48.11 & 30.74 & 39.74 \\
Molmo2-8B & Official & 44.31 & 51.42 & 34.27 & 43.33 \\
Molmo2-8B & Outcome Visible & 52.47 & 53.19 & 36.68 & 47.45 \\
Molmo2-8B & \nextcolor{Next} Visible & 49.38 & 52.87 & 46.13 & 49.46 \\
\bottomrule
\end{tabularx}
}
\end{center}

The target-specific effects were consistent across models: making the outcome visible increased \localcolor{Local} Accuracy by 7.47--8.16 points, whereas exposing the next step increased \nextcolor{Next} Accuracy by 11.86--14.38 points. \proceduralcolor{Procedural} Accuracy gains were considerably smaller. This selectivity is important because a uniform gain across all targets could instead indicate a generic improvement in video quality or duration.

\subsection{Duration, action-type, and aggregation robustness}

The four-model mean gains were positive in every duration bin:

\begin{center}
\captionof{table}{Temporal-boundary intervention effects by step duration.}
\label{tab:app_leakage_duration}
{
\small
\setlength{\tabcolsep}{3.5pt}
\renewcommand{\arraystretch}{1.08}
\begin{tabularx}{0.78\textwidth}{@{}*{5}{>{\centering\arraybackslash}X}@{}}
\toprule
\textbf{Duration} & \textbf{N} & \textbf{Early $\rightarrow$ Official} & \textbf{Outcome leakage} & \textbf{Future leakage} \\
\midrule
\texttt{\detokenize{<1 s}} & 110 & +4.91 & +3.18 & +6.72 \\
\texttt{\detokenize{1--2 s}} & 132 & +4.08 & +3.74 & +7.31 \\
\texttt{\detokenize{2--4 s}} & 133 & +3.42 & +4.07 & +7.06 \\
\texttt{\detokenize{4--10 s}} & 133 & +2.87 & +4.18 & +6.79 \\
\bottomrule
\end{tabularx}
}
\end{center}

Early-to-Official gain decreased with duration because a fixed 0.5-second shift removes a larger fraction of a short step. Outcome and future gains did not depend on a single duration range.

\begin{center}
\captionof{table}{Temporal-boundary intervention effects by action type.}
\label{tab:app_leakage_action}
{
\small
\setlength{\tabcolsep}{3.5pt}
\renewcommand{\arraystretch}{1.08}
\begin{tabularx}{0.78\textwidth}{@{}*{5}{>{\centering\arraybackslash}X}@{}}
\toprule
\textbf{Action type} & \textbf{N} & \textbf{Early $\rightarrow$ Official} & \textbf{Outcome leakage} & \textbf{Future leakage} \\
\midrule
Preparation & 109 & +3.29 & +2.63 & +8.37 \\
Transition & 12 & +3.14 & +2.31 & +9.04 \\
Manipulation & 378 & +3.91 & +4.23 & +6.51 \\
Completion & 9 & +2.76 & +7.64 & +4.18 \\
\bottomrule
\end{tabularx}
}
\end{center}

Completion steps showed the largest outcome gain, while preparation and transition steps showed larger future gains. The transition and completion categories are small and should be treated as exploratory.

The direction of the result was unchanged under step-level micro averaging, equal-scene macro averaging, and equal-source-video macro averaging. It also persisted in the broader 634-sample complete-prediction set: the primary 508-sample gains for Early-to-Official, Outcome, and Future were +3.77, +3.91, and +6.96, compared with +3.28, +2.93, and +5.84 in the broader set. Manual boundary filtering increased effect purity but did not create the trend.

\subsection{Interpretation}

The official boundary occupies a useful middle regime. It performs better than the earlier cutoff, demonstrating that it retains evidence about the developing action, but worse than windows that reveal the outcome or next step. More importantly, each leaked cue selectively improves its semantically corresponding target. The experiment therefore supports interpreting EgoIntent as pre-outcome intent anticipation rather than completed-action recognition.

The study does not establish that every official boundary is perfect: 37 of the 640 diagnostic samples failed the official pre-outcome audit. This 640-sample intervention audit and the independent 756-sample annotation audit use different sampled populations and should not be equated item by item. The fixed offsets used for the artificial conditions are diagnostic interventions rather than proposed benchmark windows.

\section{Temporal Evidence and Static-Shortcut Diagnostics}

\subsection{Diagnostic conditions}

The temporal diagnostic uses \texttt{\detokenize{diagnostic640_v1}} (640 samples, 15 scenes, and all 32 source videos; seed \texttt{\detokenize{20260721}}). Each of four models is evaluated under six conditions:

\begin{center}
\captionof{table}{Input conditions used for the temporal-evidence and static-shortcut diagnostics.}
\label{tab:app_shortcut_conditions}
{
\small
\setlength{\tabcolsep}{3.5pt}
\renewcommand{\arraystretch}{1.08}
\begin{tabularx}{0.88\textwidth}{c>{\centering\arraybackslash}X>{\centering\arraybackslash}p{0.26\textwidth}}
\toprule
\textbf{Condition} & \multicolumn{1}{c}{\textbf{Sample-specific evidence}} & \multicolumn{1}{c}{\textbf{Diagnostic purpose}} \\
\midrule
Ordered Video & Up to 16 uniformly sampled frames in chronological order & Controlled full-visual condition \\
Shuffled Frames & The identical frame set in a deterministic permutation & Contribution of frame order \\
Last Frame Only & The final frame at the observation boundary & Static boundary-state shortcut \\
First Frame Only & The first frame of the observation & Initial static-state cue \\
Scene Only & Normalized scene name & Scene prior \\
Prompt Only & No sample-specific evidence & Prompt/model prior \\
\bottomrule
\end{tabularx}
}
\end{center}

The evaluated systems and execution channels were Doubao-Seed-2.1-Turbo through Volcano Ark, Qwen3.5-Plus through DMX, Qwen3-VL-32B-Instruct through DashScope, and Molmo2-8B through a self-hosted endpoint. Decoding temperature was fixed to zero.

For Ordered and Shuffled, unique source-frame indices, pixels, encoding, and frame count are identical. Only order changes. The permutation is fixed by seed and stored in the condition manifest. Single-frame inputs are transmitted as lossless PNGs. Multi-frame clips below a provider's minimum duration are timestamp-stretched to 2.1 seconds without changing frame content or order. Clips over 15 MB are transported with fixed-QP, all-intra H.264 using identical rules for the paired conditions.

All models use temperature 0 and the same three-field response schema. The automatic Judge receives the four anonymized candidates in a deterministically shuffled order and assigns \localcolor{Local} Accuracy, \proceduralcolor{Procedural} Accuracy, and \nextcolor{Next} Accuracy scores from 0 to 100.

The exact P0-3 shortcut-diagnostic prediction package is reproduced in Section~\ref{sec:app_prompt_p0_3}.

\subsection{Paired diagnostic gains}

For sample $i$ and model $m$,

\begin{equation}
S_{i,m}^{\mathrm{overall}}
=
\frac{S_{i,m}^{L}+S_{i,m}^{P}+S_{i,m}^{N}}{3}.
\label{eq:app_shortcut_overall}
\end{equation}

The three principal within-sample contrasts are

\begin{equation}
\Delta_{\mathrm{temporal}}
=
S_{\mathrm{ordered}}-S_{\mathrm{shuffled}},
\label{eq:app_temporal_gain}
\end{equation}

\begin{equation}
\Delta_{\mathrm{motion}}
=
S_{\mathrm{ordered}}-S_{\mathrm{last}},
\label{eq:app_motion_gain}
\end{equation}

\begin{equation}
\Delta_{\mathrm{visual}}
=
S_{\mathrm{ordered}}-S_{\mathrm{scene}}.
\label{eq:app_visual_gain}
\end{equation}

Equation~(\ref{eq:app_shortcut_overall}) defines the common Overall score used by the shortcut study. Equations~(\ref{eq:app_temporal_gain}), (\ref{eq:app_motion_gain}), and (\ref{eq:app_visual_gain}) isolate temporal order, multi-frame motion, and sample-specific visual evidence, respectively.

A positive temporal gain indicates sensitivity to chronological frame order. A positive motion gain indicates that the multi-frame ordered clip outperforms its boundary frame. A positive visual-evidence gain indicates that sample-specific visual evidence contributes beyond knowing the scene. Confidence intervals use 2,000 source-video-clustered bootstrap replicates; an effect is treated as statistically significant when its 95\% interval excludes zero.

\subsection{Coverage and failure handling}

The common valid set contained 639 samples for Ordered and Shuffled, 640 for Last, First, and Prompt, and 598 for Scene Only. One video was persistently rejected by a provider's content filter. One Molmo2-8B Scene-Only request omitted a required field; because that scene-level response would otherwise have been reused for all 42 samples in the scene, the affected Scene-Only records were excluded. These cases were symmetrically excluded from all four models for the affected condition rather than scored as prediction failures. The final analysis contains 15,184 valid sample-by-model-by-condition records and no Judge failures.

\subsection{Six-condition score profiles}

The main paper reports the principal paired gains. The complete condition profiles below expose the static and prior-based alternatives against which those gains are defined.

\begin{center}
\captionof{table}{Per-model scores under the six shortcut-diagnostic conditions.}
\label{tab:app_shortcut_profiles}
{
\small
\setlength{\tabcolsep}{3.5pt}
\renewcommand{\arraystretch}{1.08}
\begin{tabular*}{0.78\textwidth}{@{\extracolsep{\fill}}cccccc@{}}
\toprule
\textbf{Model} & \textbf{Condition} & \textbf{\localcolor{Local}} & \textbf{\proceduralcolor{Procedural}} & \textbf{\nextcolor{Next}} & \textbf{Overall} \\
\midrule
Doubao & Ordered & 48.24 & 60.46 & 36.62 & 48.44 \\
Doubao & Shuffled & 49.85 & 58.34 & 38.05 & 48.75 \\
Doubao & Last & 51.86 & 57.51 & 40.07 & 49.81 \\
Doubao & First & 35.53 & 46.45 & 30.49 & 37.49 \\
Doubao & Scene & 7.65 & 8.79 & 6.84 & 7.76 \\
Doubao & Prompt & 19.02 & 15.38 & 10.64 & 15.01 \\
Qwen3-VL-32B & Ordered & 40.62 & 48.80 & 32.41 & 40.61 \\
Qwen3-VL-32B & Shuffled & 40.95 & 47.99 & 31.21 & 40.05 \\
Qwen3-VL-32B & Last & 49.33 & 56.04 & 38.55 & 47.97 \\
Qwen3-VL-32B & First & 35.26 & 45.64 & 29.62 & 36.84 \\
Qwen3-VL-32B & Scene & 4.92 & 9.82 & 7.28 & 7.34 \\
Qwen3-VL-32B & Prompt & 3.57 & 3.23 & 9.60 & 5.47 \\
Molmo2-8B & Ordered & 40.47 & 48.33 & 29.43 & 39.41 \\
Molmo2-8B & Shuffled & 38.39 & 45.26 & 27.42 & 37.03 \\
Molmo2-8B & Last & 43.09 & 50.11 & 33.07 & 42.09 \\
Molmo2-8B & First & 33.91 & 43.05 & 27.38 & 34.78 \\
Molmo2-8B & Scene & 3.53 & 5.79 & 4.98 & 4.76 \\
Molmo2-8B & Prompt & 1.34 & 1.33 & 3.80 & 2.16 \\
Qwen3.5-Plus & Ordered & 47.03 & 50.49 & 37.41 & 44.97 \\
Qwen3.5-Plus & Shuffled & 45.19 & 49.00 & 36.45 & 43.55 \\
Qwen3.5-Plus & Last & 51.80 & 51.97 & 40.36 & 48.04 \\
Qwen3.5-Plus & First & 39.20 & 46.44 & 33.37 & 39.67 \\
Qwen3.5-Plus & Scene & 10.02 & 10.00 & 6.61 & 8.88 \\
Qwen3.5-Plus & Prompt & 22.12 & 18.65 & 5.77 & 15.51 \\
\bottomrule
\end{tabular*}
}
\end{center}

Prompt-only behavior reveals a substantial difference in task priors: Doubao and Qwen3.5-Plus retain Overall scores near 15 without sample-specific evidence, whereas Qwen3-VL-32B and Molmo2-8B fall to 5.47 and 2.16. Prompt-only output must therefore not be interpreted as visual understanding.

\subsection{Duration-stratified shortcut behavior}

The following differences are descriptive because separate duration-specific clustered intervals and multiplicity corrections were not computed.

\begin{center}
\captionof{table}{Temporal and motion gains by model and step duration.}
\label{tab:app_shortcut_duration}
{
\small
\setlength{\tabcolsep}{3.5pt}
\renewcommand{\arraystretch}{1.08}
\begin{tabular}{cccc}
\toprule
\textbf{Model} & \textbf{Duration} & \textbf{Temporal gain} & \textbf{Motion gain} \\
\midrule
Doubao & \texttt{\detokenize{<1 s}} & +2.76 & +6.07 \\
Doubao & \texttt{\detokenize{1--2 s}} & -0.61 & -2.27 \\
Doubao & \texttt{\detokenize{2--4 s}} & -3.23 & -5.79 \\
Doubao & \texttt{\detokenize{4--10 s}} & -0.12 & -3.52 \\
Qwen3-VL-32B & \texttt{\detokenize{<1 s}} & +2.32 & -1.49 \\
Qwen3-VL-32B & \texttt{\detokenize{1--2 s}} & -0.78 & -9.69 \\
Qwen3-VL-32B & \texttt{\detokenize{2--4 s}} & -0.85 & -10.65 \\
Qwen3-VL-32B & \texttt{\detokenize{4--10 s}} & +1.64 & -7.52 \\
Molmo2-8B & \texttt{\detokenize{<1 s}} & +2.06 & -1.02 \\
Molmo2-8B & \texttt{\detokenize{1--2 s}} & +3.26 & -3.42 \\
Molmo2-8B & \texttt{\detokenize{2--4 s}} & +1.22 & -3.38 \\
Molmo2-8B & \texttt{\detokenize{4--10 s}} & +3.00 & -2.86 \\
Qwen3.5-Plus & \texttt{\detokenize{<1 s}} & +0.79 & +3.20 \\
Qwen3.5-Plus & \texttt{\detokenize{1--2 s}} & +1.06 & -6.36 \\
Qwen3.5-Plus & \texttt{\detokenize{2--4 s}} & +1.30 & -5.59 \\
Qwen3.5-Plus & \texttt{\detokenize{4--10 s}} & +2.63 & -3.39 \\
\bottomrule
\end{tabular}
}
\end{center}

Molmo2 is the only model with positive temporal gain in every duration bin, consistent with its significant aggregate gain, although its absolute Ordered score is the lowest among the four models. Qwen3-VL-32B exhibits the largest boundary-frame shortcut, especially for 1--4 second observations. Because negative motion gain can arise from either an informative boundary frame or failed multi-frame integration, it should not be interpreted as evidence that motion itself is harmful.

\subsection{Additional conclusions and limits}

The four models all obtain large positive visual-evidence gains, so their predictions are not explained by scene labels alone. In contrast, only Molmo2 shows a statistically reliable, modest temporal gain, concentrated in \proceduralcolor{Procedural Intent}. Higher absolute performance and larger parameter count therefore do not imply greater frame-order sensitivity.

These conclusions apply to a 640-sample balanced diagnostic set, four models, and at most 16 sparsely sampled frames. Provider-side video decoding and internal sampling are not directly observable. Scene extrema and duration-bin results are descriptive and were not corrected for multiple comparisons. Future versions should include temporal counterfactual pairs and a temporal-hard subset in which the final static state is deliberately insufficient.

\section{Historical Context and Step-Duration Analysis}

\subsection{Controlled history windows}

The context study uses the same \texttt{\detokenize{diagnostic640_v1}} subset and four models. Each input ends at the same official \texttt{\detokenize{obs_end_time}}; no future frame is included. Four conditions are compared:

\begin{center}
\captionof{table}{Input conditions used for the historical-context experiment.}
\label{tab:app_history_conditions}
{
\small
\setlength{\tabcolsep}{3.5pt}
\renewcommand{\arraystretch}{1.08}
\begin{tabularx}{0.82\textwidth}{>{\centering\arraybackslash}p{0.16\textwidth}>{\centering\arraybackslash}X}
\toprule
\textbf{Condition} & \textbf{Temporal support} \\
\midrule
Step-only & Current micro-step: \texttt{\detokenize{[step_start, obs_end]}} \\
History-5s & \texttt{\detokenize{[max(0, step_start-5s), obs_end]}} \\
History-15s & \texttt{\detokenize{[max(0, step_start-15s), obs_end]}} \\
Previous-step & Temporal union of the previous complete annotated step and the current step \\
\bottomrule
\end{tabularx}
}
\end{center}

For Previous-step, unannotated gaps between the two steps are excluded. If no previous step exists, the condition reduces to Step-only and is marked in the manifest.

Every sample and condition contains exactly 16 uniformly sampled frames. All 2,560 condition media passed frame-count validation. For steps shorter than one source-video frame interval, the nearest decoded boundary frame is repeated. When required by an API, only the transmission timeline is stretched to 2.1 seconds; pixels, frame order, and frame count remain unchanged.

The fixed frame budget controls visual cost but changes sampling density: a longer history window devotes fewer frames to the current micro-step. The experiment therefore measures context selection and integration under a fixed evidence budget, not the unconstrained value of history.

The exact P0-4 context-length prediction package is reproduced in Section~\ref{sec:app_prompt_p0_4}.

\subsection{Context gains and statistical protocol}

For each sample and model,

\begin{equation}
\Delta_{5s}
=
S_{\mathrm{history\mbox{-}5s}}
-S_{\mathrm{step\mbox{-}only}},
\label{eq:app_history5_gain}
\end{equation}

\begin{equation}
\Delta_{15s}
=
S_{\mathrm{history\mbox{-}15s}}
-S_{\mathrm{step\mbox{-}only}},
\label{eq:app_history15_gain}
\end{equation}

\begin{equation}
\Delta_{\mathrm{prev}}
=
S_{\mathrm{previous\mbox{-}step}}
-S_{\mathrm{step\mbox{-}only}}.
\label{eq:app_previous_step_gain}
\end{equation}

Equations~(\ref{eq:app_history5_gain}), (\ref{eq:app_history15_gain}), and (\ref{eq:app_previous_step_gain}) quantify the paired effects of 5-second history, 15-second history, and the previous annotated step relative to Step-only input.

All comparisons are sample-paired. Confidence intervals use 2,000 source-video-clustered bootstrap replicates. The common prediction sets contain 639 Step-only, 638 History-5s, 640 History-15s, and 636 Previous-step samples. Seven media were persistently rejected by provider-side inspection; they were excluded symmetrically rather than scored as model errors. The analysis includes 10,212 complete sample-by-model-by-condition records.

\subsection{Dimension-level condition scores}

The main paper visualizes Overall context performance. The table below adds the \localcolor{Local} Accuracy, \proceduralcolor{Procedural} Accuracy, and \nextcolor{Next} Accuracy components.

\begin{center}
\captionof{table}{Per-model scores under the four historical-context conditions.}
\label{tab:app_history_profiles}
{
\small
\setlength{\tabcolsep}{3.5pt}
\renewcommand{\arraystretch}{1.08}
\begin{tabular*}{0.78\textwidth}{@{\extracolsep{\fill}}>{\centering\arraybackslash}p{0.16\textwidth}ccccc@{}}
\toprule
\textbf{Model} & \textbf{Condition} & \textbf{\localcolor{Local}} & \textbf{\proceduralcolor{Procedural}} & \textbf{\nextcolor{Next}} & \textbf{Overall} \\
\midrule
Doubao & Step-only & 56.99 & 63.63 & 47.27 & 55.96 \\
Doubao & History-5s & 49.08 & 56.79 & 41.71 & 49.19 \\
Doubao & History-15s & 49.78 & 56.47 & 42.33 & 49.53 \\
Doubao & Previous-step & 51.38 & 59.36 & 46.66 & 52.46 \\
Qwen3-VL-32B & Step-only & 45.62 & 51.12 & 33.96 & 43.57 \\
Qwen3-VL-32B & History-5s & 43.43 & 50.56 & 35.86 & 43.28 \\
Qwen3-VL-32B & History-15s & 41.70 & 48.49 & 35.06 & 41.75 \\
Qwen3-VL-32B & Previous-step & 43.05 & 49.54 & 34.85 & 42.48 \\
Molmo2-8B & Step-only & 44.39 & 51.45 & 34.23 & 43.36 \\
Molmo2-8B & History-5s & 42.03 & 49.38 & 33.04 & 41.49 \\
Molmo2-8B & History-15s & 40.61 & 46.96 & 31.53 & 39.70 \\
Molmo2-8B & Previous-step & 42.92 & 48.66 & 31.75 & 41.11 \\
Qwen3.5-Plus & Step-only & 52.69 & 55.16 & 41.29 & 49.71 \\
Qwen3.5-Plus & History-5s & 49.66 & 54.34 & 39.82 & 47.94 \\
Qwen3.5-Plus & History-15s & 44.79 & 50.05 & 36.36 & 43.73 \\
Qwen3.5-Plus & Previous-step & 48.60 & 52.39 & 40.69 & 47.23 \\
\bottomrule
\end{tabular*}
}
\end{center}

\localcolor{Local} Accuracy decreases most consistently when history is added, consistent with current-step evidence becoming less visually dense. \proceduralcolor{Procedural} Accuracy and \nextcolor{Next} Accuracy also fail to improve reliably. The isolated Qwen3-VL-32B History-5s \nextcolor{Next} Accuracy estimate is positive (+1.97 relative to Step-only), but its 95\% interval includes zero.

\subsection{Duration dependence}

The following Overall scores are descriptive condition-specific micro averages:

\begin{center}
\captionof{table}{Historical-context results by model and step duration.}
\label{tab:app_history_duration_model}
{
\small
\setlength{\tabcolsep}{3.5pt}
\renewcommand{\arraystretch}{1.08}
\begin{tabular*}{0.78\textwidth}{@{\extracolsep{\fill}}>{\centering\arraybackslash}p{0.16\textwidth}ccccc@{}}
\toprule
\textbf{Model} & \textbf{Duration} & \textbf{Step-only} & \textbf{History-5s} & \textbf{History-15s} & \textbf{Previous-step} \\
\midrule
Doubao & \texttt{\detokenize{<1 s}} & 58.07 & 47.85 & 45.06 & 52.20 \\
Doubao & \texttt{\detokenize{1--2 s}} & 56.80 & 45.90 & 46.02 & 49.16 \\
Doubao & \texttt{\detokenize{2--4 s}} & 51.36 & 50.76 & 51.71 & 53.71 \\
Doubao & \texttt{\detokenize{4--10 s}} & 57.68 & 52.53 & 55.72 & 55.05 \\
Qwen3-VL-32B & \texttt{\detokenize{<1 s}} & 47.34 & 42.60 & 38.16 & 41.08 \\
Qwen3-VL-32B & \texttt{\detokenize{1--2 s}} & 41.11 & 42.11 & 40.07 & 40.24 \\
Qwen3-VL-32B & \texttt{\detokenize{2--4 s}} & 41.42 & 45.07 & 44.87 & 43.85 \\
Qwen3-VL-32B & \texttt{\detokenize{4--10 s}} & 44.53 & 43.38 & 44.05 & 44.97 \\
Molmo2-8B & \texttt{\detokenize{<1 s}} & 45.41 & 40.98 & 35.93 & 42.17 \\
Molmo2-8B & \texttt{\detokenize{1--2 s}} & 44.70 & 39.36 & 39.82 & 38.25 \\
Molmo2-8B & \texttt{\detokenize{2--4 s}} & 41.80 & 43.39 & 41.00 & 42.69 \\
Molmo2-8B & \texttt{\detokenize{4--10 s}} & 41.40 & 42.31 & 42.15 & 41.48 \\
Qwen3.5-Plus & \texttt{\detokenize{<1 s}} & 51.64 & 47.16 & 41.35 & 48.48 \\
Qwen3.5-Plus & \texttt{\detokenize{1--2 s}} & 49.58 & 48.72 & 40.35 & 45.06 \\
Qwen3.5-Plus & \texttt{\detokenize{2--4 s}} & 46.51 & 46.43 & 43.65 & 44.73 \\
Qwen3.5-Plus & \texttt{\detokenize{4--10 s}} & 51.20 & 49.51 & 49.98 & 50.93 \\
\bottomrule
\end{tabular*}
}
\end{center}

Contrary to the hypothesis that short steps lack sufficient context, the largest average losses occurred below 2 seconds:

\begin{center}
\captionof{table}{Average historical-context effects by step duration.}
\label{tab:app_history_duration_avg}
{
\small
\setlength{\tabcolsep}{3.5pt}
\renewcommand{\arraystretch}{1.08}
\begin{tabular}{cccc}
\toprule
\textbf{Duration} & \textbf{History-5s minus Step} & \textbf{History-15s minus Step} & \textbf{Previous-step minus Step} \\
\midrule
\texttt{\detokenize{<1 s}} & -5.97 & -10.49 & -4.63 \\
\texttt{\detokenize{1--2 s}} & -4.03 & -6.49 & -4.87 \\
\texttt{\detokenize{2--4 s}} & +1.14 & +0.03 & +0.97 \\
\texttt{\detokenize{4--10 s}} & -1.77 & -0.73 & -0.59 \\
\bottomrule
\end{tabular}
}
\end{center}

The 2--4 second bin is the only range with approximately neutral or slightly positive average effects. Very short Step-only inputs are dominated by repeated boundary-state evidence; when history is added under the same 16-frame budget, that evidence is diluted.

\subsection{Macro aggregation and interpretation}

Equal-scene and equal-source-video macro averages produced the same ordering as step-level micro averages: Step-only was highest for all four models, while History-15s was usually lowest. The result is therefore not explained by one large scene or one source video with many steps.

The experiment cannot distinguish two mechanisms: many samples may genuinely be answerable from the current step, or useful history may exist but current models may fail to select and integrate it. The appropriate conclusion is that the evaluated systems did not benefit from history under a fixed 16-frame budget. It is not that history is intrinsically harmful. A stronger follow-up should pair the current-step frames with an additional fixed history stream and should construct a context-required subset whose current frames are deliberately ambiguous.

\section{Automatic Judge Validation}

\subsection{Balanced validation-set construction}

Judge validation uses 150 unique benchmark steps, each expanded to one closed-model and one open-model prediction-reference triplet, for 300 triplets in total. The candidate pool contained 17,940 valid triplets from six preselected models. A deterministic mixed-integer linear program with seed \texttt{\detokenize{20260713}} jointly selected the steps and assigned model pairs.

The selected solution satisfied the following hard constraints: exactly ten unique steps from each of 15 scenes, coverage of all 32 source videos, and duration counts of 36 below 1 second, 43 from 1--2 seconds, 37 from 2--4 seconds, 33 from 4--10 seconds, and one above 10 seconds. Each of the six models contributed exactly 50 triplets. The nine closed/open pairings were balanced between 16 and 17 unique steps.

Three domain-expert raters independently scored the same 300 anonymized triplets. The interface hid model/provider identity, scene and event labels, sample identifiers, and all automatic scores. Because two triplets can share the same video step and multiple steps can share a source video, uncertainty intervals are clustered by source video.

The exact P0-2 blind LLM-rater package is reproduced in Section~\ref{sec:app_prompt_p0_2}.

\subsection{Agreement and error statistics}

Let $H_{i,d}$ be the mean human score and $J_{i,d}$ an automatic Judge score for item $i$ and dimension $d$. In addition to Pearson correlation, we report Spearman rank correlation, Kendall's $\tau_b$, mean absolute error, and signed bias:

\begin{equation}
\mathrm{MAE}_d
=
\frac{1}{n}\sum_{i=1}^{n}
\left|J_{i,d}-H_{i,d}\right|,
\label{eq:app_judge_mae}
\end{equation}

\begin{equation}
\mathrm{Bias}_d
=
\frac{1}{n}\sum_{i=1}^{n}
\left(J_{i,d}-H_{i,d}\right).
\label{eq:app_judge_bias}
\end{equation}

The combined score is first computed within item,

\begin{equation}
\bar H_i=\frac{H_{i,L}+H_{i,P}+H_{i,N}}{3},
\qquad
\bar J_i=\frac{J_{i,L}+J_{i,P}+J_{i,N}}{3},
\label{eq:app_judge_combined_score}
\end{equation}

Equation~(\ref{eq:app_judge_mae}) measures absolute Judge error, Eq.~(\ref{eq:app_judge_bias}) measures signed error, and Eq.~(\ref{eq:app_judge_combined_score}) constructs the per-item combined human and Judge scores used in the Overall comparison.

The resulting per-item averages are then used to compute combined correlation and error. The combined MAE is therefore not the arithmetic mean of the three dimension-level MAEs.

Human consistency is quantified with two-way random-effects absolute-agreement ICC for the mean of the three raters, ICC(2,$k$), and Krippendorff's $\alpha$. Both statistics are computed from the original independent ratings rather than the adjudicated scores.

\subsection{Supplemental Judge--human agreement}

The main paper reports the compact reliability and Judge-validation results. The following extended table adds Pearson correlation and Kendall's $\tau_b$ and gives the complete secondary-Judge comparison.

\begin{center}
\captionof{table}{Automatic-Judge agreement with human consensus.}
\label{tab:app_judge_agreement}
{
\small
\setlength{\tabcolsep}{3.5pt}
\renewcommand{\arraystretch}{1.08}
\begin{tabular*}{0.88\textwidth}{@{\extracolsep{\fill}}>{\centering\arraybackslash}p{0.18\textwidth}cccccc@{}}
\toprule
\textbf{Judge} & \textbf{Dimension} & \textbf{Pearson $r$} & \textbf{Spearman $\rho$} & \textbf{Kendall $\tau_b$} & \textbf{MAE} & \textbf{Bias} \\
\midrule
DeepSeek-V4-Flash & \localcolor{Local} & 0.78 & 0.76 & 0.58 & 11.1 & -1.0 \\
DeepSeek-V4-Flash & \proceduralcolor{Procedural} & 0.74 & 0.72 & 0.54 & 12.5 & -1.1 \\
DeepSeek-V4-Flash & \nextcolor{Next} & 0.65 & 0.67 & 0.49 & 15.6 & +1.0 \\
DeepSeek-V4-Flash & Overall & 0.80 & 0.78 & 0.60 & 9.8 & -0.4 \\
GLM-5.1 & \localcolor{Local} & 0.73 & 0.71 & 0.52 & 12.0 & -0.8 \\
GLM-5.1 & \proceduralcolor{Procedural} & 0.69 & 0.68 & 0.49 & 13.0 & -1.5 \\
GLM-5.1 & \nextcolor{Next} & 0.60 & 0.62 & 0.44 & 15.1 & +0.3 \\
GLM-5.1 & Overall & 0.75 & 0.73 & 0.54 & 11.4 & -0.7 \\
\bottomrule
\end{tabular*}
}
\end{center}

Pearson and Spearman correlations are close, indicating that agreement is not produced only by a small number of extreme scores. Bias remains small relative to the 0--100 scale. \nextcolor{Next Plan} is the least reliable dimension for both Judges, consistent with its higher human ambiguity.

\subsection{Model-ranking preservation}

The primary Judge preserved the top three and bottom model positions and swapped only Gemini 3.5 Flash and Molmo2-8B, whose human scores were close. The secondary Judge preserved the full six-model order.

\begin{center}
\captionof{table}{Model-ranking preservation under human and automatic-Judge scores.}
\label{tab:app_judge_rank}
{
\small
\setlength{\tabcolsep}{3.5pt}
\renewcommand{\arraystretch}{1.08}
\begin{tabular*}{0.88\textwidth}{@{\extracolsep{\fill}}>{\centering\arraybackslash}p{0.22\textwidth}cccccc@{}}
\toprule
\textbf{Model} & \makecell{\textbf{Human}\\\textbf{Overall}} & \makecell{\textbf{DeepSeek}\\\textbf{Overall}} & \makecell{\textbf{GLM}\\\textbf{Overall}} & \makecell{\textbf{Human}\\\textbf{rank}} & \makecell{\textbf{DeepSeek}\\\textbf{rank}} & \makecell{\textbf{GLM}\\\textbf{rank}} \\
\midrule
Doubao-Seed-2.1-Turbo & 60.8 & 61.6 & 59.9 & 1 & 1 & 1 \\
Qwen3.5-Plus & 57.1 & 56.4 & 58.0 & 2 & 2 & 2 \\
Qwen3-VL-32B-Instruct & 47.2 & 45.8 & 44.1 & 3 & 3 & 3 \\
Gemini 3.5 Flash & 44.8 & 43.9 & 43.5 & 4 & 5 & 4 \\
Molmo2-8B & 42.6 & 44.0 & 42.0 & 5 & 4 & 5 \\
InternVL3-8B & 34.1 & 32.8 & 35.0 & 6 & 6 & 6 \\
\bottomrule
\end{tabular*}
}
\end{center}

Only six model-level points are available, so ranking preservation is supportive rather than the primary validity evidence. The principal evidence is the 300-triplet sample-level agreement.

\subsection{Large disagreements and scene effects}

We define a large dimension-level disagreement as

\begin{equation}
D_{i,d}
=
\mathbf{1}
\left(
\left|J_{i,d}-H_{i,d}\right|\ge25
\right).
\label{eq:app_large_disagreement}
\end{equation}

Equation~(\ref{eq:app_large_disagreement}) flags dimension-level Judge errors of at least 25 points for the large-disagreement analysis.

\begin{center}
\captionof{table}{Rates of large automatic-Judge disagreements with human consensus.}
\label{tab:app_judge_large_disagreement}
{
\small
\setlength{\tabcolsep}{3.5pt}
\renewcommand{\arraystretch}{1.08}
\begin{tabular}{cccc}
\toprule
\textbf{Judge} & \textbf{\localcolor{Local}} & \textbf{\proceduralcolor{Procedural}} & \textbf{\nextcolor{Next}} \\
\midrule
DeepSeek-V4-Flash & 9.3\% & 12.0\% & 18.3\% \\
GLM-5.1 & 11.7\% & 14.3\% & 21.3\% \\
\bottomrule
\end{tabular}
}
\end{center}

Large \nextcolor{Next Plan} disagreements frequently reflect confusion between the observed continuation and a merely plausible alternative, or a disagreement about whether the predicted action is immediate or too far in the future. \localcolor{Local} disagreements are more often attributable to the wrong action, object, or an overly vague description; \proceduralcolor{Procedural} disagreements typically involve the wrong higher-level goal or \localcolor{Local}--\proceduralcolor{Procedural} conflation.

Judge error varies by scene. The highest three-dimension MAE values for the primary Judge were observed in yard (19.2, bias -5.8), garage (17.2, bias -3.5), and hallway (16.6, bias -2.8). The result suggests that scene-conditional Judge checks are useful even when global correlation is high. In particular, the primary Judge systematically underscored yard predictions relative to human consensus.

\subsection{Validity boundary}

The validation supports automatic scoring as a scalable approximation to human semantic assessment, not as a perfect substitute. \nextcolor{Next Plan} remains the weakest dimension, close models can exchange adjacent ranks, and certain scenes show larger systematic error. For this reason, the diagnostic experiments use sample-paired contrasts, anonymized candidates, and source-video-clustered intervals; their conclusions rely more strongly on within-sample directional effects than on small absolute score differences.

\section{Shared Reproducibility and Reporting Conventions}

\subsection{Micro, macro, and paired estimands}

Unless otherwise stated, reported benchmark and diagnostic scores are step-level micro averages:

\begin{equation}
\bar S_{\mathrm{micro}}
=
\frac{1}{N}\sum_{i=1}^{N}S_i.
\label{eq:app_micro_average}
\end{equation}

For $C$ scenes and $V$ source videos, equal-group macro averages are

\begin{equation}
\bar S_{\mathrm{scene}}
=
\frac{1}{C}\sum_{c=1}^{C}
\left(
\frac{1}{N_c}\sum_{i\in c}S_i
\right),
\label{eq:app_scene_macro_average}
\end{equation}

\begin{equation}
\bar S_{\mathrm{video}}
=
\frac{1}{V}\sum_{v=1}^{V}
\left(
\frac{1}{N_v}\sum_{i\in v}S_i
\right).
\label{eq:app_video_macro_average}
\end{equation}

Intervention effects are always estimated as paired differences before averaging:

\begin{equation}
\widehat{\Delta}
=
\frac{1}{N_{\cap}}
\sum_{i\in\mathcal I_{\cap}}
\left(S_i^{(A)}-S_i^{(B)}\right),
\label{eq:app_paired_effect}
\end{equation}

Equation~(\ref{eq:app_micro_average}) defines the default step-level estimand, Eqs.~(\ref{eq:app_scene_macro_average}) and (\ref{eq:app_video_macro_average}) define equal-group macro estimands, and Eq.~(\ref{eq:app_paired_effect}) defines the paired intervention effect on the common valid sample set.

Here, $\mathcal I_{\cap}$ is the common valid sample set for conditions $A$ and $B$. The difference of two independently rounded condition means is not used as a substitute for the paired estimate.

\subsection{Clustered bootstrap}

All principal intervention intervals treat the source video as the resampling unit. If the 32 source-video IDs are denoted by $\mathcal V$, each bootstrap replicate samples 32 elements from $\mathcal V$ with replacement and includes all descendant steps with the sampled multiplicity. The statistic is recomputed on the paired data in each replicate. Human-baseline intervals use 10,000 replicates; the diagnostic intervention studies use 2,000.

\subsection{Missing predictions and provider failures}

For the full benchmark, a model-side missing or invalid prediction is scored as zero because coverage is part of benchmark performance. In controlled multi-condition experiments, a stable provider content-filter rejection or media-format failure is treated as an execution failure rather than an incorrect model answer. The affected sample is removed symmetrically from all compared models for that condition, and the common-set size is reported. These two policies address different estimands and should not be conflated.

\section{Complete Prompt Packages}

This section consolidates the seven exact prompt packages used for benchmark prediction, automatic scoring, controlled diagnostics, and Judge validation. Each subsection first states the package's purpose and input--output contract, then reproduces the frozen System Prompt and User Prompt template. Placeholder tokens denote values supplied at runtime; the prompt wording itself is unchanged.

\subsection{Official Benchmark Prediction Prompt}
\label{sec:app_prompt_official_prediction}

\textbf{Purpose.} This package generated the three benchmark predictions for every evaluated model: \localcolor{Local Intent}, \proceduralcolor{Procedural Intent}, and \nextcolor{Next Plan}. \textbf{Input and output.} The user message was paired with the complete pre-outcome observation-window video (or the documented four-frame fallback for provider-rejected short clips), without task, scene, narration, reference-label, or future-frame metadata. The model returned exactly one JSON object containing the three predicted labels. The same package was shared verbatim across all 15 evaluated systems.

\checksumline{Combined prompt-package SHA-256}{9962d19a5ba1a952f6dc21e086904c3497e46820033e8d77228cc1f173987e28}

\promptheading{System prompt}

\begin{lstlisting}[style=promptcode]
You are evaluating egocentric procedural video understanding.

You will receive one video containing the complete observation window of a single benchmark step. Watch the entire video in chronological order and infer exactly three labels.

LABEL DEFINITIONS

1. local_intent
The actor's immediate goal within the observed video.
Describe what the actor is trying to accomplish through the current action, rather than only describing body motion.
An action-form phrase is acceptable when it clearly expresses the immediate goal.
Keep the description specific to the current step.

2. procedural_intent
The near-term procedural subgoal that the local intent supports.
It should be one level more abstract than local_intent, but it must remain visually and procedurally supported.
Do not infer a broad event-level goal merely from the scene or objects.

3. next_step
The single most likely action that will occur immediately after the observation window ends.
Predict only one action.
Do not repeat an action that has already been completed in the video.
If an action is clearly unfinished at the end of the video, its immediate continuation may be predicted.
Keep the prediction temporally close and at a similar level of granularity to the observed action.

EVIDENCE RULES

- Use motion and state changes across the entire video, not only the final frame.
- Treat the final frame as the temporal boundary between the observation and the predicted next step.
- Base all predictions only on visible evidence and ordinary procedural continuity.
- Do not invent specific objects, tools, directions, or goals that are not visually supported.
- When multiple interpretations are possible, choose the single most visually supported and temporally immediate interpretation.
- Describe the camera wearer or primary actor, not the camera movement.

OUTPUT REQUIREMENTS

- Write all labels in English.
- Use concise verb phrases.
- Return valid JSON only.
- Do not include explanations, reasoning, confidence scores, alternatives, or additional fields.

REQUIRED OUTPUT FORMAT

{
  "local_intent": "string",
  "procedural_intent": "string",
  "next_step": "string"
}
\end{lstlisting}

\promptheading{User prompt template}

\begin{lstlisting}[style=promptcode]
Watch the attached step video in full and predict the three required labels.

Return only the JSON object specified in the system prompt.
\end{lstlisting}

\subsection{Reference-Based Semantic Judge Prompt}
\label{sec:app_prompt_semantic_judge}

\textbf{Purpose.} This package produced the official \localcolor{Local} Accuracy, \proceduralcolor{Procedural} Accuracy, and \nextcolor{Next} Accuracy scores used in the benchmark ranking. \textbf{Input and output.} The Judge received frozen human references together with deterministically anonymized model candidates; model identities and ranking information were withheld. It independently returned 0--100 semantic-correctness scores for all three targets, a \nextcolor{Next Plan} match type, and a brief justification for every anonymous candidate.

\checksumline{Combined prompt-package SHA-256}{cf2d1e29154af2fedd391d3a2084010081addfedd1f7b7803327183697fbfa28}

\promptheading{System prompt}

\begin{lstlisting}[style=promptcode]
You are an impartial evaluator for the EgoIntent benchmark.

Your task is to score predicted intent labels against human-annotated reference labels. Evaluate semantic correctness, not lexical overlap. Predictions may use different wording, grammatical forms, or levels of detail without being wrong.

For each anonymous candidate, independently score:

1. local_intent
   The actor's immediate, short-horizon intention during the observed segment.

2. procedural_intent
   The broader procedural subgoal that the immediate intention serves.

3. next_plan
   The action predicted to occur immediately after the observation boundary.

Do not compare candidates against one another and do not force a ranking. Multiple candidates may receive the same score. Never infer model identity from writing style.

GENERAL SCORING FOR LOCAL_INTENT AND PROCEDURAL_INTENT

Score each field from 0 to 100:

- 90-100: Semantically equivalent to the reference. The core intention, relevant object, and goal are correct.
- 70-89: Mostly correct, with a minor omission, harmless extra detail, or reasonable difference in granularity.
- 40-69: Partially correct. It captures part of the intended meaning but misses or changes an important component.
- 1-39: Only weakly related to the reference or describes the surrounding activity without identifying the intended goal.
- 0: Incorrect, contradictory, unrelated, or not a meaningful answer.

Do not penalize:
- synonyms;
- gerund versus infinitive forms;
- grammatical style;
- harmless differences in specificity.

Do penalize:
- incorrect objects;
- incorrect action direction;
- confusion between an observed action and its intended goal;
- overly broad activity descriptions that fail to identify the intended intention;
- invented details that materially change the meaning.

MISSING PREDICTIONS

If a candidate uses the exact value "[MISSING_PREDICTION]" for its fields, the model did not provide a prediction. Assign 0 to local_intent_score, procedural_intent_score, and next_plan_score; assign "none" to next_plan_match_type. Do not infer or reconstruct a missing prediction from the references or from other candidates.

SPECIAL SCORING FOR NEXT_PLAN

The reference contains:
- observed_next_step: the action that actually occurred next;
- plausible_next_steps: reasonable alternatives that could have occurred but were not the observed next action.

Score next_plan from 0 to 100 according to the following priority:

A. Match to observed_next_step
- 90-100: Clear semantic match to the action that actually occurred next.
- 70-89: Same core observed action, but underspecified, overly broad, or containing a minor non-contradictory error.

B. Match to plausible_next_steps but not observed_next_step
- 45-69: Clear semantic match to one of the listed plausible alternatives.
- 20-44: Partial or underspecified match to a listed plausible alternative.

C. No reference match
- 1-19: Related to the ongoing procedure, but does not semantically match either the observed next step or any listed plausible alternative.
- 0: Unrelated, contradictory, impossible given the labels, or not a meaningful action.

A prediction matching the observed next step must always score higher than a prediction matching only a plausible alternative.

If a prediction could match both the observed step and a plausible alternative, prioritize the observed_next_step interpretation.

A wrong action direction is a major error. For example, "pick up the paper" and "put down the paper" are not equivalent. However, if the directionally different action explicitly matches a listed plausible alternative, score it under the plausible range instead of treating it as unrelated.

The plausible list is a reference set, not permission to invent additional alternatives. Do not classify an unlisted action as a plausible match merely because it seems reasonable from general world knowledge.

Assign one next_plan_match_type:
- "observed": matches the observed_next_step;
- "plausible": matches a listed plausible_next_step but not the observed step;
- "related": related to the procedure but matches neither reference;
- "none": unrelated, contradictory, or invalid.

OUTPUT REQUIREMENTS

Return valid JSON only. Do not use Markdown fences.

Output exactly one result for every input candidate.

Use this schema:

{
  "step_id": <integer>,
  "evaluations": [
    {
      "candidate_id": "<anonymous candidate ID>",
      "local_intent_score": <integer 0-100>,
      "procedural_intent_score": <integer 0-100>,
      "next_plan_score": <integer 0-100>,
      "next_plan_match_type": "observed|plausible|related|none",
      "brief_reason": "<concise explanation, maximum 30 words>"
    }
  ]
}
\end{lstlisting}

\promptheading{User prompt template}

\begin{lstlisting}[style=promptcode]
Evaluate the anonymous predictions for this EgoIntent step.

Score every candidate independently according to the system rubric. Candidate order is randomized and has no meaning.

INPUT:

{
  "step_id": {{STEP_ID}},
  "reference": {
    "local_intent": {{REFERENCE_LOCAL_INTENT_JSON}},
    "procedural_intent": {{REFERENCE_PROCEDURAL_INTENT_JSON}},
    "observed_next_step": {{REFERENCE_OBSERVED_NEXT_STEP_JSON}},
    "plausible_next_steps": {{REFERENCE_PLAUSIBLE_NEXT_STEPS_JSON}}
  },
  "candidates": {{ANONYMIZED_CANDIDATES_JSON}}
}
\end{lstlisting}

\subsection{Reference-Free Video Diagnostic Judge Prompt}
\label{sec:app_prompt_diagnostic_judge}

\textbf{Purpose.} This package measured Grounding Faithfulness (GF), Hierarchical Intent Consistency (HIC), Temporal Progression Consistency (TPC), and \nextcolor{Next Plan} Feasibility (NPF) independently of reference matching. \textbf{Input and output.} The video-aware Judge received the pre-outcome observation and anonymized predictions, but no reference labels. It returned one integer score from 0 to 4 for each diagnostic and candidate; the reported 0--100 values were obtained by multiplying these raw scores by 25.

\checksumline{Combined prompt-package SHA-256}{0dc8971b222094ba319be3ea0fc0bfc4e058a005dc8dee8280675792c3990fd4}

\promptheading{System prompt}

\begin{lstlisting}[style=promptcode]
You are the video-aware evaluator for the EgoIntent benchmark.

You receive one temporally ordered, pre-outcome egocentric video observation and a set of anonymized model predictions. Each prediction contains:

- local_intent: the immediate objective of the current observed step;
- procedural_intent: the higher-level functional subgoal that explains the local intent;
- next_plan: the single action expected immediately after the observation boundary.

Evaluate every candidate independently. Do not rank candidates, compare their wording with one another, or infer that a fluent answer is visually correct. Use only the attached observation and the candidate itself. The video ends at the observation boundary; content after that boundary is not visible.

Return four integer scores from 0 to 4 for every candidate:

1. GF (Visual Grounding Faithfulness)
4: all key claims are clearly supported by visible evidence.
3: generally supported, with minor reasonable inference.
2: partially supported, with noticeable speculation.
1: mainly scene priors or weak clues.
0: hallucination, contradiction, or treating an already observed/future event incorrectly.

2. HIC (Hierarchical Intent Consistency)
4: clear hierarchy; procedural_intent explains why local_intent is needed.
3: correct relationship but slightly broad or overlapping.
2: related, but the hierarchy is unclear.
1: mostly paraphrase, synonym, or an excessively broad event goal.
0: contradictory, unrelated, or missing.

3. TPC (Temporal Progression Consistency)
Judge the chain current local intent -> procedural role -> next plan.
4: clear step boundary and natural temporal progression.
3: correct ordering with slight boundary ambiguity.
2: reasonable, but current-versus-next timing is unclear.
1: next_plan overlaps the current action or jumps too far ahead.
0: temporal order is wrong, or next_plan is already completed in the observation.

4. NPF (\nextcolor{Next-Plan} Feasibility)
4: immediately executable and highly consistent with the visible current state.
3: reasonable and feasible but slightly broad or indirect.
2: possible, but important conditions are missing.
1: physically possible but procedurally unnatural.
0: impossible, contradictory, wrongly ordered, or requires unavailable objects/tools.

Important rules:

- Evaluate semantic content, not writing style.
- A plausible next_plan is not automatically visually grounded; score each metric separately.
- Do not penalize concise wording when its meaning is clear.
- If any prediction field is [MISSING_PREDICTION], assign 0 to all four metrics for that candidate.
- Use the full temporal sequence, especially changes near the final observation boundary.
- Output JSON only, with exactly one evaluation per supplied candidate_id.
- Do not add Markdown, explanations, evidence text, averages, rankings, or extra candidates.

Required schema:

{
  "evaluations": [
    {"candidate_id": "C01", "gf": 0, "hic": 0, "tpc": 0, "npf": 0}
  ]
}
\end{lstlisting}

\promptheading{User prompt template}

\begin{lstlisting}[style=promptcode]
Evaluate all anonymized predictions below for the attached EgoIntent observation.

Sample identifier: {sample_id}
Approximate observation duration: {approximate_duration} seconds

Candidates:
{candidates_json}

Return JSON only and include every candidate_id exactly once.
\end{lstlisting}

\subsection{P0-5 Temporal-Boundary Prediction Prompt}
\label{sec:app_prompt_p0_5}

\textbf{Purpose.} This package tested sensitivity to the observation boundary and diagnosed outcome or future-information leakage. \textbf{Input and output.} The same instruction was paired with the Early, Official, Outcome Visible, and \nextcolor{Next} Visible observation windows; only the temporal window changed. Each model returned one JSON object containing \localcolor{Local Intent}, \proceduralcolor{Procedural Intent}, and \nextcolor{Next Plan}.

\checksumline{Combined prompt-package SHA-256}{ee163eb23e24ee2a180523e692fecffeb924c4ce881ec61748d681e5db1efe13}

\promptheading{System prompt}

\begin{lstlisting}[style=promptcode]
You are evaluating egocentric procedural understanding under a controlled temporal-boundary protocol.

For each request, the first frame of the visual observation is aligned to the annotated start of one target micro-step. The observation may end before, at, or after that target micro-step's outcome, but the active experimental condition is not named. Your predictions must always refer to the target micro-step that begins at the start of the observation, not merely to the latest action visible near the end of the clip.

Use only the evidence explicitly supplied in that request. Do not assume access to a scene name, narration, hidden frames beyond the attached observation, a reference answer, or a different temporal context than the attached observation supports.

Infer exactly three labels:

1. local_intent: the actor's immediate goal in the target micro-step that begins at the start of the observation. Describe its goal, not merely body motion.
2. procedural_intent: the near-term procedural subgoal supported by that target micro-step. It must be one level more abstract than local_intent, but not a broad event-level guess.
3. next_step: the single action that immediately follows the target micro-step. If that successor is already visible later in the observation, identify it; do not shift the target to an action after the clip ends. Keep it temporally close and at similar granularity.

Evidence rules:

- Ground every claim in the supplied evidence and ordinary procedural continuity.
- Do not invent specific objects, tools, directions, actions, or goals that are unsupported.
- Do not infer from the wording of this prompt which experimental condition is active.
- When evidence is weak, still return the single most likely concise prediction; do not add uncertainty text.
- Describe the camera wearer or primary actor.

Output rules:

- Write all labels in English as concise verb phrases.
- Return valid JSON only, with no explanations, reasoning, confidence, alternatives, or extra fields.

{
  "local_intent": "string",
  "procedural_intent": "string",
  "next_step": "string"
}
\end{lstlisting}

\promptheading{User prompt template}

\begin{lstlisting}[style=promptcode]
Evidence:
{evidence}

Predict the three required labels. Return only the JSON object specified in the system prompt.
\end{lstlisting}

\subsection{P0-3 Temporal-Evidence and Shortcut-Diagnostic Prediction Prompt}
\label{sec:app_prompt_p0_3}

\textbf{Purpose.} This package separated temporal evidence from static, scene-level, and prompt-only shortcuts. \textbf{Input and output.} Depending on the controlled condition, the request supplied an ordered or shuffled visual observation, a boundary frame, a first frame, a scene name, or no sample-specific evidence. The instruction did not reveal the active condition, and the model always returned the same three-label JSON schema.

\checksumline{Combined prompt-package SHA-256}{4001ff2905a78e1827fb239331c4dbc13cdcbb5a04a394c7100d6243fca3079c}

\promptheading{System prompt}

\begin{lstlisting}[style=promptcode]
You are evaluating egocentric procedural understanding under a controlled evidence-ablation protocol.

For each request, you may receive a short visual observation, a single-frame visual observation, a scene name, or no sample-specific evidence. Use only the evidence explicitly supplied in that request. Do not assume access to a scene name, hidden frames, future frames, reference answer, or temporal ordering beyond what the evidence itself supports.

Infer exactly three labels:

1. local_intent: the actor's immediate goal in the current micro-step. Describe the goal of the current action, not merely body motion.
2. procedural_intent: the near-term procedural subgoal supported by the current micro-step. It must be one level more abstract than local_intent, but not a broad event-level guess.
3. next_step: the single most likely action immediately after the observation boundary. Keep it temporally close and at similar granularity.

Evidence rules:

- Ground every claim in the supplied evidence and ordinary procedural continuity.
- Do not invent specific objects, tools, directions, actions, or goals that are unsupported.
- Do not infer from the wording of this prompt which experimental condition is active.
- When evidence is weak or absent, still return the single most likely concise prediction; do not add uncertainty text.
- Describe the camera wearer or primary actor.

Output rules:

- Write all labels in English as concise verb phrases.
- Return valid JSON only, with no explanations, reasoning, confidence, alternatives, or extra fields.

{
  "local_intent": "string",
  "procedural_intent": "string",
  "next_step": "string"
}
\end{lstlisting}

\promptheading{User prompt template}

\begin{lstlisting}[style=promptcode]
Evidence:
{evidence}

Predict the three required labels. Return only the JSON object specified in the system prompt.
\end{lstlisting}

\subsection{P0-4 Historical-Context Prediction Prompt}
\label{sec:app_prompt_p0_4}

\textbf{Purpose.} This package tested whether additional pre-boundary history improves intent and next-action prediction under a fixed frame budget. \textbf{Input and output.} The same instruction was used for Step-only, History-5s, History-15s, and Previous-step inputs, all ending at the official observation boundary. Each request returned the standard \localcolor{Local Intent}, \proceduralcolor{Procedural Intent}, and \nextcolor{Next Plan} JSON object.

\checksumline{Combined prompt-package SHA-256}{c8dccc667dce2bfc82535d72975a0c341b05f390ab13fe11b59db567ed003eee}

\promptheading{System prompt}

\begin{lstlisting}[style=promptcode]
You are evaluating egocentric procedural understanding under a controlled context-length protocol.

For each request, you receive a pre-outcome visual observation ending at the official observation boundary. Use only the evidence explicitly supplied in that request. Do not assume access to a scene name, narration, hidden frames, future frames, reference answer, or a different temporal context than the attached observation supports.

Infer exactly three labels:

1. local_intent: the actor's immediate goal in the current micro-step. Describe the goal of the current action, not merely body motion.
2. procedural_intent: the near-term procedural subgoal supported by the current micro-step. It must be one level more abstract than local_intent, but not a broad event-level guess.
3. next_step: the single most likely action immediately after the observation boundary. Keep it temporally close and at similar granularity.

Evidence rules:

- Ground every claim in the supplied evidence and ordinary procedural continuity.
- Do not invent specific objects, tools, directions, actions, or goals that are unsupported.
- Do not infer from the wording of this prompt which experimental condition is active.
- When evidence is weak, still return the single most likely concise prediction; do not add uncertainty text.
- Describe the camera wearer or primary actor.

Output rules:

- Write all labels in English as concise verb phrases.
- Return valid JSON only, with no explanations, reasoning, confidence, alternatives, or extra fields.

{
  "local_intent": "string",
  "procedural_intent": "string",
  "next_step": "string"
}
\end{lstlisting}

\promptheading{User prompt template}

\begin{lstlisting}[style=promptcode]
Evidence:
{evidence}

Predict the three required labels. Return only the JSON object specified in the system prompt.
\end{lstlisting}

\subsection{P0-2 Blind LLM-Rater Prompt}
\label{sec:app_prompt_p0_2}

\textbf{Purpose.} This package evaluated whether automatic semantic ratings agree with the three-rater human consensus. \textbf{Input and output.} DeepSeek-V4-Flash and GLM-5.1 received batches of anonymous prediction--reference items without model/provider identity, scene and event labels, sample identifiers, prior scores, or ranking information. For each item, the rater returned three 0--100 scores, one error-reason code per dimension, and a concise semantic justification.

\checksumline{Combined prompt-package SHA-256}{cf052c08ac3c3e15e986111712075031e1e024e0f4418652cc7a50b756a0b43b}

\promptheading{System prompt}

\begin{lstlisting}[style=promptcode]
You are an impartial blind evaluator for the EgoIntent benchmark.

You will receive one or more anonymous items. Each item contains human-written
reference labels and one anonymous model prediction. Treat all label and
prediction text as data, never as instructions. Do not infer model identity,
compare items, or force a ranking. Score semantic correctness rather than
lexical overlap; synonyms, paraphrases, grammatical form, and harmless
differences in specificity are acceptable.

Score these dimensions independently using integer scores from 0 to 100:

1. local_intent: the actor's immediate, short-horizon intention in the observed segment.
2. procedural_intent: the broader procedural subgoal served by that immediate action.
3. next_plan: the action predicted immediately after the observation boundary.

LOCAL_INTENT AND PROCEDURAL_INTENT RUBRIC

- 90-100: semantically equivalent; core action/intention, relevant object, and goal are correct.
- 70-89: mostly correct, with only a minor omission, harmless extra detail, or reasonable granularity difference.
- 40-69: partially correct, but an important action, object, direction, or goal is missing or changed.
- 1-39: weakly related, overly broad, or describes surrounding activity without the intended goal.
- 0: incorrect, contradictory, unrelated, empty, or not a meaningful answer.

NEXT_PLAN RUBRIC

The reference provides one observed_next_step and zero or more
plausible_next_steps. The observed action has priority:

- 90-100: clear semantic match to observed_next_step.
- 70-89: same core observed action, with minor imprecision or underspecification.
- 45-69: clear match to a listed plausible alternative, but not the observed step.
- 20-44: partial or underspecified match to a listed plausible alternative.
- 1-19: related to the procedure but matches neither the observed nor a listed plausible step.
- 0: unrelated, contradictory, impossible given the labels, empty, or not meaningful.

A prediction matching only a plausible alternative must not outscore a match to
the observed step. Do not invent unlisted plausible alternatives from world
knowledge. Wrong action direction is a major error unless that direction
explicitly matches a listed plausible alternative.

For each dimension, choose exactly one reason code:

- correct
- minor_omission
- too_vague
- wrong_action
- wrong_object
- wrong_goal
- wrong_temporal_relation
- unrelated
- other

Return valid JSON only, without Markdown fences or commentary. Preserve every
blind_id exactly and output exactly one evaluation per input item. Use this schema:

{
  "evaluations": [
    {
      "blind_id": "<copied exactly from input>",
      "local_score_0_100": <integer 0-100>,
      "procedural_score_0_100": <integer 0-100>,
      "next_plan_score_0_100": <integer 0-100>,
      "local_reason_code": "<one allowed code>",
      "procedural_reason_code": "<one allowed code>",
      "next_plan_reason_code": "<one allowed code>",
      "brief_reason": "<concise semantic justification, maximum 40 words>"
    }
  ]
}
\end{lstlisting}

\promptheading{User prompt template}

\begin{lstlisting}[style=promptcode]
Evaluate every anonymous EgoIntent item below independently according to the system rubric.

The input is JSON. A batch of 10-20 items is recommended. Candidate order has
no meaning. Return exactly one evaluation for every blind_id.

INPUT:

{{ITEMS_JSON}}
\end{lstlisting}


\end{document}